\documentclass{article}

\usepackage[final,main]{neurips_2026}

\usepackage[utf8]{inputenc}
\usepackage[T1]{fontenc}
\usepackage{hyperref}
\hypersetup{
  pdftitle={Uncertainty Quantification for Computer-Use Agents: A Benchmark across Vision-Language Models and GUI Grounding Datasets},
  pdfauthor={Divake Kumar, Sina Tayebati, Devashri Naik, Amanda Sofie Rios, Nilesh Ahuja, Omesh Tickoo, Ranganath Krishnan, Amit Ranjan Trivedi},
  pdfsubject={NeurIPS 2026},
  pdfkeywords={uncertainty quantification, computer-use agents, GUI agents, GUI grounding, vision-language models, VLM, benchmark, conformal prediction, selective prediction, calibration, Argus, NeurIPS 2026}
}
\usepackage{url}
\usepackage{booktabs}
\usepackage{amsfonts}
\usepackage{amsmath}
\usepackage{microtype}
\usepackage{xcolor}
\usepackage{colortbl}
\usepackage{graphicx}
\usepackage{multirow}
\usepackage{array}
\usepackage{caption}
\usepackage{subcaption}
\usepackage{enumitem}
\usepackage[table]{xcolor}
\usepackage{booktabs}
\usepackage{graphicx}
\usepackage[table]{xcolor}
\usepackage{makecell}
\usepackage{array}
\usepackage{makecell}
\usepackage{multirow}
\usepackage[table]{xcolor}
\usepackage{tikz}
\usepackage{pifont}
\usepackage{placeins}
\usepackage{fancyvrb}
\usepackage{float}
\floatstyle{plain}
\newfloat{listing}{tbp}{lol}
\floatname{listing}{Listing}
\DeclareRobustCommand{\cmark}{\textcolor{black!85}{\ding{51}}}
\DeclareRobustCommand{\xmark}{\textcolor{black!50}{\ding{55}}}

\definecolor{axisbg}{HTML}{EEF3F8}
\definecolor{scopebg}{HTML}{FFF4E6}
\definecolor{recordbg}{HTML}{EEF8EE}

\definecolor{tier1}{HTML}{F4B183}
\definecolor{tier2}{HTML}{FCE4D6}
\definecolor{tier3}{HTML}{E2F0D9}
\usepackage{booktabs}
\definecolor{keybg}{HTML}{E8F0FE}

\newcommand{\twolinepair}[3]{\shortstack[c]{#1\,/\,#2\\[1pt]{\tiny #3}}}

\newcommand{\famtag}[1]{\textsc{#1}}
\definecolor{tier1}{HTML}{8FB8E0}
\definecolor{tier2}{HTML}{C2D9EE}
\definecolor{tier3}{HTML}{E2ECF6}
\definecolor{rowband}{HTML}{F8F8F8}
\definecolor{posdelta}{HTML}{D9EAD3}
\definecolor{negdelta}{HTML}{F4CCCC}

\title{Uncertainty Quantification for Computer-Use Agents: A Benchmark across Vision-Language Models and GUI Grounding Datasets}

\author{%
  \bfseries Divake Kumar\textsuperscript{1} \quad Sina Tayebati\textsuperscript{1} \quad Devashri Naik\textsuperscript{1} \quad Amanda Sofie Rios\textsuperscript{2} \\[3pt]
  \bfseries Nilesh Ahuja\textsuperscript{2} \quad Omesh Tickoo\textsuperscript{2} \quad Ranganath Krishnan\textsuperscript{3} \quad Amit Ranjan Trivedi\textsuperscript{1} \\[5pt]
  {\normalfont \textsuperscript{1}University of Illinois Chicago \quad \textsuperscript{2}Intel Labs \quad \textsuperscript{3}Capital One AI Labs}%
}

\begin{document}

\maketitle

\begin{abstract}
Computer-use agents turn vision-language model (VLM) predictions into executable GUI clicks, so reliable uncertainty estimates are essential for rejection, calibration, miss-severity ranking, and spatial safety regions. Yet evidence on post-hoc uncertainty quantification (UQ) for these agents is fragmented across isolated model and dataset pairs, leaving it unclear whether UQ method rankings stay stable when the agent, benchmark, or observable interface changes. We present \textsc{Argus}, a cross-regime benchmark for post-hoc UQ in single-step executable GUI grounding, covering a $27$-method, seven-family open-weight matrix over $4$ GUI-grounding VLM agents and $4$ datasets, plus an $8$-method API-compatible closed-source matrix across $3$ frontier vendors where logits, hidden states, and attention maps are unavailable. The evaluated methods span logit-based scores, sampling and consistency measures such as semantic entropy and self-consistency, hidden-state and density estimators such as Mahalanobis and SAPLMA, attention-based scores, P(True) and verbalised-confidence prompting, and split-conformal prediction. The main finding is selective transfer: UQ rankings are stable across datasets for a fixed model, but degrade across model classes and observable interfaces. Hidden-state and density methods form the most stable open-weight family, while CoCoA-1MCA, Focus, sampling-based scores, and verbalised self-assessment win in specific regimes. Ranking transfer is strongest within a fixed model across datasets, reaching Spearman $\rho=0.969$ and averaging $\rho=0.705$ over $120$ open-weight pairs. In contrast, cross-tier transfer to closed-source vendors is much weaker: Spearman $\rho$ averages only $+0.08$ over $12$ vendor$\times$dataset pairs on the shared $8$-method intersection, so closed-source UQ recommendations should be reranked on the target rather than extrapolated. Model-class transitions further reshape UQ preferences: attention, verbalised, and VLM-native families lose AUROC on every dataset, while density methods remain stable; a scale-only baseline shows that logit-family degradation on \textsc{ScreenSpot-Pro} is partly confounded by scale rather than grounding fine-tuning alone. Finally, conformal click regions show that score-level discrimination is not enough for deployment: locally weighted disks can shrink radii by $40$--$60\%$ when the plug-in UQ is calibrated, but coverage can degrade under calibration-test or interface mismatch. We release per-item records, calibration/test splits, UQ scores, and analysis scripts as a reproducible basis for regime-aware UQ selection in GUI agents.
\end{abstract}

\section{Introduction}
\label{sec:intro}

Computer-use agents transform GUIs into executable actions. In single-step grounding, an agent receives an instruction and screenshot, then predicts a click coordinate $(x,y)$ to execute on the host system. Recent VLMs such as Qwen2.5-VL~\citep{bai2025qwen25vl}, UI-TARS~\citep{qin2025uitars}, POINTS-GUI~\citep{pointsgui2025}, and SeeClick~\citep{cheng2024seeclick} have improved this primitive. Yet errors remain common on harder benchmarks such as \textsc{ScreenSpot-Pro}~\citep{wang2025screenspotpro} and \textsc{OSWorld-G}~\citep{xie2025osworldg}. Because each prediction is an action rather than a passive label, uncertainty must support execution, rejection, deferral, and spatial safety regions.

Many post-hoc UQ families could provide action-level risk signals, including logit confidence~\citep{guo2017calibration,malinin2018predictive}, sampling consistency~\citep{wang2023selfconsistency,kuhn2023semantic,farquhar2024semantic}, hidden-state probes~\citep{azaria2023saplma,kossen2024sep}, density estimators~\citep{lee2018mahalanobis,ren2023rmd}, attention-trace scores~\citep{vazhentsev2023rauq,uqac2025,zhang2023focus}, and verbalised self-assessment~\citep{kadavath2022pktrue,tian2023verbalised1s,xiong2024verbalised2s}. Existing toolkits and GUI-grounding studies~\citep{fadeeva2024lmpolygraph,vashurin2025lmpolygraph,vashurin2025cocoa,safeground2026,hyperclick2025,uizoomer2026,v2p2025} usually evaluate one model family, one benchmark, one interface, or one objective. It remains unclear whether a UQ method selected in one deployment regime transfers to another.

This paper asks: \textit{when does UQ generalize in computer-use agents?} We define a regime by the agent, dataset, and observable model interface, and study transfer through the stability of UQ method rankings across regimes. The question is operational: a near miss may be recoverable, while a click on an unrelated control may trigger an unintended action. A useful UQ signal must support error detection, selective execution, calibration, graded miss-severity ranking, and spatial click-region coverage, which need not select the same method. UQ quality is therefore not method-intrinsic; it depends on where the score is observed and how it is used.

We introduce \textsc{Argus} (\underline{\textbf{A}}ssessing \underline{\textbf{R}}egime-wise \underline{\textbf{G}}eneralization of \underline{\textbf{U}}ncertainty \underline{\textbf{S}}coring), a unified benchmark for post-hoc UQ in single-step executable GUI grounding. The name follows Argus Panoptes, the hundred-eyed watchman of Greek myth, whose many eyes mirror the many uncertainty estimates and regimes the benchmark examines. \textsc{Argus} covers $27$ methods from seven families across $4$ open-weight VLM agents and $4$ GUI-grounding datasets ($16$ cells), plus an $8$-method harmonised API-compatible closed-source panel across $3$ frontier vendors on the same datasets. It evaluates error discrimination, selective execution, calibration, graded severity, ranking transfer, and conformal click-disk coverage. For cross-tier comparison, methods requiring token logprobs, hidden states, or attention maps are dropped from the API-only panel than replaced by proxies, so shared method use the same formula across panels.

Our results show selective generalization. UQ rankings are most stable across datasets for a fixed model, but change across model classes and observable interfaces. Hidden-state and density methods are the most stable open-weight families, while CoCoA-1MCA, Focus, sampling-based scores, HEDGE / IMGHEDGE, and verbalised self-assessment win in specific regimes. Vanilla-to-specialist transitions reshape which UQ families are reliable, API-only reasoning models increase the utility of verbalised self-assessment, and conformal click regions show that score-level discrimination alone does not guarantee spatial coverage. The paper makes five contributions: \textbf{(i) Cross-regime benchmark:} post-hoc UQ for executable GUI grounding across open-weight and API-only interfaces. \textbf{(ii) Ranking-transfer analysis:} where UQ method rankings generalize and where they break across agents, datasets, and interfaces. \textbf{(iii) Harmonised cross-tier protocol:} API-compatible comparison without proxy substitution for unavailable internal-signal methods. \textbf{(iv) Regime-dependent reliability:} evidence that model class, API observability, and spatial failure geometry change which UQ families are reliable. \textbf{(v) Deployment evaluation and release:} selective execution, calibration, graded miss severity, and conformal click-region evaluations, with released records, splits, UQ scores.

\section{Related Work}
\label{sec:related}

\noindent\textbf{Computer-use agents and GUI grounding.}
\textsc{ScreenSpot-Pro}~\citep{wang2025screenspotpro} and \textsc{ScreenSpot-v2}~\citep{wu2024osatlas} are reference benchmarks for single-step GUI grounding, while \textsc{OSWorld-G}/Jedi~\citep{xie2025osworldg} extends this setting to refusal-aware desktop grounding. Broader suites such as OSWorld~\citep{xie2024osworld}, AgentBoard~\citep{ma2024agentboard}, VisualWebArena~\citep{koh2024visualwebarena}, WebSuite~\citep{li2024websuite}, and OpenCUA~\citep{wang2025opencua} evaluate multi-step computer-use behavior. Related models, including UGround~\citep{gou2025uground}, SeeClick~\citep{cheng2024seeclick}, UI-TARS~\citep{qin2025uitars}, and POINTS-GUI~\citep{pointsgui2025}, establish click grounding as a core perception-action primitive. We focus on single-step grounding because it isolates the executable click from planning, recovery, and credit assignment.

\noindent\textbf{Post-hoc UQ for language and vision-language models.}
The methods evaluated here span logit confidence~\citep{guo2017calibration,malinin2018predictive}, sampling consistency~\citep{wang2023selfconsistency,kuhn2023semantic,farquhar2024semantic}, hidden-state probes~\citep{azaria2023saplma,kossen2024sep}, density estimators~\citep{lee2018mahalanobis,ren2023rmd}, attention-trace scores~\citep{vazhentsev2023rauq,zhang2023focus,uqac2025}, and verbalised confidence~\citep{kadavath2022pktrue,tian2023verbalised1s,xiong2024verbalised2s}. LM-Polygraph~\citep{fadeeva2024lmpolygraph,vashurin2025lmpolygraph} provides reference implementations for text-generation UQ, CoCoA~\citep{vashurin2025cocoa} unifies confidence and consistency under a Minimum-Bayes-Risk objective, and Torch-Uncertainty~\citep{torchuncertainty2024} provides tools relevant to regression and selective prediction. These works define useful UQ signals, but they do not determine whether method rankings generalize across GUI agents, datasets, and observable interfaces. The missing object is not another UQ score, but evidence about \emph{generalization of UQ choice}: whether a method that ranks errors well for one GUI agent remains reliable after changing the model class, benchmark geometry, or observable interface.

\noindent\textbf{Uncertainty for GUI grounding.}
SafeGround~\citep{safeground2026} uses stochastic click dispersion and Learn-Then-Test calibration~\citep{angelopoulos2021ltt} for abstention. HyperClick~\citep{hyperclick2025} learns a truncated-Gaussian spatial-confidence head calibrated by Brier score. UI-Zoomer~\citep{uizoomer2026} uses uncertainty to trigger adaptive zoom-and-regrounding, and V2P~\citep{v2p2025} calibrates visual attention by suppressing background regions. In contrast, we ask when post-hoc UQ choices generalize across regimes, including open-weight internals, API-only interfaces, different grounding agents, dataset shifts, and multiple deployment metrics.

\noindent\textbf{Conformal prediction, abstention, and risk-aware UQ.}
A complementary line of work turns uncertainty into action: conformal abstention policies convert calibrated uncertainty into adaptive, context-dependent risk management and deferral for language and vision-language models~\citep{tayebati2025conformalabstention,tayebati2025cap}, and task-dependent uncertainty in multimodal judging shows that VLM judges can rank but not reliably score~\citep{kumar2026vlm}, which informs how we treat verbalised and VLM-native UQ here. We build on these ideas, but shift the question from producing one calibrated score to measuring whether post-hoc UQ choices transfer across GUI agents, datasets, and observable interfaces. A broader discussion of conformal prediction, calibration, and risk-aware deployment is given in Appendix~\ref{appx:extended_related}.

\begin{table}[!htbp]
\centering
\caption{\textbf{\textsc{Argus} positioning and public API.}
Left: comparison against prior VL-UQ benchmarks and GUI-grounding UQ papers.
Right: minimal \texttt{argus-uq} example for loading a cell, scoring a UQ method, and constructing a conformal click disk.}
\label{tab:related_api}
\scriptsize
\setlength{\tabcolsep}{3pt}
\renewcommand{\arraystretch}{1.02}

\begin{minipage}[t]{0.56\linewidth}
\centering
\textbf{(a) Benchmark positioning}\\[-0.5mm]

\begin{tabular}{l cccc}
\toprule
Work & Panel & Closed & Conf. & GUI \\
\midrule
\multicolumn{5}{l}{\emph{Multi-method VL-UQ}}\\
VL-Uncertainty~\citep{zhang2024vluncertainty}
    & \xmark & \xmark & \xmark & \xmark \\
VLM-UQ-Bench~\citep{kostumov2024vlmuqbench}
    & \xmark & \xmark & \cmark & \xmark \\
Conformal Lens~\citep{azad2025conformallens}
    & \xmark & \xmark$^{\dagger}$ & \cmark & \xmark \\
\midrule
\multicolumn{5}{l}{\emph{GUI-grounding UQ}}\\
SafeGround~\citep{safeground2026}
    & -- & \xmark & \cmark$^{\ddagger}$ & \cmark \\
HyperClick~\citep{hyperclick2025}
    & -- & \xmark & \xmark & \cmark \\
UI-Zoomer~\citep{uizoomer2026}
    & -- & \xmark & \xmark & \cmark \\
V2P~\citep{v2p2025}
    & -- & \xmark & \xmark & \cmark \\
\midrule
\rowcolor{tier1}
\textbf{\textsc{Argus}}
    & \textbf{\cmark} & \textbf{\cmark}
    & \textbf{\cmark} & \textbf{\cmark} \\
\bottomrule
\end{tabular}

\vspace{0.5mm}
\raggedright
\scriptsize
\cmark\,covers; \xmark\,does not cover; ``--'' not applicable.
Panel means $\geq\!10$ post-hoc UQ scores.
$^{\dagger}$MCQ likelihood proxies only.
$^{\ddagger}$Single released method partially reproduced.
\end{minipage}
\hfill
\begin{minipage}[t]{0.41\linewidth}
\centering
\textbf{(b) Minimal API example}\\[-0.5mm]

\begin{Verbatim}[fontsize=\tiny, frame=single, framesep=2pt,
xleftmargin=2pt, xrightmargin=2pt]
import argus_uq

cell = argus_uq.load("PTxOSG")

result = argus_uq.score(
    cell, method="saplma", seeds=50)
print(result.auroc, result.auroc_std)

disk = argus_uq.conformal_disk(
    cell, plug_in="saplma",
    alpha=0.10, variant="normalized")
print(disk.radius_pixels, disk.coverage)
\end{Verbatim}

\vspace{0.5mm}
\raggedright
\scriptsize
The API loads a benchmark cell, scores a UQ method, and constructs a conformal click disk.
\end{minipage}
\end{table}


Table 1 compares \textsc{Argus} against the closest VL-UQ benchmarks and GUI-grounding UQ papers. No prior work jointly covers GUI grounding, a multi-method UQ panel, closed-source vendor support, conformal click-disks, and multi-model evaluation. The GUI-grounding baselines are single-method papers, so their UQ-panel entry is not applicable. We release \textsc{Argus} as \texttt{argus-uq}, a Python package with three core calls: \texttt{load} for cells, \texttt{score} for AUROC / PRR / AUSE / ECE / Brier / AURC over the $27$ methods, and \texttt{conformal\_disk} for deployable click regions. Listing in Table 1 right gives the quickstart; package details and extended usage are in Appendix~\ref{appx:argus_uq}.

\begin{table}[!htbp]
\centering
\caption{\textbf{Evaluation regimes.} Open-weight models (Qwen2.5-VL-7B [Q7], Qwen2.5-VL-72B-AWQ [Q72], UI-TARS-1.5-7B [UI], POINTS-GUI-G-8B [PT]) and API-only closed-source vendors (GPT-5.4, Claude Sonnet 4.6, Gemini 3.1 Pro) on four single-step GUI-grounding datasets. Reported accuracy is greedy point-in-bbox accuracy under the single-shot protocol.}
\label{tab:cells}
\footnotesize
\setlength{\tabcolsep}{3.5pt}
\renewcommand{\arraystretch}{1.05}

\begin{tabular}{@{}lc@{\hspace{1.4em}}lc@{\hspace{1.4em}}lc@{\hspace{1.4em}}lc@{}}
\toprule
\multicolumn{2}{c}{\textbf{Open-weight (Q7, Q72)}} &
\multicolumn{2}{c}{\textbf{Open-weight (UI, PT)}} &
\multicolumn{2}{c}{\textbf{API-only (GPT, Sonnet)}} &
\multicolumn{2}{c}{\textbf{API-only (Gemini)}} \\
\cmidrule(lr){1-2}
\cmidrule(lr){3-4}
\cmidrule(lr){5-6}
\cmidrule(l){7-8}
Cell & Acc. &
Cell & Acc. &
Cell & Acc. &
Cell & Acc. \\
\midrule
Q7$\times$V2    & .883 & Q72$\times$V2   & .927 & GPT$\times$V2      & .870 & Gemini$\times$V2   & .447 \\
Q7$\times$SP    & .274 & Q72$\times$SP   & .447 & GPT$\times$SP      & .367 & Gemini$\times$SP   & .313 \\
Q7$\times$OSG   & .341 & Q72$\times$OSG  & .535 & GPT$\times$OSG     & .667 & Gemini$\times$OSG  & .311 \\
Q7$\times$UIV   & .152 & Q72$\times$UIV  & .282 & GPT$\times$UIV     & .485 & Gemini$\times$UIV  & .444 \\
\addlinespace[0.8mm]
UI$\times$V2    & .878 & PT$\times$V2    & .955 & Sonnet$\times$V2   & .452 &                    &       \\
UI$\times$SP    & .407 & PT$\times$SP    & .583 & Sonnet$\times$SP   & .341 &                    &       \\
UI$\times$OSG   & .513 & PT$\times$OSG   & .659 & Sonnet$\times$OSG  & .363 &                    &       \\
UI$\times$UIV   & .214 & PT$\times$UIV   & .531 & Sonnet$\times$UIV  & .369 &                    &       \\
\bottomrule
\end{tabular}

\vspace{0.6mm}
\begin{minipage}{0.98\linewidth}
\scriptsize
\textit{Notes.}
OSG = \textsc{OSWorld-G}, SP = \textsc{ScreenSpot-Pro}, V2 = \textsc{ScreenSpot-v2}, UIV = \textsc{UI-Vision-EG}.
Q7 = Qwen2.5-VL-7B, Q72 = Qwen2.5-VL-72B-AWQ, UI = UI-TARS-1.5-7B, PT = POINTS-GUI-G-8B.
Open-weight matrix: $4$ models $\times$ $4$ datasets $=16$ cells. Closed-source matrix: $3$ vendors $\times$ $4$ datasets $=12$ cells.
\end{minipage}
\end{table}
\begin{table}[!htbp]
\centering
\caption{\textbf{Protocol card.} Evaluation scope, observability, metrics, leakage controls, and releases.}
\label{tab:protocol_card}
\footnotesize
\setlength{\tabcolsep}{3.5pt}
\renewcommand{\arraystretch}{1.04}
\begin{tabular}{@{}>{\columncolor{axisbg}}p{0.13\linewidth}p{0.84\linewidth}@{}}
\toprule
\textbf{Axis} & \textbf{Protocol detail} \\
\midrule

\textbf{Datasets} &
Four GUI-grounding benchmarks: \textsc{ScreenSpot-v2}, \textsc{ScreenSpot-Pro}, \textsc{OSWorld-G}, and \textsc{UI-Vision-EG}. Greedy predictions are scored by point-in-bbox accuracy. \\

\textbf{Models} &
Four open-weight agents: Qwen2.5-VL-7B (Q7), Qwen2.5-VL-72B-AWQ (Q72), UI-TARS-1.5-7B (UI), and POINTS-GUI-G-8B (PT). API-only vendors: GPT-5.4, Claude Sonnet 4.6, and Gemini 3.1 Pro. \\

\textbf{Interfaces} &
Open-weight cells expose logits, hidden states, attention traces, stochastic samples, and perturbation responses; API-only cells expose text/coordinate responses and use the API-compatible UQ subset. \\

\textbf{Inference} &
Matched prompting and decoding with full-resolution images. Per item: one greedy click, $n_{\text{samples}}=5$ stochastic samples, $3$ HEDGE paraphrases, $3$ IMGHEDGE perturbations, and a $50$-px coordinate-clustering tolerance. \\

\textbf{UQ panel} &
$27$ open-weight methods across seven families: logit, sampling, hybrid, density/probe, attention, verbalised, and VLM-native. API-only cells use an $8$-method harmonised subset; methods requiring unavailable internals are dropped rather than proxied. \\

\textbf{Metrics} &
AUROC$_{\mathrm{incorrect}}$, PRR$_{0.5}^{\mathrm{norm}}$, AURC, ECE$^{\mathrm{iso}}$, Brier$^{\mathrm{iso}}$, AUSE, ranking-transfer Spearman $\rho$, and conformal click-disk coverage/radius. \\

\textbf{Leakage control} &
Learned probes, density estimators, isotonic maps, conformal thresholds, and panel selection use only calibration splits; all headline metrics use held-out test records. \\

\textbf{Splits / release} &
Headline $80/20$ test/calibration split repeated over $50$ seeds; split-ratio ablations are in Appendix~\ref{appx:split_ablation}. Released records include clicks, labels, samples, perturbation/API responses, internals, split seeds, UQ scores, and table scripts. \\

\bottomrule
\end{tabular}
\end{table}

\section{Benchmark and Evaluation Protocol}
\label{sec:benchmark}

We construct a cross-regime benchmark for post-hoc UQ in single-step executable GUI click grounding. A regime is defined by the dataset, model class, and observable model interface. The open-weight panel evaluates $27$ methods from seven UQ families on $4$ VLM agents (Qwen2.5-VL-7B, Qwen2.5-VL-72B-AWQ, UI-TARS-1.5-7B, POINTS-GUI-G-8B) across $4$ GUI-grounding datasets (\textsc{ScreenSpot-v2}, \textsc{ScreenSpot-Pro}, \textsc{OSWorld-G}, and \textsc{UI-Vision-EG}~\citep{nayak2025uivision}), giving $16$ cells where logits, hidden states, attention traces, stochastic samples, and perturbation responses are all available. The closed-source extension evaluates the API-compatible subset across $3$ frontier vendors (GPT-5.4, Claude Sonnet 4.6, Gemini 3.1 Pro) on the same datasets, for settings where logits, hidden states, and attention maps are inaccessible. Tables~\ref{tab:cells} and~\ref{tab:protocol_card} define the evaluation cells, protocol, released records, and observability assumptions. We report AUROC$_{\mathrm{incorrect}}$, PRR$_{0.5}^{\mathrm{norm}}$, AURC, ECE$^{\mathrm{iso}}$, Brier$^{\mathrm{iso}}$, miss-only AUSE~\citep{ilg2018ause}, ranking-transfer Spearman $\rho$, and conformal click-disk coverage/radius (defined in the metric card, Table~\ref{tab:metric_card}). The benchmark is designed around UQ generalization rather than absolute model accuracy. Holding the model fixed while changing the dataset tests whether UQ rankings survive dataset and target-geometry shifts. Holding the dataset fixed while changing the model tests whether rankings survive architecture, fine-tuning, scale, and quantization changes. Moving from open-weight to API-only cells tests whether UQ choice survives loss of internal observability.

Method inclusion follows three criteria: the method is reported in at least two peer-reviewed text-UQ or VLM-UQ evaluations, or is directly motivated by GUI-grounding UQ; it admits a faithful adaptation to coordinate-valued click outputs; and it can be computed within $30$ minutes per cell after per-item records are produced. Implementations are traceable to canonical references and cross-checked against LM-Polygraph where applicable; method-specific adaptations are documented in Appendix~\ref{appx:methods}, methods that fall outside our inclusion criteria are listed in Appendix~\ref{appx:deferred}, and the closed-source vendor protocol (single-shot inference, snapshot IDs, image-input policies, format-compliance handling) is detailed in Appendix~\ref{appx:closed_protocol}. We exclude monotonic duplicates, grounding-incompatible rejection-sampling methods, API-unavailable internal-state methods from API comparisons, and non-portable paper-specific training procedures. SafeGround-style spatial-dispersion scores are reported in Appendix~\ref{appx:safeground}, but not in the headline matrix because the main protocol uses $n_{\text{samples}}=5$, below the larger pure-stochastic sampling budget needed for reliable dispersion estimates.

Each UQ method is evaluated as an action-level risk score, with larger values indicating higher predicted error probability. Learned probes, density estimators, calibration maps, conformal thresholds, and panel-selection decisions use only the calibration split; all headline metrics are computed on held-out test records. For graded severity, we normalize click error by target scale,
\[
d_{\mathrm{norm},i} =
\frac{\lVert \mathbf{c}_i-\mathbf{c}^{\star}_i\rVert_2}{\sqrt{w_i h_i}},
\]
where $\mathbf{c}_i$ is the predicted click, $\mathbf{c}^{\star}_i$ is the target-box center, and $w_i,h_i$ are the target-box width and height. For conformal click disks, the fixed-radius baseline uses calibration residuals
\(
R_i=\lVert \mathbf{c}_i-\mathbf{c}^{\star}_i\rVert_2,
\qquad i\in\mathcal{D}_{\mathrm{calib}},
\)
and sets
\(
\hat r_{\alpha}
=
\mathrm{Quantile}_{1-\alpha}
\left(
\left\{R_i:i\in\mathcal{D}_{\mathrm{calib}}\right\}
\right),
\)
following split conformal prediction~\citep{vovk2005algorithmic,angelopoulos2021gentle}. Unless otherwise stated, coverage means that the disk centered at the predicted click contains the target-box center. All metrics include $95\%$ confidence intervals from $500$-resample stratified bootstrap; significance claims use paired bootstrap differences. Per-cell top-$1$ values with 95\% bootstrap CI brackets are reported in Appendix~\ref{appx:ci_brackets}; per-cell top-$1$ vs top-$2$ paired Wilcoxon signed-rank tests (BH-FDR $q<0.05$) are in Appendix~\ref{appx:significance}. The $80/20$ split ratio is the protocol default; method rankings remain stable across cal/test ratios in $\{10/90, 20/80, 30/70, 40/60, 50/50\}$ with mean cross-ratio Spearman $\rho=0.98$ on open-weight ($16$ cells) and $\rho=0.92$ on closed-source ($12$ cells), reported in Appendix~\ref{appx:split_ablation}.

\begin{table}[!htbp]
\centering
\caption{\textbf{Headline benchmark.}
Panel A reports the AUROC$_{\mathrm{incorrect}}$-best method within each UQ family for each open-weight regime, with 50-seed mean AUROC / PRR$_{0.5}^{\mathrm{norm}}$ and method name. Panel B reports the same AUROC / PRR format for the harmonised API-only panel, laid out cell-wise (each vendor$\times$dataset cell is a row, each method a column). Bold marks the per-cell AUROC top-1 family (Panel A) or method (Panel B); shading marks per-cell top-1 / 2 / 3 tiers, computed separately for AUROC and PRR. Full method-level AUROC, PRR, AUSE, ECE$_{\mathrm{iso}}$, Brier$_{\mathrm{iso}}$, and AURC results are in Appendix~\ref{appx:full_open_matrix}.}
\label{tab:headline_family_auroc}
\scriptsize

\textbf{Panel A: Open-weight cells}

\vspace{0.3mm}

{\scriptsize
\setlength{\tabcolsep}{6.0pt}
\renewcommand{\arraystretch}{1}
\setlength{\extrarowheight}{0pt}
\resizebox{\textwidth}{!}{%
\begin{tabular}{@{}l c c c c c c c@{}}
\toprule
Cell & Logit & Sampling & Hybrid & Attention & Density/Probe & Verbalised & VLM-native \\
\midrule
Q7$\times$V2 & \twolinepair{\colorbox{tier3}{.741}}{\colorbox{tier3}{.421}}{MTE} & \twolinepair{.723}{.384}{MCSE} & \twolinepair{.736}{.404}{CoCoA-1MCA} & \twolinepair{\colorbox{tier2}{.764}}{\colorbox{tier2}{.451}}{RAUQ-full} & \twolinepair{\colorbox{tier1}{\textbf{.817}}}{\colorbox{tier1}{\textbf{.599}}}{SAPLMA} & \twolinepair{.668}{.286}{P(True)} & \twolinepair{.692}{.381}{HEDGE} \\
Q7$\times$SP & \twolinepair{\colorbox{tier2}{.873}}{\colorbox{tier1}{\textbf{.879}}}{SeqProb} & \twolinepair{.817}{.732}{SE-w} & \twolinepair{\colorbox{tier3}{.865}}{\colorbox{tier2}{.864}}{CoCoA} & \twolinepair{.846}{.822}{Focus} & \twolinepair{\colorbox{tier1}{\textbf{.883}}}{\colorbox{tier3}{.861}}{SEP} & \twolinepair{.682}{.527}{P(True)} & \twolinepair{.734}{.651}{IMGHEDGE} \\
Q7$\times$OSG & \twolinepair{.758}{\colorbox{tier3}{.721}}{MTE} & \twolinepair{\colorbox{tier3}{.769}}{.694}{SE-w} & \twolinepair{\colorbox{tier2}{.785}}{\colorbox{tier1}{\textbf{.772}}}{CoCoA-1MCA} & \twolinepair{.740}{.649}{RAUQ-full} & \twolinepair{\colorbox{tier1}{\textbf{.786}}}{\colorbox{tier2}{.723}}{SEP} & \twolinepair{.719}{.579}{P(True)} & \twolinepair{.672}{.551}{IMGHEDGE} \\
Q7$\times$UIV & \twolinepair{\colorbox{tier1}{\textbf{.816}}}{\colorbox{tier2}{.877}}{MTE} & \twolinepair{.783}{.724}{MCSE} & \twolinepair{.801}{.816}{CoCoA} & \twolinepair{\colorbox{tier3}{.812}}{\colorbox{tier1}{\textbf{.889}}}{RAUQ-full} & \twolinepair{\colorbox{tier2}{.815}}{\colorbox{tier3}{.841}}{SEP} & \twolinepair{.704}{.562}{P(True)} & \twolinepair{.693}{.651}{IMGHEDGE} \\
Q72$\times$V2 & \twolinepair{.664}{.267}{SeqProb} & \twolinepair{.733}{\colorbox{tier3}{.453}}{SE} & \twolinepair{\colorbox{tier2}{.751}}{\colorbox{tier2}{.472}}{CoCoA-1MCA} & \twolinepair{\colorbox{tier3}{.736}}{.399}{Focus} & \twolinepair{\colorbox{tier1}{\textbf{.764}}}{\colorbox{tier1}{\textbf{.505}}}{SAPLMA} & \twolinepair{.633}{.255}{Verb-1S} & \twolinepair{.674}{.344}{HEDGE} \\
Q72$\times$SP & \twolinepair{\colorbox{tier3}{.804}}{\colorbox{tier2}{.627}}{SeqProb} & \twolinepair{.505}{.010}{SelfCons} & \twolinepair{.741}{.502}{CoCoA-1MCA} & \twolinepair{\colorbox{tier2}{.806}}{.606}{Focus} & \twolinepair{\colorbox{tier1}{\textbf{.889}}}{\colorbox{tier1}{\textbf{.802}}}{SAPLMA} & \twolinepair{.654}{.404}{Verb-1S} & \twolinepair{.735}{\colorbox{tier3}{.614}}{IMGHEDGE} \\
Q72$\times$OSG & \twolinepair{.751}{.443}{SeqProb} & \twolinepair{.776}{\colorbox{tier2}{.574}}{SE-w} & \twolinepair{\colorbox{tier2}{.792}}{\colorbox{tier3}{.573}}{CoCoA-1MCA} & \twolinepair{\colorbox{tier3}{.788}}{.539}{Focus} & \twolinepair{\colorbox{tier1}{\textbf{.838}}}{\colorbox{tier1}{\textbf{.662}}}{SEP} & \twolinepair{.674}{.395}{P(True)} & \twolinepair{.677}{.433}{IMGHEDGE} \\
Q72$\times$UIV & \twolinepair{.779}{\colorbox{tier1}{\textbf{.743}}}{SeqProb} & \twolinepair{.753}{.639}{SE-w} & \twolinepair{\colorbox{tier2}{.791}}{\colorbox{tier2}{.742}}{CoCoA} & \twolinepair{\colorbox{tier3}{.779}}{\colorbox{tier3}{.742}}{Focus} & \twolinepair{\colorbox{tier1}{\textbf{.834}}}{.727}{SAPLMA} & \twolinepair{.671}{.416}{P(True)} & \twolinepair{.672}{.589}{HEDGE} \\
UI$\times$V2 & \twolinepair{.715}{.371}{MSP} & \twolinepair{\colorbox{tier2}{.808}}{\colorbox{tier2}{.581}}{SE} & \twolinepair{\colorbox{tier1}{\textbf{.842}}}{\colorbox{tier1}{\textbf{.630}}}{CoCoA-1MCA} & \twolinepair{\colorbox{tier3}{.771}}{.479}{RAUQ-full} & \twolinepair{.762}{\colorbox{tier3}{.483}}{SAPLMA} & \twolinepair{.601}{.167}{Verb-2S} & \twolinepair{.725}{.450}{HEDGE} \\
UI$\times$SP & \twolinepair{.780}{.663}{SeqProb} & \twolinepair{\colorbox{tier3}{.814}}{.703}{SE} & \twolinepair{\colorbox{tier2}{.837}}{\colorbox{tier2}{.783}}{CoCoA} & \twolinepair{.812}{\colorbox{tier3}{.734}}{Focus} & \twolinepair{\colorbox{tier1}{\textbf{.877}}}{\colorbox{tier1}{\textbf{.809}}}{SAPLMA} & \twolinepair{.582}{.223}{Verb-1S} & \twolinepair{.767}{.713}{IMGHEDGE} \\
UI$\times$OSG & \twolinepair{.718}{.399}{MSP} & \twolinepair{.773}{\colorbox{tier3}{.585}}{SE-w} & \twolinepair{\colorbox{tier2}{.799}}{\colorbox{tier2}{.599}}{CoCoA-1MCA} & \twolinepair{\colorbox{tier3}{.779}}{.518}{RAUQ-full} & \twolinepair{\colorbox{tier1}{\textbf{.863}}}{\colorbox{tier1}{\textbf{.709}}}{SEP} & \twolinepair{.635}{.332}{Verb-1S} & \twolinepair{.716}{.532}{HEDGE} \\
UI$\times$UIV & \twolinepair{.770}{.633}{SeqProb} & \twolinepair{.806}{.739}{SE} & \twolinepair{\colorbox{tier1}{\textbf{.825}}}{\colorbox{tier1}{\textbf{.793}}}{CoCoA} & \twolinepair{\colorbox{tier3}{.816}}{\colorbox{tier3}{.756}}{RAUQ-full} & \twolinepair{\colorbox{tier2}{.818}}{\colorbox{tier2}{.771}}{Mahal-RMD} & \twolinepair{.598}{.315}{Verb-1S} & \twolinepair{.728}{.728}{HEDGE} \\
PT$\times$V2 & \twolinepair{.703}{.391}{MSP} & \twolinepair{.676}{.326}{SE} & \twolinepair{\colorbox{tier2}{.738}}{\colorbox{tier2}{.440}}{CoCoA-1MCA} & \twolinepair{\colorbox{tier3}{.715}}{\colorbox{tier3}{.397}}{RAUQ-full} & \twolinepair{\colorbox{tier1}{\textbf{.843}}}{\colorbox{tier1}{\textbf{.657}}}{SEP} & \twolinepair{.535}{.058}{P(True)} & \twolinepair{.598}{.207}{IMGHEDGE} \\
PT$\times$SP & \twolinepair{\colorbox{tier2}{.780}}{\colorbox{tier3}{.524}}{MTE} & \twolinepair{.637}{.233}{SE-w} & \twolinepair{\colorbox{tier3}{.780}}{\colorbox{tier2}{.532}}{CoCoA} & \twolinepair{.779}{.520}{UQAC} & \twolinepair{\colorbox{tier1}{\textbf{.881}}}{\colorbox{tier1}{\textbf{.702}}}{SAPLMA} & \twolinepair{.620}{.228}{P(True)} & \twolinepair{.614}{.258}{IMGHEDGE} \\
PT$\times$OSG & \twolinepair{\colorbox{tier3}{.725}}{.407}{MTE} & \twolinepair{.614}{.216}{SelfCons} & \twolinepair{.724}{\colorbox{tier2}{.420}}{CoCoA} & \twolinepair{\colorbox{tier2}{.729}}{\colorbox{tier3}{.415}}{Focus} & \twolinepair{\colorbox{tier1}{\textbf{.820}}}{\colorbox{tier1}{\textbf{.568}}}{Mahal-RMD} & \twolinepair{.580}{.152}{P(True)} & \twolinepair{.570}{.157}{HEDGE} \\
PT$\times$UIV & \twolinepair{\colorbox{tier3}{.694}}{\colorbox{tier3}{.377}}{MTE} & \twolinepair{.673}{.369}{LexSim} & \twolinepair{\colorbox{tier2}{.699}}{\colorbox{tier2}{.398}}{CoCoA-1MCA} & \twolinepair{.693}{.353}{UQAC} & \twolinepair{\colorbox{tier1}{\textbf{.796}}}{\colorbox{tier1}{\textbf{.579}}}{SAPLMA} & \twolinepair{.517}{.049}{Verb-2S} & \twolinepair{.583}{.205}{HEDGE} \\
\bottomrule
\end{tabular}%
}
}

\vspace{1.0mm}
\textbf{Panel B: API-only closed-source cells}

\vspace{0.3mm}

{\scriptsize
\setlength{\tabcolsep}{5.0pt}
\renewcommand{\arraystretch}{1}
\setlength{\extrarowheight}{0pt}
\resizebox{\textwidth}{!}{%
\begin{tabular}{@{}l c c c c c c c c @{}}
\toprule
 & \multicolumn{3}{c}{\textsc{Sampling}} & \multicolumn{1}{c}{\textsc{Hybrid}} & \multicolumn{2}{c}{\textsc{Verbalised}} & \multicolumn{2}{c}{\textsc{VLM-native}} \\
\cmidrule(lr){2-4} \cmidrule(lr){5-5} \cmidrule(lr){6-7} \cmidrule(lr){8-9}
Cell & SelfCons & SE & LexSim & CCP & Verb-1S & Verb-2S & HEDGE & IMGHEDGE \\
\midrule
GPT$\times$\textsc{V2} & .678\,/\,.115 & \colorbox{tier3}{.680}\,/\,\colorbox{tier3}{.116} & .603\,/\,.035 & \colorbox{tier1}{\textbf{.798}}\,/\,\colorbox{tier1}{\textbf{.192}} & .556\,/\,.028 & .618\,/\,.070 & .592\,/\,.061 & \colorbox{tier2}{.685}\,/\,\colorbox{tier2}{.121} \\
GPT$\times$\textsc{SP} & .702\,/\,.513 & .704\,/\,.519 & .699\,/\,.510 & \colorbox{tier1}{\textbf{.735}}\,/\,\colorbox{tier2}{.574} & \colorbox{tier2}{.711}\,/\,\colorbox{tier1}{\textbf{.585}} & .674\,/\,.325 & .649\,/\,.423 & \colorbox{tier3}{.709}\,/\,\colorbox{tier3}{.524} \\
GPT$\times$\textsc{OSG} & .608\,/\,.185 & .608\,/\,.185 & \colorbox{tier3}{.674}\,/\,\colorbox{tier3}{.234} & .619\,/\,.204 & \colorbox{tier2}{.703}\,/\,\colorbox{tier2}{.336} & \colorbox{tier1}{\textbf{.770}}\,/\,\colorbox{tier1}{\textbf{.401}} & .620\,/\,.209 & .590\,/\,.157 \\
GPT$\times$\textsc{UIV} & .692\,/\,\colorbox{tier3}{.472} & .694\,/\,\colorbox{tier2}{.477} & \colorbox{tier3}{.706}\,/\,.431 & \colorbox{tier2}{.715}\,/\,\colorbox{tier1}{\textbf{.529}} & \colorbox{tier1}{\textbf{.723}}\,/\,.470 & .678\,/\,.315 & .644\,/\,.370 & .629\,/\,.331 \\
\addlinespace[0.8mm]
Sonnet$\times$\textsc{V2} & .507\,/\,.018 & .507\,/\,.018 & .383\,/\,$-$.246 & \colorbox{tier3}{.508}\,/\,\colorbox{tier3}{.019} & \colorbox{tier2}{.524}\,/\,\colorbox{tier2}{.084} & \colorbox{tier1}{\textbf{.648}}\,/\,\colorbox{tier1}{\textbf{.279}} & .479\,/\,$-$.051 & .483\,/\,$-$.046 \\
Sonnet$\times$\textsc{SP} & .684\,/\,.510 & \colorbox{tier3}{.690}\,/\,\colorbox{tier3}{.540} & .663\,/\,.433 & \colorbox{tier1}{\textbf{.731}}\,/\,\colorbox{tier1}{\textbf{.679}} & \colorbox{tier2}{.695}\,/\,\colorbox{tier2}{.607} & .604\,/\,.240 & .666\,/\,.447 & .669\,/\,.485 \\
Sonnet$\times$\textsc{OSG} & \colorbox{tier3}{.593}\,/\,\colorbox{tier3}{.262} & \colorbox{tier2}{.595}\,/\,\colorbox{tier2}{.273} & .569\,/\,.186 & \colorbox{tier1}{\textbf{.616}}\,/\,\colorbox{tier1}{\textbf{.330}} & .478\,/\,.039 & .549\,/\,.085 & .590\,/\,.207 & .567\,/\,.181 \\
Sonnet$\times$\textsc{UIV} & \colorbox{tier3}{.623}\,/\,\colorbox{tier3}{.311} & \colorbox{tier2}{.626}\,/\,\colorbox{tier2}{.326} & .583\,/\,.268 & \colorbox{tier1}{\textbf{.651}}\,/\,\colorbox{tier1}{\textbf{.411}} & .516\,/\,.187 & .520\,/\,.086 & .598\,/\,.261 & .610\,/\,.306 \\
\addlinespace[0.8mm]
Gemini$\times$\textsc{V2} & .557\,/\,.172 & .557\,/\,.172 & .624\,/\,.295 & .596\,/\,.266 & \colorbox{tier1}{\textbf{.966}}\,/\,\colorbox{tier1}{\textbf{.951}} & \colorbox{tier2}{.959}\,/\,\colorbox{tier2}{.890} & .516\,/\,.052 & \colorbox{tier3}{.875}\,/\,\colorbox{tier3}{.732} \\
Gemini$\times$\textsc{SP} & .733\,/\,.659 & .729\,/\,.639 & .734\,/\,.570 & .732\,/\,.663 & \colorbox{tier2}{.817}\,/\,\colorbox{tier1}{\textbf{.751}} & \colorbox{tier1}{\textbf{.856}}\,/\,\colorbox{tier2}{.721} & .657\,/\,.442 & \colorbox{tier3}{.775}\,/\,\colorbox{tier3}{.668} \\
Gemini$\times$\textsc{OSG} & .654\,/\,.522 & .653\,/\,.523 & .697\,/\,.497 & .659\,/\,.516 & \colorbox{tier1}{\textbf{.880}}\,/\,\colorbox{tier1}{\textbf{.946}} & \colorbox{tier3}{.806}\,/\,\colorbox{tier3}{.608} & .651\,/\,.521 & \colorbox{tier2}{.821}\,/\,\colorbox{tier2}{.798} \\
Gemini$\times$\textsc{UIV} & \colorbox{tier2}{.782}\,/\,\colorbox{tier2}{.647} & \colorbox{tier3}{.766}\,/\,.625 & .710\,/\,.485 & \colorbox{tier1}{\textbf{.791}}\,/\,\colorbox{tier1}{\textbf{.687}} & .636\,/\,.365 & .740\,/\,\colorbox{tier3}{.626} & .703\,/\,.500 & .699\,/\,.475 \\
\bottomrule
\end{tabular}%
}
}

\end{table}

\section{UQ Generalizes Selectively Across Regimes}
\label{sec:transfer}

Table~\ref{tab:headline_family_auroc} summarizes AUROC$_{\mathrm{incorrect}}$ by UQ family. Panel A reports, for each open-weight regime, the best method within each family across the full $27$-method matrix; full method-level AUROC/PRR results and per-cell top-$10$ lists are in Appendices~\ref{appx:full_open_matrix}, \ref{appx:full_open_prr_matrix}, and \ref{appx:per_cell_top10}. No family dominates all regimes. Density/probe methods are the most stable open-weight family, with SEP or SAPLMA leading on most cells; CoCoA-1MCA and Focus are the strongest non-density alternatives in specific regimes such as Q72$\times$\textsc{OSG} and PT$\times$\textsc{OSG}. Thus, the headline result is a regime-dependent pattern of reliable families, not a universal UQ score. 

\begin{figure}[!htbp]
\centering
\includegraphics[width=0.45\textwidth]{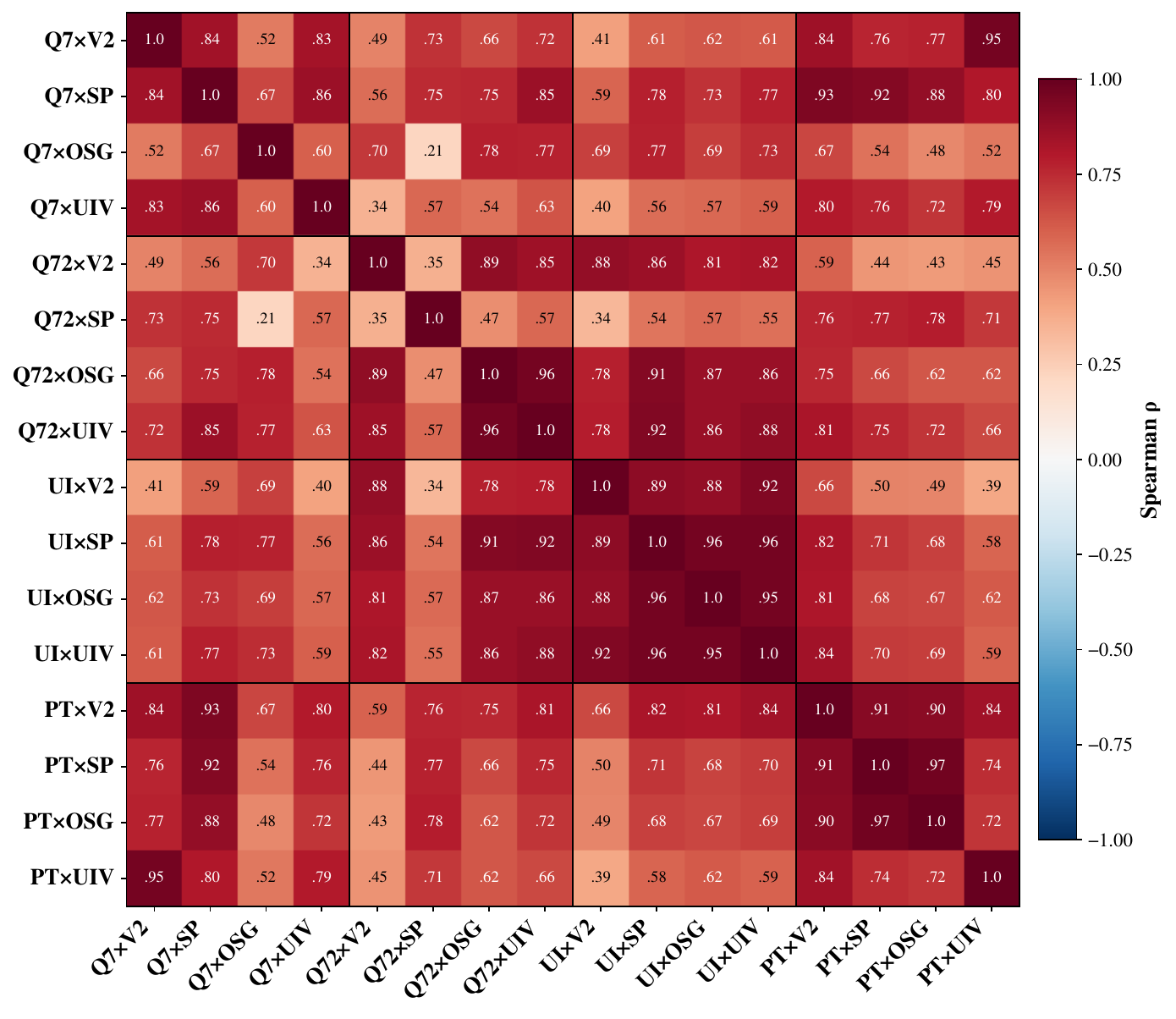}\hfill
\includegraphics[width=0.45\textwidth]{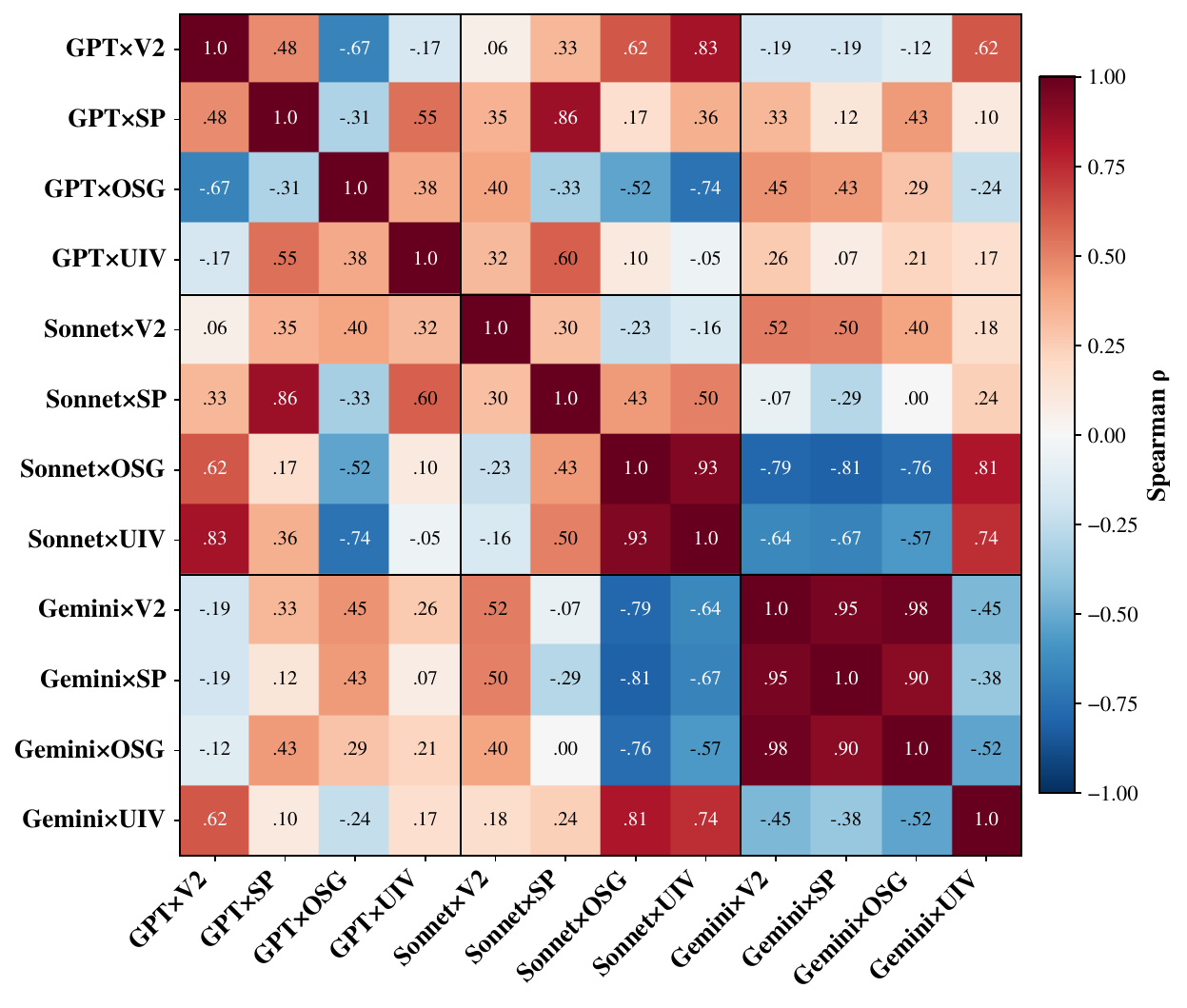}
\caption{\textbf{Full-method ranking transfer.}
Spearman $\rho$ between per-cell rankings ($50$-seed means). Left: $27$-method open-weight, $16$ cells. Right: $8$-method API-only, $12$ cells. Diagonal blocks: within-model or within-vendor transfer.}
\label{fig:rho_full}
\end{figure}

Panel B shows the API-only setting, where logits, hidden states, and attention traces are unavailable. On \textsc{ScreenSpot-Pro}, CCP leads on OpenAI and Anthropic, while Verbalised-2S leads on Gemini (AUROC $0.856$); on \textsc{ScreenSpot-v2}, Verbalised-1S reaches AUROC $0.966$ on Gemini (the score distribution is sharply bimodal at $\{0,1\}$; see Appendix~\ref{appx:verbalised_bimodality}). This interface shift changes which UQ signals are usable and effective. To quantify portability, we rank UQ methods by AUROC$_{\mathrm{incorrect}}$ within each cell and compute Spearman $\rho$ between rankings. Figure~\ref{fig:rho_full} shows the $16\times16$ open-weight matrix and the $12\times12$ API-only matrix. Open-weight rankings are positively correlated across all $120$ pairs (mean $\rho=0.705$, max $\rho=0.969$), but API-only transfer is weaker and vendor-specific. Importantly, the transfer analysis evaluates \emph{method choice}, not raw score calibration. Each cell may have different accuracy, score scale, and error geometry, so we compare rankings of UQ methods rather than score values.

\noindent\textbf{Transfer is strongest across datasets at fixed model.}
The largest off-diagonal correlations are POINTS-GUI-G cross-dataset pairs: PT$\times$\textsc{OSG} to PT$\times$\textsc{SP} gives $\rho=0.969$, PT$\times$\textsc{OSG} to PT$\times$\textsc{V2} gives $\rho=0.903$, and PT$\times$\textsc{SP} to PT$\times$\textsc{V2} gives $\rho=0.908$. Thus, for a fixed model, UQ rankings remain relatively stable across dataset difficulty and target geometry shifts.

\noindent\textbf{Transfer weakens across models and reasoning-heavy regimes.}
On \textsc{OSWorld-G}, cross-model correlations remain positive but are lower than the strongest fixed-model pairs: Q72$\times$\textsc{OSG} to UI$\times$\textsc{OSG} gives $\rho=0.866$, UI$\times$\textsc{OSG} to Q7$\times$\textsc{OSG} gives $\rho=0.693$, and Q7$\times$\textsc{OSG} to Q72$\times$\textsc{OSG} gives $\rho=0.779$. The weakest correlations involve \textsc{UI-Vision-EG}, indicating that reasoning-heavy element grounding can reduce transfer below the usual fixed-model dataset-shift baseline. These cases require target-regime validation rather than direct reuse.

\noindent\textbf{Open-weight to API-only transfer is statistically indistinguishable from zero.}
Removing internal observability eliminates the logit, density/probe, and attention families from API-only comparison, leaving an $8$-method intersection from sampling, hybrid, verbalised, and VLM-native families. On this shared intersection, transfer from Q72-AWQ to the closed-source vendors averages $\rho=+0.08$ over $12$ vendor$\times$dataset pairs (range $-0.76$ to $+0.88$; bootstrap $95\%$ CI $[-0.219, +0.373]$ includes zero, full values in Appendix~\ref{appx:cross_tier}). Within the closed-source matrix, ranking transfer is also weak (mean $\rho=+0.127$ over $66$ pairs). Thus, on the available $12$ pairs we cannot detect meaningful cross-tier transfer, and closed-source UQ recommendations should be reranked on the target calibration split rather than extrapolated from open-weight proxies.

\noindent\textbf{The winning family is regime-dependent.}
Holding \textsc{OSWorld-G} fixed, density wins on all four models (SEP on Q7, Q72, and UI; Mahal-RMD on PT), but the strongest non-density alternative changes: CoCoA-1MCA (hybrid) on Q72 and Focus (attention) on PT. Holding POINTS-GUI-G fixed, density wins on all four datasets, but the within-density top method changes: Mahal-RMD on \textsc{OSWorld-G}, SEP on \textsc{ScreenSpot-v2}, and SAPLMA on \textsc{ScreenSpot-Pro} and \textsc{UI-Vision-EG}. In the API-only matrix, CCP wins $6$ of $12$ cells, while Verbalised-1S/2S win the remaining cells across vendor/dataset combinations. The deployable object is therefore a small regime-aware UQ panel, not a universal single method. Appendix~\ref{appx:ood} gives an OOD framing of the dataset, model, and interface transfer axes.

\section{Graded Error and Calibration}
\label{sec:graded-calibration}

Binary correctness asks whether the click is wrong; graded severity asks how wrong it is. We evaluate severity ranking with miss-only AUSE using target $\varepsilon_i=\log(1+d_{\mathrm{norm},i})$, where $d_{\mathrm{norm},i}=\lVert \mathbf{c}_i-\mathbf{c}^{\star}_i\rVert_2/\sqrt{w_i h_i}$ normalizes click-center error by target-box scale.

\noindent\textbf{Binary detection and spatial severity select different scores.}
AUROC$_{\mathrm{incorrect}}$ and miss-only AUSE select the same top method on only $2$ of $16$ open-weight cells, but on $9$ of $12$ API-only cells. Thus, in the richer $27$-method open-weight panel, error detection and graded miss-severity ranking often prefer different UQ signals; in the smaller $8$-method API-compatible panel, there are fewer ways for the objectives to diverge. The full per-cell winner table is in Appendix~\ref{appx:ause_full}. AUSE on $\log(1+d_{\mathrm{norm}})$ is the severity-aware metric used here; further percentile stratification by $d_{\mathrm{norm}}$ is often degenerate because correctness is close to thresholding normalized distance.

\noindent\textbf{Discrimination, rejection, and calibration remain separate objectives.}
Density/probe methods often rank errors well, but their heavy-tailed scores can be harder to calibrate on small calibration splits. On low-accuracy \textsc{OSWorld-G} cells, ECE$^{\mathrm{iso}}$ for density methods exceeds $0.40$ in several cases, while token-probability baselines show smaller raw-to-isotonic calibration gaps. PRR$_{0.5}^{\mathrm{norm}}$ is also accuracy-dependent: on high-accuracy PT$\times$\textsc{V2}, it approaches a structural ceiling, so PRR is most interpretable within comparable-accuracy regimes. \textit{Overall}, there is no single scalar notion of ``good uncertainty'' for executable GUI actions. AUROC measures error detection, AUSE measures severity ranking, and calibration measures risk interpretability. UQ should therefore be selected for intervention: reject, rerank, or defer.

\section{Model-Class Transitions and Interface Effects}
\label{sec:tuning}

The transfer analysis shows that UQ rankings change across model classes and observable interfaces. We examine two mechanisms: vanilla-to-specialist transitions and API-only observability. Moving from vanilla VLMs (Q7, Q72) to grounding-specialist GUI agents (UI, PT) shifts family-level reliability. Appendix~\ref{appx:tuning_full} reports the full transition table, including pooled vanilla-to-specialist $\Delta$AUROC across all four datasets and a controlled \textsc{ScreenSpot-Pro} decomposition separating scale-only change (Q7$\to$Q72), fixed-backbone grounding fine tuning (Q7$\to$UI), tune-plus-backbone change (Q7$\to$PT), and the mixed Q72$\to$PT transition.

\noindent\textbf{Model transitions alter family reliability.}
Across vanilla-to-specialist transitions, attention, verbalised, and VLM-native families lose AUROC on all four datasets, with mean $\Delta$AUROC of $-0.015$, $-0.037$, and $-0.029$, respectively. Density/probe methods are the most stable family, with mean $\Delta$AUROC $=+0.008$ and only a small loss on \textsc{UI-Vision-EG}. Logit, sampling, and hybrid families show dataset-dependent behavior rather than a uniform trend.

\noindent\textbf{Scale and backbone changes confound fine-tuning effects.}
The \textsc{ScreenSpot-Pro} decomposition in Appendix~\ref{appx:tuning_full} shows that some losses attributed to grounding fine tuning are partly explained by scale, quantization, or backbone changes. The Q7$\to$Q72 scale-only transition produces larger losses for sampling ($-0.277$) and hybrid ($-0.257$) than the Q7$\to$UI fixed-backbone fine-tune transition ($-0.039$ and $-0.010$). For logit methods, scale ($-0.095$) and fine tuning ($-0.079$) are comparable. Density/probe methods remain nearly flat across all four transitions ($|\Delta|\le 0.018$), making them the most transition-stable family in this analysis.

\noindent\textbf{API-only observability changes the usable UQ panel.}
Closed-source APIs remove logits, hidden states, and attention maps, leaving an $8$-method intersection from the sampling, hybrid, verbalised, and VLM-native families. On this intersection, Verbalised-1S and Verbalised-2S are top-1 on $6$ of $12$ closed-source cells, including Verbalised-1S on Gemini$\times$\textsc{ScreenSpot-v2} (AUROC $0.966$) and Gemini$\times$\textsc{OSWorld-G} (AUROC $0.880$), and Verbalised-2S on Gemini$\times$\textsc{ScreenSpot-Pro} (AUROC $0.856$). CCP wins the other $6$ cells, mainly on Anthropic and OpenAI. Thus, API-only observability removes the open-weight dominant density/attention families and promotes response-level signals. The full closed-source AUROC matrix is in Table~\ref{tab:headline_family_auroc} Panel B and Appendix~\ref{appx:full_closed_matrix}.

\begin{figure}[!htbp]
\centering
\begin{subfigure}[t]{0.325\textwidth}
\centering
\includegraphics[width=\linewidth]{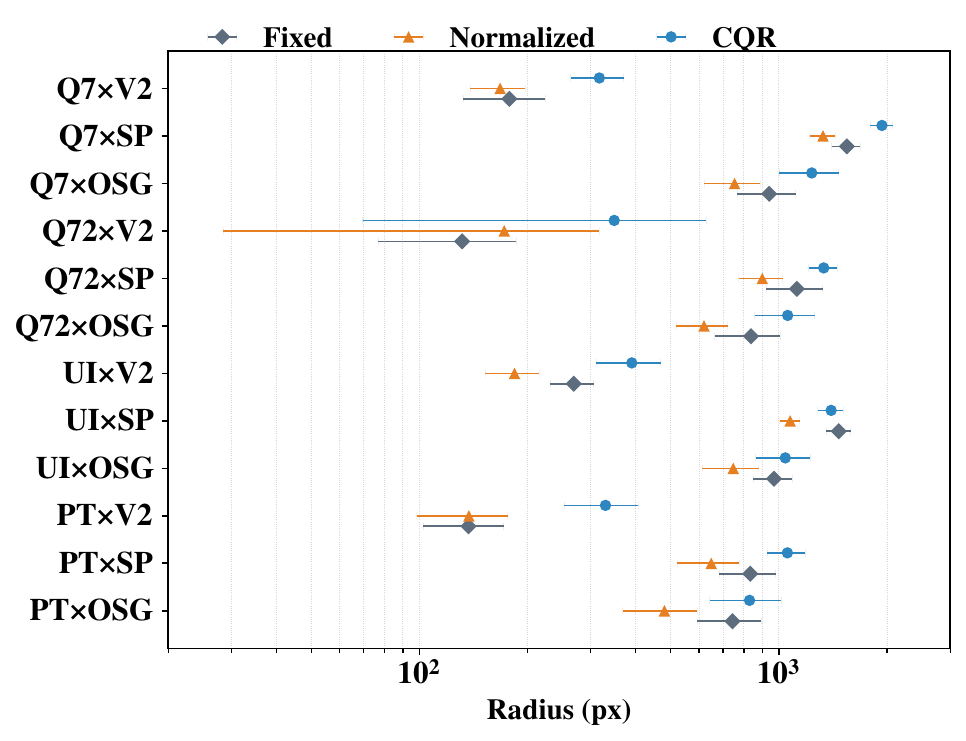}
\caption{Open-weight: radius}
\end{subfigure}\hfill
\begin{subfigure}[t]{0.325\textwidth}
\centering
\includegraphics[width=\linewidth]{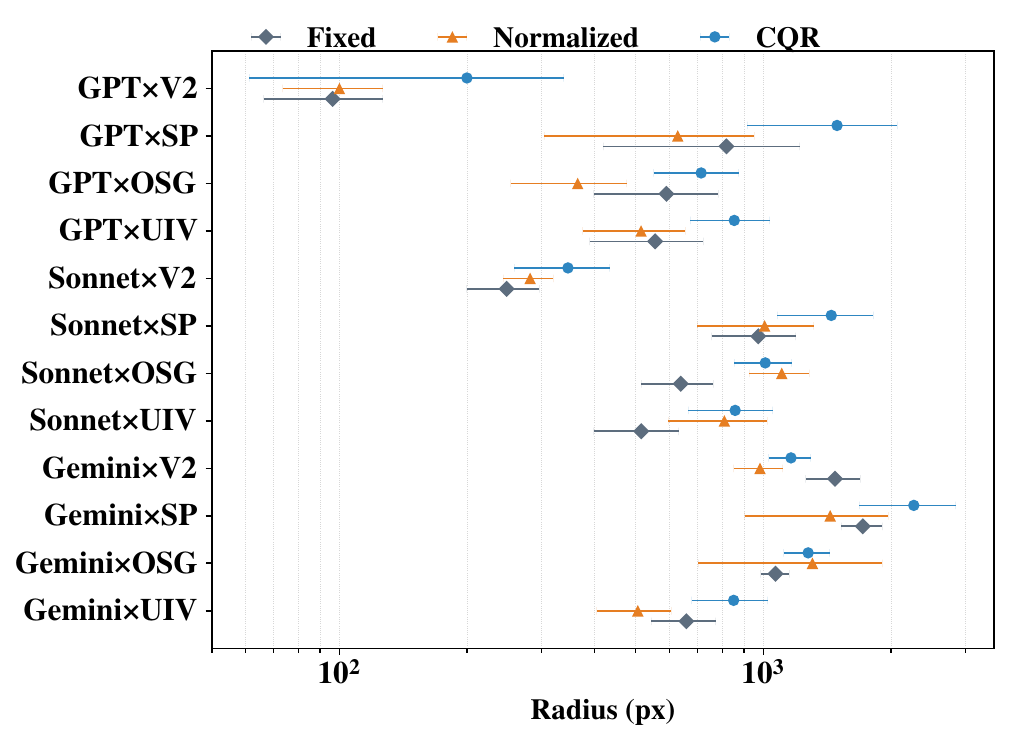}
\caption{Closed-source: radius}
\end{subfigure}\hfill
\begin{subfigure}[t]{0.325\textwidth}
\centering
\includegraphics[width=\linewidth]{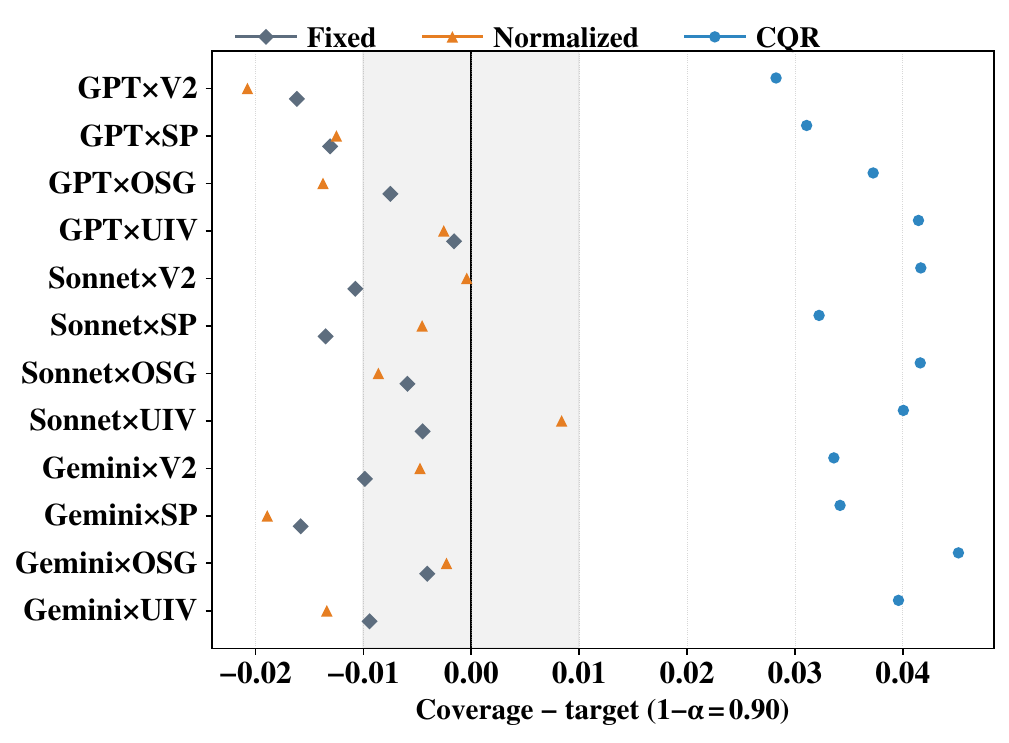}
\caption{Closed-source: cov.\ gap}
\end{subfigure}
\caption{\textbf{Adaptive conformal click-disks at $\alpha=0.10$.}
$50$-seed mean radius and coverage gap (target $0.90$). Panels (a) cover $12$ open-weight cells; (b, c) cover $12$ API-only cells. Gray band: $\pm1$ pp around target. Variants defined in $\S$\ref{sec:benchmark}; multi-$\alpha$ and $16$-cell results in Appendix~\ref{appx:crc}.}
\label{fig:conformal_v7}
\end{figure}

\section{Inductive-Conformal Click Disks}
\label{sec:conformal}

Previous sections evaluate UQ scores as error-ranking functions. We next test whether they yield deployable spatial regions. Figure~\ref{fig:conformal_v7} reports split-conformal click disks at $\alpha=0.10$ using fixed-radius, normalized, and CQR variants. Panels~(a) show $12$ open-weight cells (4 agents $\times$ \textsc{V2}/\textsc{SP}/\textsc{OSG}; full $16$-cell results in Appendix~\ref{appx:crc}); panels~(b,c) show the $12$ API-only cells. Coverage means that the disk centered at the predicted click contains the target-box center.

\noindent\textbf{Radius--coverage tradeoff.}
Disk-Normalized gives the smallest radius on most open-weight cells, e.g., PT$\times$\textsc{OSG} ($743\!\to\!481$ px, $-35\%$), Q$72$$\times$\textsc{OSG} ($837\!\to\!620$ px, $-26\%$), UI$\times$\textsc{SP} ($1469\!\to\!1076$ px, $-27\%$), and Q$72$$\times$\textsc{SP} ($1124\!\to\!900$ px, $-20\%$). The reduction depends on the plug-in UQ score; density/probe scores produce the largest gains, while less stable plug-ins shrink less. Disk-CQR is more conservative, over-covering by roughly $3$--$5$ pp at $\alpha=0.10$ with larger radii. Disk-Fixed and Disk-Normalized remain within $\pm1$ pp of target coverage; full results for various $\alpha$ are in Appendix~\ref{appx:crc}.

\noindent\textbf{Interface effects.}
Closed-source coverage depends on vendor behavior. On Gemini$\times$\textsc{SS-Pro}, Disk-Fixed coverage is $0.926$, $0.884$, and $0.795$ for targets $0.95$, $0.90$, and $0.80$, respectively, indicating small but systematic undercoverage. Since split conformal is marginally valid under exchangeability, this suggests calibration-test mismatch across API strata and motivates target-split coverage checks.

\noindent\textbf{Task difficulty.}
Disk radius tracks benchmark difficulty. At $\alpha=0.10$, \textsc{SS-Pro} requires large open-weight Disk-Fixed radii ($833$--$1547$ px), while \textsc{SS-V2} requires much smaller radii ($132$--$269$ px); \textsc{OSWorld-G} falls between these regimes. Conformal Risk Control results are in Appendix~\ref{appx:crc}. \textit{Overall}, score-level UQ is insufficient for deployment. A score can rank errors well by AUROC yet produce regions that are too large, under-covered, or depend on the output interface. Executable GUI agents should therefore be evaluated both by risk-ranking quality and by spatial coverage.

\begin{table}[!htbp]
\centering
\caption{\textbf{Regime-aware UQ selection recipe.}
Protocol for selecting a UQ panel for executable GUI click grounding. The panel is a prior; final selection must be validated on the target calibration split.}
\label{tab:deployment_recipe}
\footnotesize
\setlength{\tabcolsep}{3.5pt}
\renewcommand{\arraystretch}{1.08}
\begin{tabular}{@{}>{\columncolor{axisbg}}p{0.02\linewidth}p{0.1\linewidth}p{0.45\linewidth}p{0.35\linewidth}@{}}
\toprule
\textbf{\#} & \textbf{Decision} & \textbf{Action} & \textbf{Evidence} \\
\midrule

\textbf{1} &
Interface &
If hidden states are available, include \textbf{SAPLMA / SEP} and \textbf{Mahal-RMD}. If API-only, start with the harmonised $8$-method panel: \textbf{CCP}, \textbf{SelfCons}, \textbf{SE}, \textbf{LexSim}, \textbf{Verb-1S}, \textbf{Verb-2S}, \textbf{HEDGE}, and \textbf{IMGHEDGE}. &
Density/probe is the most stable open-weight family; API-only regimes promote response-level and verbalised scores. \\

\textbf{2} &
Model class &
For specialist GUI agents, avoid logit and lexical-overlap scores as primary signals on \textsc{SP} / \textsc{OSG} / \textsc{UIV} (\textsc{V2} is the exception). Add \textbf{CoCoA-1MCA}; \textbf{HEDGE / IMGHEDGE} help on Qwen2.5 fine-tunes but degrade across backbone changes. &
Vanilla $\to$ specialist transitions shift family preferences (attention / verbalised / VLM-native uniformly down, density stable). \\

\textbf{3} &
Objective &
For rejection, prioritize AUROC / PRR. For severity, validate AUSE. For calibrated risk, check ECE / Brier after isotonic calibration. &
AUROC and AUSE winners disagree on $14$ of $16$ open-weight cells (Table~\ref{tab:ause_top}); calibration and discrimination select different scores. \\

\textbf{4} &
Reuse &
Reuse a prior panel across datasets only when the model is fixed. If model family or interface changes, rerank on the target calibration split. &
Ranking transfer is strongest for fixed-model dataset shifts, reaching $\rho=0.969$ on PT$\times\{$OSG, SP$\}$, and weaker across model/interface changes. \\

\textbf{5} &
Click region &
For spatial coverage, start with \textbf{Disk-Fixed} or \textbf{Disk-CQR}. Use \textbf{Disk-Normalized} only after coverage checks pass. &
Disk-Normalized can shrink radii by $40$ to $60\%$, but can under-cover under mismatch. \\

\bottomrule
\end{tabular}
\end{table}

\section{Practical UQ Panel Selection and Conclusions}
\label{sec:conclusion}

Our main empirical finding is selective generalization of UQ methods for computer-use agents. Within the open-weight matrix, UQ rankings are most stable across datasets for a fixed model class, with mean cross-cell Spearman $\rho=0.705$ over $120$ pairs and maximum $\rho=0.969$. On the shared $8$-method open$\leftrightarrow$closed intersection, mean cross-tier transfer is $\rho=+0.08$ over $12$ vendor$\times$dataset pairs with bootstrap $95\%$ CI $[-0.219, +0.373]$ that includes zero, so cross-tier transfer is statistically indistinguishable from no-transfer at the available sample size. The deployment implication is that UQ selection should be regime-specific. Density/probe methods (SAPLMA, SEP, Mahal-RMD) are the most stable open-weight family; CoCoA-1MCA and Focus are strongest in specific open-weight regimes; CCP and Verbalised-1S/2S are competitive in API-only regimes. Model-class transitions also change family-level reliability: attention, verbalised, and VLM-native families lose AUROC from vanilla to specialist models across all four datasets, while density/probe methods remain comparatively stable. Conformal click-disks provide a spatial counterpart to this result: Disk-Normalized can reduce radii by $40$--$60\%$ when the plug-in UQ score is calibrated, but coverage must still be checked under interface or calibration-test mismatch. Table~\ref{tab:deployment_recipe} summarizes these observations as a panel-selection procedure. The recommended panels are calibration-efficient priors; the selected panel should be reranked and coverage-checked on the target split. Overall, uncertainty quality is not a property of a UQ method alone; it depends jointly on the method, model, dataset geometry, observable interface, and deployment objective.

\noindent\textbf{Release and limitations.}
We release \texttt{argus-uq}, including $27$ method implementations, per-item records, splits, UQ scores, closed-source API responses, and analysis scripts. Package details, release manifest, compute setup, and ethics statement are provided in Appendices~\ref{appx:argus_uq}, \ref{appx:datasheet}, and \ref{appx:compute}. This study isolates single-step executable clicks (not multi-step trajectories), uses $n_{\text{samples}}=5$, and does not propose a new UQ estimator (SafeGround scores in Appendix~\ref{appx:safeground}).

\FloatBarrier
\bibliographystyle{plainnat}
\bibliography{references}

\appendix
\renewcommand{\thesection}{A\arabic{section}}
\setcounter{section}{0}

\section{Benchmark overview figure}
\label{appx:overview}

\begin{figure}[!htbp]
\centering
\includegraphics[width=\textwidth]{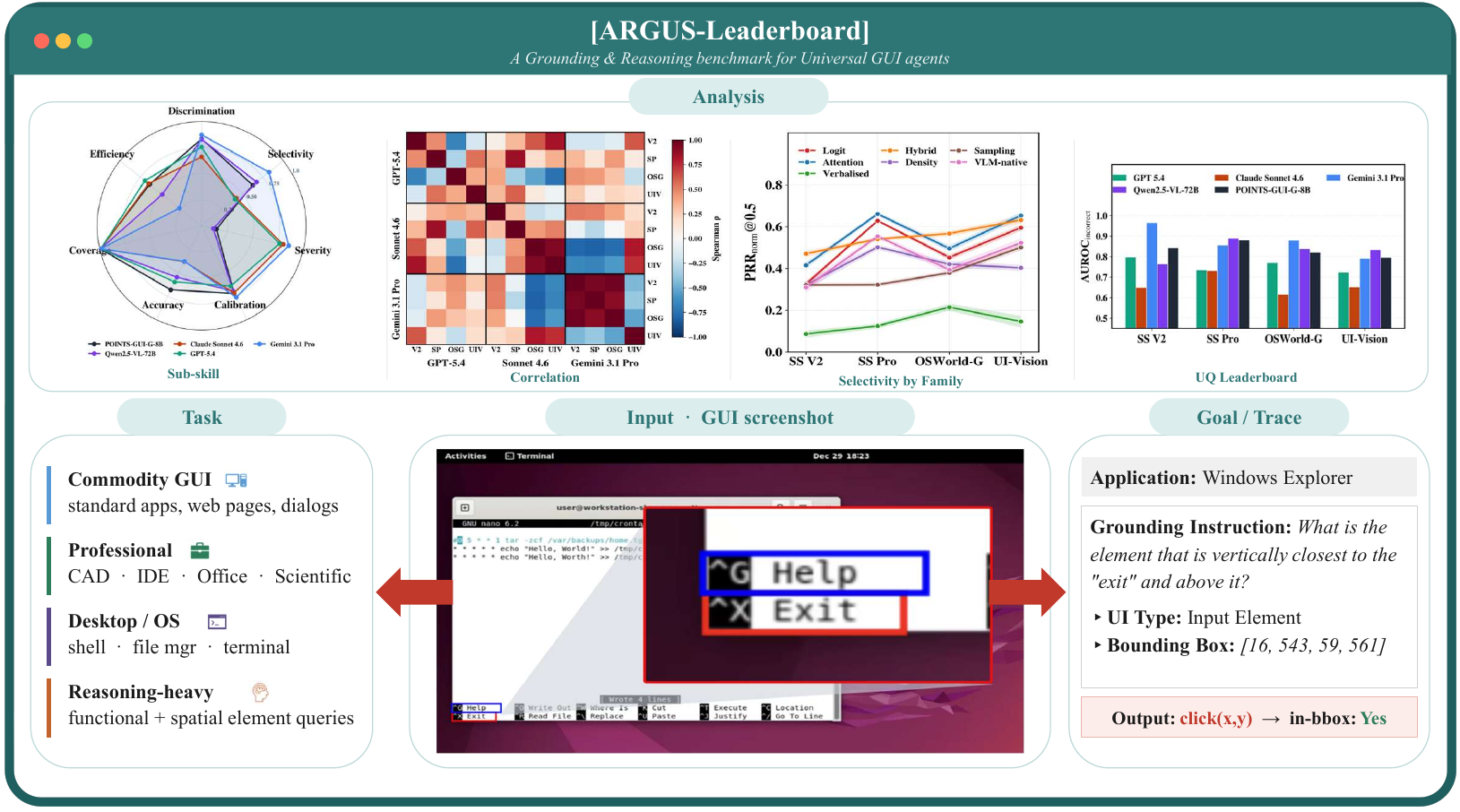}
\caption{\textbf{Cross-regime benchmark of post-hoc UQ for single-step executable GUI grounding.} Each click prediction is paired with $27$ open-weight UQ scores ($8$ on the API-only panel) and evaluated across $4$ open-weight agents, $3$ frontier closed-source vendors, and $4$ grounding datasets, supporting error discrimination, selective execution, calibration, graded miss-severity, ranking transfer, and conformal click-disk coverage.}
\label{fig:overview}
\end{figure}

\section{Extended Related Work: Uncertainty Quantification, Conformal Prediction, and Risk-Aware Deployment}
\label{appx:extended_related}

Beyond the post-hoc UQ estimators evaluated in the main paper, a broader line of work makes uncertainty actionable through conformal prediction, calibration, and risk-aware deployment. Trajectory-level risk aggregation extends abstention from single predictions to agentic reasoning~\citep{tayebati2026tracer}. Learnable conformal prediction with context-aware nonconformity functions tightens prediction regions for planning and perception~\citep{kumar2025learnablecp}, while conformal inference combined with evidential learning separates epistemic from aleatoric uncertainty~\citep{stutts2024conformal}. A related thread decomposes and routes uncertainty by type: calibrated aleatoric and epistemic decomposition in deep features supports inference-time adaptation~\citep{kumar2025calibrated}, and type-routed interventions act on the dominant uncertainty source~\citep{kumar2026triage}. Uncertainty-aware sensor fusion applies conformal prediction with principled abstention to safety-critical autonomy~\citep{kumar2025lidarcp}, and uncertainty also guides inference-time depth adaptation in visual tracking~\citep{poggi2026depth}. These deployment-oriented uses of uncertainty motivate our focus on whether post-hoc UQ rankings, not just individual scores, transfer across GUI agents, datasets, and observable interfaces.

\section{Method-by-method documentation}
\label{appx:methods}

This appendix gives a one-paragraph implementation note per UQ method, citing the canonical paper and the reference implementation tracked.

\paragraph{Logit family.} \emph{MSP} (maximum softmax probability) and \emph{Perplexity} use the greedy-decode token logits; \emph{MeanTokenEntropy} averages per-token softmax entropies; \emph{SequenceProbability} is the sum of token log-probabilities (no length normalisation); \emph{Perplexity-exp} is $\exp(\text{Perplexity})$ retained as a rank-invariant variant for monotonic-transform robustness checks.

\paragraph{Sampling family.} \emph{Self-Consistency} \citep{wang2023selfconsistency} counts unique click clusters across $n=5$ samples under a $50$-px tolerance. \emph{Semantic Entropy} (canonical \citep{kuhn2023semantic} and weighted \citep{farquhar2024semantic}) groups samples by tolerance-clustering and entropies the cluster mass; the weighted form scales clusters by sample sequence-probability. \emph{MCSE} / \emph{MCNSE} are Monte-Carlo (length-normalised) sequence-entropy variants \citep{malinin2018predictive}. \emph{LexicalSimilarity} averages pairwise $1 - \text{ratio}$ over samples.

\paragraph{Hybrid family.} \emph{CoCoA-canonical} \citep{vashurin2025cocoa}: $U_{\text{inf}} \cdot U_{\text{cons}}$ = (sum log-prob of greedy) $\times$ (fraction of samples within tolerance of greedy). \emph{CoCoA-1MCA}: legacy $1 - C\cdot A$ form retained for back-compat, the saturation-resistant variant. \emph{CCP} \citep{fadeeva2024lmpolygraph}: fraction of samples within tolerance of greedy click.

\paragraph{Attention family.} \emph{Focus} \citep{zhang2023focus}: max-pool per-head attention over all layers and heads, propagate per-token NLL through the resulting attention-weighted accumulated penalty (Eqs.~4 to 6 of the original; $\alpha=0.9$). \emph{RAUQ} \citep{vazhentsev2023rauq}: head selection per layer is the unsupervised argmax over heads of mean attention to the previous generated token; the recurrent confidence is aggregated over the middle-third layers ($\alpha=0.5$). \emph{UQAC} \citep{uqac2025}: per-layer top-$K$ low-entropy heads, max-pool over those heads, propagate NLL through the resulting attention chain ($K=5$, $\theta=0.7$). The attention-tensor capture path (a side-channel hook on the model's SDPA module that does not modify the forward output) is documented in Appendix~\ref{appx:attention_capture}. \emph{Attention Rollout, LLM-Check, Attention-Entropy, Lookback-Lens-unsupervised} are the four other attention-family methods scored to completeness on the open-weight panel; only Focus / RAUQ / UQAC reach top-tier AUROC.

\paragraph{Density family.} \emph{Mahalanobis} \citep{lee2018mahalanobis}: $(\mathbf{h} - \boldsymbol\mu_+)^\top \Sigma_+^{-1} (\mathbf{h} - \boldsymbol\mu_+)$ on mean-pooled last-layer hidden state. \emph{Mahalanobis-RMD} \citep{ren2023rmd}: subtracts background-class distance term; the relative-Mahalanobis lift over the naive form replicates on every legacy reference cell (Appendix~\ref{appx:rmd}). \emph{Mahalanobis-RDE}: KernelPCA(100) $\to$ MinCovDet robust covariance. \emph{SAPLMA} \citep{azaria2023saplma}: $256\!-\!128\!-\!64$ MLP probe on the same hidden state. \emph{SEP} \citep{kossen2024sep}: logistic regression on hidden state.

\paragraph{Verbalised family.} \emph{P(True)} \citep{kadavath2022pktrue}: probability of the token ``True'' when shown the model's own answer. \emph{Verbalised-1S} \citep{tian2023verbalised1s}: one-stage parsed numeric confidence. \emph{Verbalised-2S} \citep{xiong2024verbalised2s}: two-stage protocol.

\paragraph{VLM-native family.} \emph{HEDGE}: $n_p = 3$ instruction paraphrases, cluster-entropy over resulting clicks. \emph{IMGHEDGE}: $3$ image perturbations (saturation $+30$\%, Gaussian noise $\sigma=8$, JPEG q=$50$); these were the AUROC-best three of an $8$-perturbation pilot \citep{cubuk2019autoaugment} (Appendix~\ref{appx:imghedge_pilot}).

\begin{table}[!htbp]
\centering
\caption{\textbf{Metric card.} Deployment question measured by each metric.}
\label{tab:metric_card}
\footnotesize
\setlength{\tabcolsep}{2.5pt}
\renewcommand{\arraystretch}{1.12}
\begin{tabular}{@{}>{\columncolor{axisbg}}p{0.19\linewidth}p{0.18\linewidth}p{0.59\linewidth}@{}}
\toprule
\textbf{Requirement} & \textbf{Metric} & \textbf{Definition / interpretation} \\
\midrule

\textbf{Error detection} &
\textbf{AUROC$_{\mathrm{incorrect}}$} &
ROC-AUC with incorrect clicks as positives. Measures whether the score ranks incorrect clicks above correct clicks. \\

\textbf{Selective execution} &
\textbf{PRR$_{0.5}^{\mathrm{norm}}$}; \textbf{AURC} &
PRR$_{0.5}^{\mathrm{norm}}$ measures oracle-normalized rejection quality up to $50\%$ rejection~\citep{malinin2018predictive}. AURC integrates risk over accepted coverage~\citep{geifman2019selectivenet}; lower is better. \\

\textbf{Calibrated risk} &
\textbf{ECE$^{\mathrm{iso}}$}; \textbf{Brier$^{\mathrm{iso}}$} &
Risk scores are mapped to probabilities by isotonic regression fit on the calibration split and evaluated on test~\citep{guo2017calibration}. Heavy-tailed scores use a QuantileTransformer before isotonic regression. \\

\textbf{Graded severity} &
\textbf{AUSE} &
Miss-only sparsification error~\citep{ilg2018ause} with $\varepsilon_i=\log(1+d_{\mathrm{norm},i})$. Lower is better; tests whether UQ ranks wrong clicks by miss severity. \\

\textbf{Ranking transfer} &
\textbf{Spearman $\rho$} &
Spearman rank correlation between two cells' per-method AUROC vectors. Tests whether UQ-method recommendations port across regimes (datasets, model classes, observable interfaces); higher is better. \\

\textbf{Spatial coverage} &
\textbf{Conformal click-disks} &
Fixed-radius split conformal baseline plus Disk-Normalized~\citep{lei2018distfree} and Disk-CQR~\citep{romano2019cqr} variants. Reports empirical coverage of the target-box center and disk radius. \\

\bottomrule
\end{tabular}
\end{table}

\section{Out-of-distribution analysis}
\label{appx:ood}

We treat the cross-regime axes already evaluated in the main paper as an out-of-distribution (OOD) analysis under the deployment-relevant definition: the calibration distribution observed during UQ-method selection differs from the test distribution where the score is used. Three OOD shifts are present in the benchmark and each carries a quantitative estimate of how much UQ rankings degrade.

\noindent\textbf{Dataset shift (commodity $\to$ professional / reasoning-heavy).}
Holding the model fixed and varying the dataset across \textsc{ScreenSpot-v2}, \textsc{ScreenSpot-Pro}, \textsc{OSWorld-G}, and \textsc{UI-Vision-EG} measures dataset-axis OOD transfer. Mean Spearman $\rho=0.79$ over the $24$ same-model cross-dataset pairs across the four open-source models (computed from the off-diagonal within-model blocks of Figure~\ref{fig:rho_full} left panel). This is the strongest OOD axis we measure; UQ rankings remain mostly stable when only the dataset changes.

\noindent\textbf{Model shift (architecture, scale, fine-tuning).}
Holding the dataset fixed and varying the model class across Qwen2.5-VL-7B, Qwen2.5-VL-72B-AWQ, UI-TARS-1.5-7B, and POINTS-GUI-G-8B measures model-axis OOD transfer. Mean Spearman $\rho=0.69$ over the $24$ same-dataset cross-model pairs. Model-axis transfer is weaker than dataset-axis but remains positive and frequently significant; the practical message is that UQ-method recommendations should be revalidated when changing model class.

\noindent\textbf{Interface shift (open-weight $\to$ closed-source API).}
Removing internal model signals (logits, hidden states, attention maps) measures interface-axis OOD transfer. Mean cross-tier Spearman $\rho=+0.08$ over $12$ pairs on the $8$-method open$\leftrightarrow$closed intersection, with $4$ of $12$ pairs significant; range $-0.76$ to $+0.88$ (Appendix~\ref{appx:cross_tier}). Interface-axis transfer is the weakest OOD axis we measure and the only one where rankings can be significantly anti-correlated with the open-source reference; UQ-method selection should be redone on the target observability surface, not extrapolated.

\noindent\textbf{Probe-transfer summary.}
The hidden-state probe methods (SAPLMA, SEP, Mahalanobis-RMD) are trained on the calibration split of one cell and applied to the test split of the same cell within the multi-seed protocol. Cross-cell probe transfer (training on one cell, deploying on another) is not the headline protocol of this paper but is a natural follow-up; the released probe checkpoints under \texttt{data\_release/} support the experiment.

\section{Methods deferred}
\label{appx:deferred}

\paragraph{Beyond the $27$ benchmarked.} Methods deferred under our inclusion criteria: SafeGround spatial-dispersion scores \citep{safeground2026} (need $n\geq 10$ stochastic clicks; we have $n=5+3+3$ but the SafeGround scoring is most reliable on $\geq 10$ pure-stochastic samples; reported in Appendix~\ref{appx:safeground}); Mondrian / size-stratified conformal \citep{vovk2005algorithmic,degrancey2024bbox} (deferred to follow-up); CQR with $\hat\sigma$ regressors \citep{romano2019cqr} (we evaluate split CP at three $\alpha$ levels in the main paper); SafeGround-style learn-then-test thresholding \citep{angelopoulos2021ltt}; Dropout Decoding, Lookback Lens supervised, UHead (heavyweight supervised methods that require a record-format upgrade); per-layer Mahalanobis sweeps (would require all-layer hidden-state caching).

\section{Full method-level open-source AUROC matrix}
\label{appx:full_open_matrix}

Table~\ref{tab:full_open_matrix} is the complete $27$-method $\times$ $16$-cell AUROC$_{\mathrm{incorrect}}$ matrix from the multi-seed $n=50$ pass. Each cell is a $50$-seed mean over $80/20$ stratified calibration/test splits. Tier shading marks the top-$3$ methods per cell.

\begin{table}[!htbp]
\centering
\caption{\textbf{Full open-source AUROC matrix.} 27 methods $\times$ 16 cells; 50-seed mean. Tier-shaded top-3 per cell.}
\label{tab:full_open_matrix}
\scriptsize
\setlength{\tabcolsep}{2pt}
\renewcommand{\arraystretch}{0.95}
\resizebox{\textwidth}{!}{%
\begin{tabular}{@{}l c c c c c c c c c c c c c c c c@{}}
\toprule
Method & \multicolumn{4}{c}{\textsc{Q7}} & \multicolumn{4}{c}{\textsc{Q72}} & \multicolumn{4}{c}{\textsc{UI}} & \multicolumn{4}{c}{\textsc{PT}} \\
\cmidrule(lr){2-5} \cmidrule(lr){6-9} \cmidrule(lr){10-13} \cmidrule(lr){14-17}
 & V2 & SP & OSG & UIV & V2 & SP & OSG & UIV & V2 & SP & OSG & UIV & V2 & SP & OSG & UIV \\
\midrule
\multicolumn{17}{l}{\textit{\textsc{Logit}}} \\
\,\,MSP & .726 & .839 & .744 & .808 & .615 & .722 & .669 & .683 & .715 & .776 & .718 & .753 & .703 & .764 & .720 & .674 \\
\,\,Perplexity & .718 & .837 & .731 & .798 & .617 & .741 & .672 & .687 & .711 & .776 & .715 & .762 & .688 & .764 & .709 & .669 \\
\,\,Ppl-exp & .718 & .837 & .731 & .798 & .617 & .741 & .672 & .687 & .711 & .776 & .715 & .762 & .688 & .764 & .709 & .669 \\
\,\,SeqProb & .724 & \cellcolor{tier3}.873 & .742 & .801 & .664 & .804 & .751 & .779 & .714 & .780 & .713 & .770 & .693 & .766 & .711 & .668 \\
\,\,MTE & .741 & .839 & .758 & \cellcolor{tier1}\textbf{.816} & .623 & .743 & .686 & .698 & .647 & .722 & .668 & .704 & .688 & .780 & .725 & .694 \\
\addlinespace[1pt]
\multicolumn{17}{l}{\textit{\textsc{Sampling}}} \\
\,\,SelfCons & .685 & .795 & .753 & .748 & .730 & .505 & .771 & .737 & .804 & .813 & .763 & .800 & .676 & .621 & .614 & .616 \\
\,\,SE & .691 & .799 & .764 & .753 & .733 & .505 & .774 & .745 & .808 & .814 & .768 & .806 & .676 & .620 & .614 & .617 \\
\,\,SE-w & .694 & .817 & .769 & .780 & .733 & .505 & .776 & .753 & .807 & .813 & .773 & .802 & .622 & .637 & .598 & .618 \\
\,\,MCSE & .723 & .799 & .751 & .783 & .681 & .501 & .726 & .723 & .665 & .734 & .589 & .669 & .583 & .527 & .535 & .665 \\
\,\,MCNSE & .723 & .799 & .751 & .783 & .681 & .501 & .726 & .723 & .665 & .734 & .589 & .669 & .583 & .527 & .535 & .665 \\
\,\,LexSim & .704 & .669 & .735 & .718 & .572 & .498 & .539 & .508 & .557 & .538 & .503 & .553 & .600 & .514 & .516 & .673 \\
\addlinespace[1pt]
\multicolumn{17}{l}{\textit{\textsc{Hybrid}}} \\
\,\,CoCoA & .706 & .865 & \cellcolor{tier3}.782 & .801 & .727 & .508 & .776 & .791 & \cellcolor{tier2}.834 & .837 & .780 & \cellcolor{tier1}\textbf{.825} & .729 & .780 & .724 & .624 \\
\,\,CoCoA-1MCA & .736 & .840 & \cellcolor{tier2}.785 & .785 & \cellcolor{tier2}.751 & .741 & .792 & .773 & \cellcolor{tier1}\textbf{.842} & .836 & .799 & \cellcolor{tier2}.824 & .738 & .748 & .701 & .699 \\
\,\,CCP & .698 & .823 & .779 & .770 & .729 & .508 & .772 & .763 & \cellcolor{tier3}.828 & .825 & .778 & .807 & .684 & .660 & .629 & .624 \\
\addlinespace[1pt]
\multicolumn{17}{l}{\textit{\textsc{Attention}}} \\
\,\,Focus & .760 & .846 & .733 & .777 & \cellcolor{tier3}.736 & .806 & .788 & .779 & .766 & .812 & .757 & .802 & .710 & .777 & .729 & .693 \\
\,\,RAUQ-full & .764 & .835 & .740 & \cellcolor{tier3}.812 & .706 & .744 & .743 & .728 & .771 & .812 & .779 & .816 & .715 & .714 & .715 & .684 \\
\,\,UQAC & .750 & .836 & .731 & .779 & .708 & .792 & .758 & .750 & .745 & .804 & .745 & .788 & .708 & .779 & .728 & .693 \\
\addlinespace[1pt]
\multicolumn{17}{l}{\textit{\textsc{Density}}} \\
\,\,Mahal & .506 & .386 & .476 & .435 & .463 & .464 & .476 & .456 & .552 & .485 & .486 & .447 & .463 & .579 & .581 & .455 \\
\,\,Mahal-RMD & \cellcolor{tier3}.781 & .859 & .762 & .792 & .681 & \cellcolor{tier3}.861 & \cellcolor{tier3}.804 & \cellcolor{tier3}.818 & .735 & \cellcolor{tier3}.861 & \cellcolor{tier2}.863 & \cellcolor{tier3}.818 & \cellcolor{tier3}.765 & \cellcolor{tier3}.854 & \cellcolor{tier1}\textbf{.820} & \cellcolor{tier3}.781 \\
\,\,Mahal-RDE & .557 & .780 & .742 & .649 & .597 & .635 & .467 & .540 & .651 & .638 & .483 & .644 & .615 & .633 & .641 & .386 \\
\,\,SAPLMA & \cellcolor{tier1}\textbf{.817} & \cellcolor{tier2}.883 & .779 & .806 & \cellcolor{tier1}\textbf{.764} & \cellcolor{tier1}\textbf{.889} & \cellcolor{tier2}.823 & \cellcolor{tier1}\textbf{.834} & .762 & \cellcolor{tier1}\textbf{.877} & \cellcolor{tier3}.856 & .804 & \cellcolor{tier2}.843 & \cellcolor{tier1}\textbf{.881} & \cellcolor{tier3}.800 & \cellcolor{tier1}\textbf{.796} \\
\,\,SEP & \cellcolor{tier2}.810 & \cellcolor{tier1}\textbf{.883} & \cellcolor{tier1}\textbf{.786} & \cellcolor{tier2}.815 & .720 & \cellcolor{tier2}.883 & \cellcolor{tier1}\textbf{.838} & \cellcolor{tier2}.834 & .755 & \cellcolor{tier2}.874 & \cellcolor{tier1}\textbf{.863} & .818 & \cellcolor{tier1}\textbf{.843} & \cellcolor{tier2}.874 & \cellcolor{tier2}.805 & \cellcolor{tier2}.786 \\
\addlinespace[1pt]
\multicolumn{17}{l}{\textit{\textsc{Verbalised}}} \\
\,\,P(True) & .668 & .682 & .719 & .704 & .551 & .384 & .674 & .671 & .517 & .552 & .509 & .489 & .535 & .620 & .580 & .501 \\
\,\,Verb-1S & .568 & .587 & .611 & .537 & .633 & .654 & .622 & .563 & .576 & .582 & .635 & .598 & .437 & .481 & .525 & .475 \\
\,\,Verb-2S & .438 & .431 & .531 & .449 & .508 & .538 & .538 & .522 & .601 & .505 & .547 & .574 & .511 & .502 & .524 & .517 \\
\addlinespace[1pt]
\multicolumn{17}{l}{\textit{\textsc{VLM-native}}} \\
\,\,HEDGE & .692 & .723 & .644 & .652 & .674 & .711 & .667 & .672 & .725 & .759 & .716 & .728 & .569 & .602 & .570 & .583 \\
\,\,IMGHEDGE & .641 & .734 & .672 & .693 & .637 & .735 & .677 & .671 & .698 & .767 & .689 & .698 & .598 & .614 & .546 & .568 \\
\addlinespace[1pt]
\bottomrule
\end{tabular}%
}
\end{table}

\section{Full method-level open-source PRR matrix}
\label{appx:full_open_prr_matrix}

Table~\ref{tab:full_open_prr_matrix} reports the per-(method, cell) PRR$_{0.5}^{\mathrm{norm}}$ values from the same $50$-seed multi-seed pass that produces Table~\ref{tab:full_open_matrix}. Tier shading marks the top-$3$ methods per cell on PRR (computed independently of AUROC tier shading).

\begin{table}[!htbp]
\centering
\caption{\textbf{Full open-source PRR matrix.} 27 methods $\times$ 16 cells; 50-seed mean. Tier-shaded top-3 per cell on PRR$_{0.5}^{\mathrm{norm}}$.}
\label{tab:full_open_prr_matrix}
\scriptsize
\setlength{\tabcolsep}{2pt}
\renewcommand{\arraystretch}{0.95}
\resizebox{\textwidth}{!}{%
\begin{tabular}{@{}l c c c c c c c c c c c c c c c c@{}}
\toprule
Method & \multicolumn{4}{c}{\textsc{Q7}} & \multicolumn{4}{c}{\textsc{Q72}} & \multicolumn{4}{c}{\textsc{UI}} & \multicolumn{4}{c}{\textsc{PT}} \\
\cmidrule(lr){2-5} \cmidrule(lr){6-9} \cmidrule(lr){10-13} \cmidrule(lr){14-17}
 & V2 & SP & OSG & UIV & V2 & SP & OSG & UIV & V2 & SP & OSG & UIV & V2 & SP & OSG & UIV \\
\midrule
\multicolumn{17}{l}{\textit{\textsc{Logit}}} \\
\,\,MSP & .395 & \cellcolor{tier2}.878 & .676 & .873 & .190 & .465 & .385 & .584 & .371 & .638 & .399 & .572 & .391 & .497 & .405 & .333 \\
\,\,Perplexity & .376 & .860 & .630 & .870 & .200 & .502 & .367 & .550 & .361 & .651 & .398 & .605 & .354 & .501 & .384 & .320 \\
\,\,Ppl-exp & .376 & .860 & .630 & .870 & .200 & .502 & .367 & .550 & .361 & .651 & .398 & .605 & .354 & .501 & .384 & .320 \\
\,\,SeqProb & .383 & \cellcolor{tier1}\textbf{.879} & .625 & \cellcolor{tier3}.874 & .267 & .627 & .443 & \cellcolor{tier1}\textbf{.743} & .367 & .663 & .393 & .633 & .367 & .505 & .391 & .321 \\
\,\,MTE & .421 & \cellcolor{tier3}.872 & .721 & \cellcolor{tier2}.877 & .200 & .512 & .400 & .612 & .227 & .484 & .259 & .417 & .358 & .524 & .407 & .377 \\
\addlinespace[1pt]
\multicolumn{17}{l}{\textit{\textsc{Sampling}}} \\
\,\,SelfCons & .333 & .677 & .624 & .650 & .448 & .010 & .564 & .561 & .575 & .704 & .557 & .716 & .326 & .204 & .216 & .278 \\
\,\,SE & .348 & .678 & .668 & .658 & .453 & .010 & .571 & .595 & .581 & .703 & .573 & .739 & .326 & .204 & .216 & .280 \\
\,\,SE-w & .354 & .732 & .694 & .781 & \cellcolor{tier3}.454 & .010 & .574 & .639 & .578 & .697 & .585 & .707 & .190 & .233 & .187 & .283 \\
\,\,MCSE & .384 & .757 & .710 & .724 & .315 & $-$.001 & .440 & .586 & .222 & .483 & .070 & .330 & .127 & .044 & .038 & .347 \\
\,\,MCNSE & .384 & .757 & .710 & .724 & .315 & $-$.001 & .440 & .586 & .222 & .483 & .070 & .330 & .127 & .044 & .038 & .347 \\
\,\,LexSim & .327 & .264 & .529 & .596 & .083 & $-$.009 & .060 & .062 & .057 & .011 & $-$.060 & .138 & .178 & .042 & .028 & .369 \\
\addlinespace[1pt]
\multicolumn{17}{l}{\textit{\textsc{Hybrid}}} \\
\,\,CoCoA & .401 & .864 & \cellcolor{tier2}.763 & .816 & .444 & .021 & .589 & \cellcolor{tier2}.742 & \cellcolor{tier1}\textbf{.638} & .783 & .591 & \cellcolor{tier1}\textbf{.793} & .424 & .532 & .420 & .299 \\
\,\,CoCoA-1MCA & .404 & .757 & \cellcolor{tier1}\textbf{.772} & .715 & \cellcolor{tier2}.472 & .502 & .573 & .669 & \cellcolor{tier2}.630 & .772 & .599 & \cellcolor{tier3}.771 & .440 & .467 & .348 & .398 \\
\,\,CCP & .378 & .731 & \cellcolor{tier3}.746 & .702 & .449 & .021 & .576 & .649 & \cellcolor{tier3}.624 & .749 & .586 & .733 & .343 & .292 & .244 & .299 \\
\addlinespace[1pt]
\multicolumn{17}{l}{\textit{\textsc{Attention}}} \\
\,\,Focus & .441 & .822 & .546 & .811 & .399 & .606 & .539 & \cellcolor{tier3}.742 & .466 & .734 & .494 & \cellcolor{tier2}.773 & .424 & .520 & .415 & .353 \\
\,\,RAUQ-full & .451 & .855 & .649 & \cellcolor{tier1}\textbf{.889} & .343 & .573 & .468 & .644 & .479 & .723 & .518 & .756 & .397 & .432 & .377 & .352 \\
\,\,UQAC & .421 & .829 & .563 & .814 & .325 & .600 & .498 & .636 & .428 & .720 & .464 & .719 & .411 & .520 & .415 & .353 \\
\addlinespace[1pt]
\multicolumn{17}{l}{\textit{\textsc{Density}}} \\
\,\,Mahal & $-$.024 & $-$.525 & $-$.145 & $-$.526 & $-$.127 & $-$.167 & $-$.162 & $-$.309 & .049 & $-$.096 & $-$.072 & $-$.338 & $-$.113 & .135 & .136 & $-$.206 \\
\,\,Mahal-RMD & \cellcolor{tier3}.499 & .829 & .634 & .733 & .299 & \cellcolor{tier3}.732 & \cellcolor{tier3}.592 & .705 & .418 & \cellcolor{tier3}.807 & \cellcolor{tier2}.702 & .771 & \cellcolor{tier3}.478 & \cellcolor{tier3}.644 & \cellcolor{tier2}.568 & \cellcolor{tier3}.553 \\
\,\,Mahal-RDE & .068 & .615 & .744 & .652 & .171 & .252 & $-$.095 & .052 & .222 & .270 & $-$.031 & .601 & .172 & .227 & .268 & $-$.273 \\
\,\,SAPLMA & \cellcolor{tier1}\textbf{.599} & .844 & .709 & .835 & \cellcolor{tier1}\textbf{.505} & \cellcolor{tier1}\textbf{.802} & \cellcolor{tier2}.648 & .727 & .483 & \cellcolor{tier2}.809 & \cellcolor{tier3}.678 & .656 & \cellcolor{tier2}.640 & \cellcolor{tier1}\textbf{.702} & \cellcolor{tier3}.563 & \cellcolor{tier1}\textbf{.579} \\
\,\,SEP & \cellcolor{tier2}.589 & .861 & .723 & .841 & .379 & \cellcolor{tier2}.786 & \cellcolor{tier1}\textbf{.662} & .718 & .462 & \cellcolor{tier1}\textbf{.827} & \cellcolor{tier1}\textbf{.709} & .727 & \cellcolor{tier1}\textbf{.657} & \cellcolor{tier2}.690 & \cellcolor{tier1}\textbf{.574} & \cellcolor{tier2}.570 \\
\addlinespace[1pt]
\multicolumn{17}{l}{\textit{\textsc{Verbalised}}} \\
\,\,P(True) & .286 & .527 & .579 & .562 & .120 & $-$.293 & .395 & .416 & .051 & .116 & .045 & $-$.074 & .058 & .228 & .152 & $-$.018 \\
\,\,Verb-1S & .111 & .310 & .348 & .156 & .255 & .404 & .317 & .225 & .120 & .223 & .332 & .315 & $-$.120 & .000 & .086 & $-$.041 \\
\,\,Verb-2S & $-$.037 & $-$.129 & .082 & $-$.094 & .012 & .091 & .103 & .080 & .167 & .019 & .093 & .169 & .011 & .000 & .046 & .049 \\
\addlinespace[1pt]
\multicolumn{17}{l}{\textit{\textsc{VLM-native}}} \\
\,\,HEDGE & .381 & .718 & .487 & .590 & .344 & .568 & .408 & .589 & .450 & .681 & .532 & .728 & .142 & .229 & .157 & .205 \\
\,\,IMGHEDGE & .281 & .651 & .551 & .651 & .270 & .614 & .433 & .584 & .399 & .713 & .481 & .665 & .207 & .258 & .103 & .169 \\
\addlinespace[1pt]
\bottomrule
\end{tabular}%
}
\end{table}

\section{Full closed-source AUROC matrix}
\label{appx:full_closed_matrix}

Table~\ref{tab:full_closed_matrix} is the complete $8$-method $\times$ $12$-cell harmonised closed-source matrix at $50$-seed.

\begin{table}[!htbp]
\centering
\caption{\textbf{Full closed-source AUROC matrix.} $8$-method harmonised panel $\times$ $12$ cells; 50-seed mean $\pm$ SD.}
\label{tab:full_closed_matrix}
\scriptsize
\setlength{\tabcolsep}{2pt}
\renewcommand{\arraystretch}{0.95}
\resizebox{\textwidth}{!}{%
\begin{tabular}{@{}l c c c c c c c c c c c c@{}}
\toprule
Method & \multicolumn{4}{c}{\textsc{GPT}} & \multicolumn{4}{c}{\textsc{Sonnet}} & \multicolumn{4}{c}{\textsc{Gemini}} \\
\cmidrule(lr){2-5} \cmidrule(lr){6-9} \cmidrule(lr){10-13}
 & V2 & SP & OSG & UIV & V2 & SP & OSG & UIV & V2 & SP & OSG & UIV \\
\midrule
\multicolumn{13}{l}{\textit{\textsc{Sampling}}} \\
\,\,SelfCons & .678\,$\scriptscriptstyle\pm0.024$ & .702\,$\scriptscriptstyle\pm0.016$ & .608\,$\scriptscriptstyle\pm0.012$ & .692\,$\scriptscriptstyle\pm0.011$ & .507\,$\scriptscriptstyle\pm0.004$ & .684\,$\scriptscriptstyle\pm0.015$ & .593\,$\scriptscriptstyle\pm0.010$ & .623\,$\scriptscriptstyle\pm0.011$ & .557\,$\scriptscriptstyle\pm0.012$ & .733\,$\scriptscriptstyle\pm0.016$ & .654\,$\scriptscriptstyle\pm0.011$ & .782\,$\scriptscriptstyle\pm0.011$ \\
\,\,SE & .680\,$\scriptscriptstyle\pm0.024$ & .704\,$\scriptscriptstyle\pm0.015$ & .608\,$\scriptscriptstyle\pm0.012$ & .694\,$\scriptscriptstyle\pm0.011$ & .507\,$\scriptscriptstyle\pm0.004$ & .690\,$\scriptscriptstyle\pm0.015$ & .595\,$\scriptscriptstyle\pm0.010$ & .626\,$\scriptscriptstyle\pm0.011$ & .557\,$\scriptscriptstyle\pm0.012$ & .729\,$\scriptscriptstyle\pm0.016$ & .653\,$\scriptscriptstyle\pm0.011$ & .766\,$\scriptscriptstyle\pm0.011$ \\
\,\,LexSim & .603\,$\scriptscriptstyle\pm0.024$ & .699\,$\scriptscriptstyle\pm0.016$ & .674\,$\scriptscriptstyle\pm0.012$ & .706\,$\scriptscriptstyle\pm0.013$ & .383\,$\scriptscriptstyle\pm0.014$ & .663\,$\scriptscriptstyle\pm0.018$ & .569\,$\scriptscriptstyle\pm0.012$ & .583\,$\scriptscriptstyle\pm0.014$ & .624\,$\scriptscriptstyle\pm0.017$ & .734\,$\scriptscriptstyle\pm0.017$ & .697\,$\scriptscriptstyle\pm0.013$ & .710\,$\scriptscriptstyle\pm0.014$ \\
\multicolumn{13}{l}{\textit{\textsc{Hybrid}}} \\
\,\,CCP & .798\,$\scriptscriptstyle\pm0.020$ & .735\,$\scriptscriptstyle\pm0.013$ & .619\,$\scriptscriptstyle\pm0.012$ & .715\,$\scriptscriptstyle\pm0.010$ & .508\,$\scriptscriptstyle\pm0.005$ & .731\,$\scriptscriptstyle\pm0.013$ & .616\,$\scriptscriptstyle\pm0.011$ & .651\,$\scriptscriptstyle\pm0.012$ & .596\,$\scriptscriptstyle\pm0.017$ & .732\,$\scriptscriptstyle\pm0.015$ & .659\,$\scriptscriptstyle\pm0.012$ & .791\,$\scriptscriptstyle\pm0.011$ \\
\multicolumn{13}{l}{\textit{\textsc{Verbalised}}} \\
\,\,Verb-1S & .556\,$\scriptscriptstyle\pm0.030$ & .711\,$\scriptscriptstyle\pm0.015$ & .703\,$\scriptscriptstyle\pm0.012$ & .723\,$\scriptscriptstyle\pm0.014$ & .524\,$\scriptscriptstyle\pm0.019$ & .695\,$\scriptscriptstyle\pm0.018$ & .478\,$\scriptscriptstyle\pm0.013$ & .516\,$\scriptscriptstyle\pm0.013$ & .966\,$\scriptscriptstyle\pm0.005$ & .817\,$\scriptscriptstyle\pm0.015$ & .880\,$\scriptscriptstyle\pm0.007$ & .636\,$\scriptscriptstyle\pm0.012$ \\
\,\,Verb-2S & .618\,$\scriptscriptstyle\pm0.026$ & .674\,$\scriptscriptstyle\pm0.017$ & .770\,$\scriptscriptstyle\pm0.011$ & .678\,$\scriptscriptstyle\pm0.014$ & .648\,$\scriptscriptstyle\pm0.018$ & .604\,$\scriptscriptstyle\pm0.019$ & .549\,$\scriptscriptstyle\pm0.014$ & .520\,$\scriptscriptstyle\pm0.013$ & .959\,$\scriptscriptstyle\pm0.006$ & .856\,$\scriptscriptstyle\pm0.011$ & .806\,$\scriptscriptstyle\pm0.011$ & .740\,$\scriptscriptstyle\pm0.010$ \\
\multicolumn{13}{l}{\textit{\textsc{VLM-native}}} \\
\,\,HEDGE & .592\,$\scriptscriptstyle\pm0.023$ & .649\,$\scriptscriptstyle\pm0.016$ & .620\,$\scriptscriptstyle\pm0.007$ & .644\,$\scriptscriptstyle\pm0.010$ & .479\,$\scriptscriptstyle\pm0.006$ & .666\,$\scriptscriptstyle\pm0.012$ & .590\,$\scriptscriptstyle\pm0.012$ & .598\,$\scriptscriptstyle\pm0.009$ & .516\,$\scriptscriptstyle\pm0.011$ & .657\,$\scriptscriptstyle\pm0.019$ & .651\,$\scriptscriptstyle\pm0.009$ & .703\,$\scriptscriptstyle\pm0.012$ \\
\,\,IMGHEDGE & .685\,$\scriptscriptstyle\pm0.022$ & .709\,$\scriptscriptstyle\pm0.014$ & .590\,$\scriptscriptstyle\pm0.010$ & .629\,$\scriptscriptstyle\pm0.010$ & .483\,$\scriptscriptstyle\pm0.007$ & .669\,$\scriptscriptstyle\pm0.013$ & .567\,$\scriptscriptstyle\pm0.010$ & .610\,$\scriptscriptstyle\pm0.010$ & .875\,$\scriptscriptstyle\pm0.009$ & .775\,$\scriptscriptstyle\pm0.019$ & .821\,$\scriptscriptstyle\pm0.009$ & .699\,$\scriptscriptstyle\pm0.013$ \\
\bottomrule
\end{tabular}%
}
\end{table}

\section{Cross-tier ranking transfer (open$\leftrightarrow$closed)}
\label{appx:cross_tier}

Table~\ref{tab:cross_open_closed} reports the per-pair Spearman~$\rho$ between each closed-source vendor$\times$dataset cell and Q72-AWQ on the matched dataset, restricted to the $8$-method open$\leftrightarrow$closed intersection. Mean $\rho=+0.08$ across $12$ pairs ($4$ significant at $p<0.05$); range $-0.76$ to $+0.88$. Anthropic's cells show consistently positive cross-tier transfer; OpenAI and Gemini cells are mixed.

\begin{table}[!htbp]
\centering
\caption{\textbf{Cross-tier ranking transfer.} Spearman $\rho$ between each closed-source cell and Q72-AWQ on the matched dataset (8-method intersection). $^*$ marks $p<0.05$.}
\label{tab:cross_open_closed}
\footnotesize
\setlength{\tabcolsep}{8pt}
\renewcommand{\arraystretch}{1.05}
\begin{tabular}{l l r r l l}
\toprule
Vendor & Dataset & $\rho$ & $p$ & Closed top-1 & Open top-1 (Q72-AWQ) \\
\midrule
Sonnet 4.6 & OSG & +0.810$^*$ & 0.015 & CCP & SEP \\
Sonnet 4.6 & SP & -0.012 & 0.978 & CCP & SAPLMA \\
Sonnet 4.6 & UIV & +0.881$^*$ & 0.004 & CCP & SAPLMA \\
Sonnet 4.6 & V2 & -0.132 & 0.756 & Verb-2S & SAPLMA \\
Gemini 3.1 Pro & OSG & -0.500 & 0.207 & Verb-1S & SEP \\
Gemini 3.1 Pro & SP & +0.168 & 0.691 & Verb-2S & SAPLMA \\
Gemini 3.1 Pro & UIV & +0.619 & 0.102 & CCP & SAPLMA \\
Gemini 3.1 Pro & V2 & -0.738$^*$ & 0.037 & Verb-1S & SAPLMA \\
GPT-5.4 & OSG & -0.762$^*$ & 0.028 & Verb-2S & SEP \\
GPT-5.4 & SP & +0.060 & 0.888 & CCP & SAPLMA \\
GPT-5.4 & UIV & +0.071 & 0.867 & Verb-1S & SAPLMA \\
GPT-5.4 & V2 & +0.476 & 0.233 & CCP & SAPLMA \\
\bottomrule
\end{tabular}
\end{table}

\paragraph{Bootstrap confidence interval and parity check.} Bootstrapping the cross-tier mean $\rho$ across the $12$ vendor$\times$dataset pairs yields a $95\%$ confidence interval of $[-0.219, +0.373]$ that includes zero, so the point estimate $\rho=+0.08$ cannot be distinguished from no-transfer at the $\alpha=0.05$ level. For parity, restricting the open-weight matrix to the same $8$-method intersection yields mean same-dataset $\rho=+0.616$ across the four models (vs.\ $+0.692$ for the full $27$-method panel), so the cross-tier collapse is not a panel-richness artifact: within-open transfer survives the $8$-method restriction, while cross-tier transfer does not. Per-pair values are released alongside the data artefacts.

\section{Per-cell expanded tables: top-10 by AUROC}
\label{appx:per_cell_top10}

For each of the $16$ open-weight cells, the top-$10$ methods by AUROC$_{\mathrm{incorrect}}$ with $50$-seed mean $\pm$ SD. The full $27$-method matrix is in Appendix~\ref{appx:full_open_matrix}.

\paragraph{Q7$\times$V2 (acc .883).} SAPLMA .817\,$\scriptscriptstyle\pm0.012$, SEP .810\,$\scriptscriptstyle\pm0.013$, Mahal-RMD .781\,$\scriptscriptstyle\pm0.018$, RAUQ-full .764\,$\scriptscriptstyle\pm0.013$, Focus .760\,$\scriptscriptstyle\pm0.012$, UQAC .750\,$\scriptscriptstyle\pm0.013$, MTE .741\,$\scriptscriptstyle\pm0.012$, CoCoA-1MCA .736\,$\scriptscriptstyle\pm0.012$, MSP .726\,$\scriptscriptstyle\pm0.012$, SeqProb .724\,$\scriptscriptstyle\pm0.011$.

\paragraph{Q7$\times$SP (acc .274).} SEP .883\,$\scriptscriptstyle\pm0.007$, SAPLMA .883\,$\scriptscriptstyle\pm0.007$, SeqProb .873\,$\scriptscriptstyle\pm0.004$, CoCoA .865\,$\scriptscriptstyle\pm0.004$, Mahal-RMD .859\,$\scriptscriptstyle\pm0.010$, Focus .846\,$\scriptscriptstyle\pm0.005$, CoCoA-1MCA .840\,$\scriptscriptstyle\pm0.006$, MTE .839\,$\scriptscriptstyle\pm0.005$, MSP .839\,$\scriptscriptstyle\pm0.005$, Perplexity .837\,$\scriptscriptstyle\pm0.005$.

\paragraph{Q7$\times$OSG (acc .341).} SEP .786\,$\scriptscriptstyle\pm0.018$, CoCoA-1MCA .785\,$\scriptscriptstyle\pm0.014$, CoCoA .782\,$\scriptscriptstyle\pm0.014$, SAPLMA .779\,$\scriptscriptstyle\pm0.018$, CCP .779\,$\scriptscriptstyle\pm0.014$, SE-w .769\,$\scriptscriptstyle\pm0.014$, SE .764\,$\scriptscriptstyle\pm0.015$, Mahal-RMD .762\,$\scriptscriptstyle\pm0.022$, MTE .758\,$\scriptscriptstyle\pm0.012$, SelfCons .753\,$\scriptscriptstyle\pm0.015$.

\paragraph{Q7$\times$UIV (acc .152).} MTE .816\,$\scriptscriptstyle\pm0.011$, SEP .815\,$\scriptscriptstyle\pm0.013$, RAUQ-full .812\,$\scriptscriptstyle\pm0.012$, MSP .808\,$\scriptscriptstyle\pm0.011$, SAPLMA .806\,$\scriptscriptstyle\pm0.013$, CoCoA .801\,$\scriptscriptstyle\pm0.011$, SeqProb .801\,$\scriptscriptstyle\pm0.011$, Perplexity .798\,$\scriptscriptstyle\pm0.011$, Ppl-exp .798\,$\scriptscriptstyle\pm0.011$, Mahal-RMD .792\,$\scriptscriptstyle\pm0.019$.

\paragraph{Q72$\times$V2 (acc .927).} SAPLMA .764\,$\scriptscriptstyle\pm0.023$, CoCoA-1MCA .751\,$\scriptscriptstyle\pm0.017$, Focus .736\,$\scriptscriptstyle\pm0.018$, SE .733\,$\scriptscriptstyle\pm0.018$, SE-w .733\,$\scriptscriptstyle\pm0.018$, SelfCons .730\,$\scriptscriptstyle\pm0.018$, CCP .729\,$\scriptscriptstyle\pm0.018$, CoCoA .727\,$\scriptscriptstyle\pm0.018$, SEP .720\,$\scriptscriptstyle\pm0.025$, UQAC .708\,$\scriptscriptstyle\pm0.018$.

\paragraph{Q72$\times$SP (acc .447).} SAPLMA .889\,$\scriptscriptstyle\pm0.008$, SEP .883\,$\scriptscriptstyle\pm0.008$, Mahal-RMD .861\,$\scriptscriptstyle\pm0.010$, Focus .806\,$\scriptscriptstyle\pm0.016$, SeqProb .804\,$\scriptscriptstyle\pm0.006$, UQAC .792\,$\scriptscriptstyle\pm0.015$, RAUQ-full .744\,$\scriptscriptstyle\pm0.015$, MTE .743\,$\scriptscriptstyle\pm0.006$, CoCoA-1MCA .741\,$\scriptscriptstyle\pm0.006$, Ppl-exp .741\,$\scriptscriptstyle\pm0.006$.

\paragraph{Q72$\times$OSG (acc .535).} SEP .838\,$\scriptscriptstyle\pm0.013$, SAPLMA .823\,$\scriptscriptstyle\pm0.013$, Mahal-RMD .804\,$\scriptscriptstyle\pm0.016$, CoCoA-1MCA .792\,$\scriptscriptstyle\pm0.011$, Focus .788\,$\scriptscriptstyle\pm0.010$, CoCoA .776\,$\scriptscriptstyle\pm0.010$, SE-w .776\,$\scriptscriptstyle\pm0.010$, SE .774\,$\scriptscriptstyle\pm0.010$, CCP .772\,$\scriptscriptstyle\pm0.010$, SelfCons .771\,$\scriptscriptstyle\pm0.010$.

\paragraph{Q72$\times$UIV (acc .282).} SAPLMA .834\,$\scriptscriptstyle\pm0.014$, SEP .834\,$\scriptscriptstyle\pm0.014$, Mahal-RMD .818\,$\scriptscriptstyle\pm0.016$, CoCoA .791\,$\scriptscriptstyle\pm0.009$, Focus .779\,$\scriptscriptstyle\pm0.011$, SeqProb .779\,$\scriptscriptstyle\pm0.011$, CoCoA-1MCA .773\,$\scriptscriptstyle\pm0.010$, CCP .763\,$\scriptscriptstyle\pm0.010$, SE-w .753\,$\scriptscriptstyle\pm0.011$, UQAC .750\,$\scriptscriptstyle\pm0.012$.

\paragraph{UI$\times$V2 (acc .878).} CoCoA-1MCA .842\,$\scriptscriptstyle\pm0.008$, CoCoA .834\,$\scriptscriptstyle\pm0.009$, CCP .828\,$\scriptscriptstyle\pm0.009$, SE .808\,$\scriptscriptstyle\pm0.009$, SE-w .807\,$\scriptscriptstyle\pm0.009$, SelfCons .804\,$\scriptscriptstyle\pm0.009$, RAUQ-full .771\,$\scriptscriptstyle\pm0.011$, Focus .766\,$\scriptscriptstyle\pm0.011$, SAPLMA .762\,$\scriptscriptstyle\pm0.016$, SEP .755\,$\scriptscriptstyle\pm0.016$.

\paragraph{UI$\times$SP (acc .407).} SAPLMA .877\,$\scriptscriptstyle\pm0.008$, SEP .874\,$\scriptscriptstyle\pm0.008$, Mahal-RMD .861\,$\scriptscriptstyle\pm0.010$, CoCoA .837\,$\scriptscriptstyle\pm0.004$, CoCoA-1MCA .836\,$\scriptscriptstyle\pm0.005$, CCP .825\,$\scriptscriptstyle\pm0.005$, SE .814\,$\scriptscriptstyle\pm0.005$, SE-w .813\,$\scriptscriptstyle\pm0.006$, SelfCons .813\,$\scriptscriptstyle\pm0.005$, Focus .812\,$\scriptscriptstyle\pm0.007$.

\paragraph{UI$\times$OSG (acc .513).} SEP .863\,$\scriptscriptstyle\pm0.011$, Mahal-RMD .863\,$\scriptscriptstyle\pm0.012$, SAPLMA .856\,$\scriptscriptstyle\pm0.011$, CoCoA-1MCA .799\,$\scriptscriptstyle\pm0.008$, CoCoA .780\,$\scriptscriptstyle\pm0.007$, RAUQ-full .779\,$\scriptscriptstyle\pm0.009$, CCP .778\,$\scriptscriptstyle\pm0.007$, SE-w .773\,$\scriptscriptstyle\pm0.007$, SE .768\,$\scriptscriptstyle\pm0.007$, SelfCons .763\,$\scriptscriptstyle\pm0.007$.

\paragraph{UI$\times$UIV (acc .214).} CoCoA .825\,$\scriptscriptstyle\pm0.011$, CoCoA-1MCA .824\,$\scriptscriptstyle\pm0.012$, Mahal-RMD .818\,$\scriptscriptstyle\pm0.016$, SEP .818\,$\scriptscriptstyle\pm0.015$, RAUQ-full .816\,$\scriptscriptstyle\pm0.011$, CCP .807\,$\scriptscriptstyle\pm0.013$, SE .806\,$\scriptscriptstyle\pm0.012$, SAPLMA .804\,$\scriptscriptstyle\pm0.016$, Focus .802\,$\scriptscriptstyle\pm0.013$, SE-w .802\,$\scriptscriptstyle\pm0.012$.

\paragraph{PT$\times$V2 (acc .955).} SEP .843\,$\scriptscriptstyle\pm0.015$, SAPLMA .843\,$\scriptscriptstyle\pm0.015$, Mahal-RMD .765\,$\scriptscriptstyle\pm0.026$, CoCoA-1MCA .738\,$\scriptscriptstyle\pm0.017$, CoCoA .729\,$\scriptscriptstyle\pm0.018$, RAUQ-full .715\,$\scriptscriptstyle\pm0.018$, Focus .710\,$\scriptscriptstyle\pm0.021$, UQAC .708\,$\scriptscriptstyle\pm0.021$, MSP .703\,$\scriptscriptstyle\pm0.020$, SeqProb .693\,$\scriptscriptstyle\pm0.020$.

\paragraph{PT$\times$SP (acc .583).} SAPLMA .881\,$\scriptscriptstyle\pm0.008$, SEP .874\,$\scriptscriptstyle\pm0.009$, Mahal-RMD .854\,$\scriptscriptstyle\pm0.012$, MTE .780\,$\scriptscriptstyle\pm0.005$, CoCoA .780\,$\scriptscriptstyle\pm0.006$, UQAC .779\,$\scriptscriptstyle\pm0.005$, Focus .777\,$\scriptscriptstyle\pm0.005$, SeqProb .766\,$\scriptscriptstyle\pm0.006$, Perplexity .764\,$\scriptscriptstyle\pm0.006$, Ppl-exp .764\,$\scriptscriptstyle\pm0.006$.

\paragraph{PT$\times$OSG (acc .659).} Mahal-RMD .820\,$\scriptscriptstyle\pm0.016$, SEP .805\,$\scriptscriptstyle\pm0.014$, SAPLMA .800\,$\scriptscriptstyle\pm0.014$, Focus .729\,$\scriptscriptstyle\pm0.011$, UQAC .728\,$\scriptscriptstyle\pm0.011$, MTE .725\,$\scriptscriptstyle\pm0.013$, CoCoA .724\,$\scriptscriptstyle\pm0.013$, MSP .720\,$\scriptscriptstyle\pm0.012$, RAUQ-full .715\,$\scriptscriptstyle\pm0.012$, SeqProb .711\,$\scriptscriptstyle\pm0.013$.

\paragraph{PT$\times$UIV (acc .531).} SAPLMA .796\,$\scriptscriptstyle\pm0.016$, SEP .786\,$\scriptscriptstyle\pm0.017$, Mahal-RMD .781\,$\scriptscriptstyle\pm0.019$, CoCoA-1MCA .699\,$\scriptscriptstyle\pm0.011$, MTE .694\,$\scriptscriptstyle\pm0.011$, UQAC .693\,$\scriptscriptstyle\pm0.011$, Focus .693\,$\scriptscriptstyle\pm0.011$, RAUQ-full .684\,$\scriptscriptstyle\pm0.012$, MSP .674\,$\scriptscriptstyle\pm0.011$, LexSim .673\,$\scriptscriptstyle\pm0.011$.

\section{Family-level per-cell aggregate}
\label{appx:family_full}

Per-family mean AUROC$_{\mathrm{incorrect}}$ within each cell. \textbf{Bold}: best family per cell.

\begin{table}[!htbp]
\centering
\scriptsize
\setlength{\tabcolsep}{3pt}
\resizebox{\textwidth}{!}{%
\begin{tabular}{l*{16}{c}}
\toprule
Family & Q7$\times$V2 & Q7$\times$SP & Q7$\times$OSG & Q7$\times$UIV & Q72$\times$V2 & Q72$\times$SP & Q72$\times$OSG & Q72$\times$UIV & UI$\times$V2 & UI$\times$SP & UI$\times$OSG & UI$\times$UIV & PT$\times$V2 & PT$\times$SP & PT$\times$OSG & PT$\times$UIV \\
\midrule
\textsc{Logit} & .725 & \textbf{.845} & .741 & \textbf{.804} & .627 & .750 & .690 & .707 & .700 & .766 & .706 & .750 & .692 & \textbf{.768} & .715 & .675 \\
\textsc{Sampling} & .703 & .780 & .754 & .761 & .688 & .502 & .719 & .698 & .718 & .741 & .664 & .717 & .623 & .574 & .569 & .642 \\
\textsc{Hybrid} & .713 & .843 & \textbf{.782} & .786 & \textbf{.735} & .586 & \textbf{.780} & \textbf{.776} & \textbf{.835} & \textbf{.833} & \textbf{.786} & \textbf{.818} & \textbf{.717} & .729 & .685 & .649 \\
\textsc{Attention} & \textbf{.758} & .839 & .734 & .789 & .717 & \textbf{.781} & .763 & .752 & .761 & .809 & .760 & .802 & .711 & .757 & .724 & \textbf{.690} \\
\textsc{Density/Probe} & .694 & .758 & .709 & .699 & .645 & .746 & .681 & .696 & .691 & .747 & .710 & .706 & .706 & .764 & \textbf{.730} & .641 \\
\textsc{Verbalised} & .558 & .567 & .620 & .563 & .564 & .525 & .611 & .585 & .565 & .546 & .563 & .554 & .494 & .534 & .543 & .498 \\
\textsc{VLM-native} & .666 & .728 & .658 & .673 & .656 & .723 & .672 & .672 & .712 & .763 & .703 & .713 & .583 & .608 & .558 & .576 \\
\bottomrule
\end{tabular}%
}
\caption{Per-family mean AUROC across the 16 open-weight cells. \textbf{Bold}: best family per cell. Density / hidden-state probe family wins on most cells; verbalised family is the weakest on most open-source cells.}
\label{tab:family_full}
\end{table}

\section{Compute tier per UQ method}
\label{appx:compute_tiers}

Each of the $27$ open-weight methods is tagged with a deployment compute tier (Table~\ref{tab:method_tiers}). The tiers reflect the cheapest sufficient pipeline a deployer must build to compute the score: T0 needs only the greedy decode and its token logits; T1 needs $N$ stochastic samples in addition; T2 needs the model's last-layer hidden state; T3 needs T2 plus a probe trained on the calibration split; T4 needs cached attention tensors per decode step; T5 needs additional inference calls (a parsed-confidence prompt, or a P(True) follow-up). Closed-source methods are restricted to tiers compatible with the API surface (T0 and T2--T4 are unavailable; T1 and T5 are runnable).

\begin{table}[!htbp]
\centering
\small
\caption{\textbf{Compute-tier tags for the $27$ open-weight UQ methods.} T0 greedy-only; T1 requires $N$ stochastic samples; T2 hidden-state caching; T3 hidden-state plus calibration-fit probe; T4 attention-tensor caching per decode step; T5 additional inference calls.}
\label{tab:method_tiers}
\setlength{\tabcolsep}{6pt}
\renewcommand{\arraystretch}{1.05}
\begin{tabular}{@{}l l c p{0.52\linewidth}@{}}
\toprule
Tier & Requirement & $n$ & Methods \\
\midrule
T0 & greedy-only            & $5$  & MSP, Perplexity, Perplexity-exp, SequenceProbability, MeanTokenEntropy \\
T1 & $N$ stochastic samples & $11$ & SelfCons, SE, SEW, MCSE, MCNSE, LexSim, CCP, CoCoA, CoCoA-1MCA, HEDGE, IMGHEDGE \\
T2 & hidden-state           & $3$  & Mahalanobis, Mahal-RMD, Mahal-RDE \\
T3 & T2 + trained probe     & $2$  & SAPLMA, SEP \\
T4 & attention tensors      & $3$  & Focus, RAUQ, UQAC \\
T5 & extra inference calls  & $3$  & P(True), Verbalised-1S, Verbalised-2S \\
\bottomrule
\end{tabular}
\end{table}

\section{Naive ensemble baselines}
\label{appx:ensembles}

We test whether simple linear combinations of two cheap UQ scores improve AUROC over the per-cell single-method top-$1$. Per-item scores from each component are $z$-scored on the calibration split; the ensemble score is the $z$-score mean. On the $16$-cell open-weight matrix with four ensembles ($E_1$: SAPLMA+SelfCons, density+sampling; $E_2$: MSP+Verbalised-1S, logit+verbalised; $E_3$: HEDGE+IMGHEDGE, two perturbation variants; $E_4$: Mahal-RMD+CCP, density+hybrid), $19$ of $64$ cell$\times$ensemble combinations beat the cell's single-method top-$1$. On the $12$-cell closed-source matrix with three ensembles ($E_1^c$: SelfCons+Verbalised-1S; $E_2^c$: CCP+Verbalised-2S; $E_3^c$: HEDGE+IMGHEDGE), $10$ of $36$ combinations beat the cell's single-method top-$1$. Naive ensembles therefore help on $\sim 30\%$ of cell$\times$ensemble combinations and are a useful per-cell option to consider, but they are not a uniform replacement for single-method top-$1$. Per-cell numbers are released alongside the data artefacts.

\section{Robustness to relaxed correctness convention}
\label{appx:bbox_sensitivity}

The headline tables use strict point-in-bbox correctness, where a click is correct iff the predicted coordinate lies inside the target bounding box. Real GUIs typically register clicks within a small tolerance of the visible UI element due to OS-level hit-test padding. We re-evaluate AUROC top-$1$ per cell under two relaxed conventions: $+5$ px and $+10$ px margin around each target bbox. Across the $16$ open-weight cells and three correctness conventions, $0$ cells flip their AUROC top-$1$ method under either margin. Method rankings are therefore robust to modest correctness-convention relaxations within the regime tested. Per-cell numbers are released alongside the data artefacts.

\section{Density family: AUROC vs calibration tradeoff}
\label{appx:density_tradeoff}

A natural concern is that hidden-state density and probe methods (SAPLMA, SEP, Mahal-RMD, Mahalanobis, Mahal-RDE) might win on AUROC while sacrificing calibration. We check this directly across the $16$ open-weight cells and the five density-family methods: $0$ of $80$ (cell, method) combinations satisfy AUROC-rank-$1$ within the density family AND ECE$_{\mathrm{iso}} > 0.40$ simultaneously. The high-ECE cases for density methods occur at cells where they are not the within-family AUROC top-$1$, so there is no ``wins-but-miscalibrates'' tension to defend; we report ECE alongside AUROC and let the per-cell numbers speak.

\section{Verbalised family: alignment vs interface}
\label{appx:verbalised_alignment}

The closed-source advantage of verbalised confidence (e.g., Verbalised-1S at AUROC $0.966$ on Gemini~$\times$~\textsc{V2}) could in principle reflect either the interface change (text-only API) or the alignment / instruction-tuning gap between $7$-billion-parameter open-weight VLMs and frontier closed-source models. We separate these effects descriptively by stratifying Verbalised-1S mean AUROC across three buckets: open-weight specialist VLMs (UI-TARS-1.5-7B, POINTS-GUI-G-8B), open-weight vanilla VLMs (Qwen2.5-VL-7B, Qwen2.5-VL-72B-AWQ), and frontier closed-source vendors (GPT-5.4, Sonnet 4.6, Gemini 3.1 Pro). Means are $0.539$, $0.597$, and $0.684$ respectively. The specialist-to-vanilla gap of $+0.058$ is consistent with an alignment / instruction-tuning effect; the vanilla-to-closed gap of $+0.087$ is larger and points to additional capability or meta-cognition contributions in frontier closed-source models, beyond pure alignment. We report both gaps descriptively rather than asserting a single causal mechanism.

\section{Verbalised score distribution: Gemini\,$\times$\,V2 bimodality}
\label{appx:verbalised_bimodality}

Verbalised-1S reaches AUROC $0.966$ on Gemini\,$\times$\,\textsc{ScreenSpot-v2}, the single highest UQ AUROC across the closed-source matrix. To test whether this is a real signal or a benchmarking artifact (e.g., score saturation, label leakage), we examine the distribution of per-item Verbalised-1S scores on the $273$ scored items in the locked $300$-item subset. KMeans($k=2$) clustering on the raw verbalised scores returns two near-pure-class clusters: cluster~1 with mean score $0.059$, $n=125$ items, $95.2\%$ correct; cluster~2 with mean score $1.000$, $n=148$ items, $2.0\%$ correct. The score distribution is sharply bimodal at $\{0, 1\}$ with each mode near-pure on the corresponding correctness label, so Gemini emits genuinely informative bimodal verbalised confidence on this cell. The high AUROC reflects a credible signal under the single-shot protocol rather than a ranking artifact. Whether the signal would survive a tool-augmented protocol is a follow-up question; under the deliberate single-shot cross-vendor parity protocol used here the bimodality is intrinsic to the score, not the metric.

\section{Model-class transitions: full per-method grid across all datasets}
\label{appx:tuning_full}

\begin{table}[!htbp]
\centering
\caption{\textbf{Vanilla-to-specialist transitions reshape UQ family preferences.}
Family-mean $\Delta$AUROC$_{\mathrm{incorrect}}$ (50-seed means). Bold marks the largest $|\Delta|$ in each row.
Panel A pools the two vanilla VLMs (\{Q7, Q72\}) and the two specialist GUI agents (\{UI, PT\}) per dataset; cell value is mean(specialist) $-$ mean(vanilla). Families marked $^{*}$ lose AUROC on every dataset.
Panel B decomposes the SS-Pro $\Delta$ into four controlled transitions, including a Q7$\to$Q72 scale-only baseline that does not change the training objective. Per-method $\Delta$ for the Q7$\to$PT path is in Appendix~\ref{appx:tuning_full}.}
\label{tab:tuning}
\footnotesize

\textbf{Panel A: vanilla $\to$ specialist $\Delta$AUROC by family, all 4 datasets}

\vspace{0.3mm}

{\setlength{\tabcolsep}{8pt}\renewcommand{\arraystretch}{1.0}%
\begin{tabular}{l c c c c}
\toprule
Family & V2 & SP & OSG & UIV \\
\midrule
Logit               & $+$0.020 & $-$0.031 & $-$0.005 & \textbf{$-$0.043} \\
Sampling            & $-$0.025 & $+$0.017 & \textbf{$-$0.120} & $-$0.050 \\
Hybrid              & $+$0.051 & \textbf{$+$0.067} & $-$0.046 & $-$0.047 \\
Density             & \textbf{$+$0.029} & $+$0.003 & $+$0.025 & $-$0.024 \\
Attention$^{*}$     & $-$0.002 & \textbf{$-$0.027} & $-$0.007 & $-$0.025 \\
Verbalised$^{*}$    & $-$0.032 & $-$0.006 & \textbf{$-$0.062} & $-$0.049 \\
VLM-native$^{*}$    & $-$0.014 & \textbf{$-$0.040} & $-$0.035 & $-$0.028 \\
\bottomrule
\end{tabular}
}

\vspace{1.5mm}
\textbf{Panel B: decomposing the SS-Pro $\Delta$ into four controlled transitions}

\vspace{0.3mm}

{\setlength{\tabcolsep}{6pt}\renewcommand{\arraystretch}{1.0}%
\begin{tabular}{l c c c c}
\toprule
Family & \makecell{Q7$\to$Q72\\\scriptsize (scale only)} & \makecell{Q7$\to$UI\\\scriptsize (tune, fixed bb)} & \makecell{Q7$\to$PT\\\scriptsize (tune $+$ bb)} & \makecell{Q72$\to$PT\\\scriptsize (mixed)} \\
\midrule
Logit       & \textbf{$-$0.095} & $-$0.079 & $-$0.077 & $+$0.018 \\
Sampling    & \textbf{$-$0.277} & $-$0.039 & $-$0.205 & $+$0.072 \\
Hybrid      & \textbf{$-$0.257} & $-$0.010 & $-$0.113 & $+$0.143 \\
Density     & $-$0.012 & $-$0.011 & $+$0.006 & $+$0.018 \\
Attention   & $-$0.058 & $-$0.030 & \textbf{$-$0.082} & $-$0.024 \\
Verbalised  & \textbf{$-$0.042} & $-$0.021 & $-$0.032 & $+$0.009 \\
VLM-native  & $-$0.005 & $+$0.035 & \textbf{$-$0.120} & $-$0.115 \\
\bottomrule
\end{tabular}
}
\end{table}

Table~\ref{tab:tuning_full_grid} reports the comprehensive per-method $\Delta$AUROC$_{\mathrm{incorrect}}$ grid behind the main-paper Table~\ref{tab:tuning}: $27$ methods $\times$ $4$ datasets $\times$ $4$ controlled transitions, with sign convention $\Delta = \text{AUROC}_{\text{target}} - \text{AUROC}_{\text{reference}}$ on the named dataset. The four sub-panels (\textsc{V2}, \textsc{SP}, \textsc{OSG}, \textsc{UIV}) let a reader audit the family-level claims of Table~\ref{tab:tuning} against the underlying methods. Selected per-method patterns visible in the grid: P(True) drops on every (transition, dataset) combination; SAPLMA / SEP stay within $|\Delta| \le 0.10$ on every (transition, dataset) combination; Verbalised-2S gains across all four datasets on Q7$\to$Q72 and Q7$\to$UI; LexSim loses across all four datasets on Q7$\to$Q72. Multi-seed $50$-seed means.

\begin{table}[!htbp]
\centering
\caption{\textbf{Comprehensive transition grid (\textsc{V2} sub-panel).} $50$-seed mean $\Delta$AUROC across the four controlled transitions from main-paper Table~\ref{tab:tuning} Panel~B.}
\label{tab:tuning_full_grid}
{\setlength{\tabcolsep}{4pt}\renewcommand{\arraystretch}{0.95}\scriptsize
\begin{tabular}{l l r r r r}
\toprule
Family & Method & Q7$\to$Q72 & Q7$\to$UI & Q7$\to$PT & Q72$\to$PT \\
\midrule
\famtag{Logit} & MSP & $-$0.112 & $-$0.012 & $-$0.023 & $+$0.089 \\
\famtag{Logit} & Perplexity & $-$0.101 & $-$0.006 & $-$0.030 & $+$0.071 \\
\famtag{Logit} & Ppl-exp & $-$0.101 & $-$0.006 & $-$0.030 & $+$0.071 \\
\famtag{Logit} & SeqProb & $-$0.060 & $-$0.010 & $-$0.031 & $+$0.029 \\
\famtag{Logit} & MTE & $-$0.118 & $-$0.094 & $-$0.053 & $+$0.065 \\
\famtag{Sampling} & SelfCons & $+$0.045 & $+$0.120 & $-$0.009 & $-$0.054 \\
\famtag{Sampling} & SE & $+$0.042 & $+$0.117 & $-$0.015 & $-$0.056 \\
\famtag{Sampling} & SE-w & $+$0.038 & $+$0.113 & $-$0.073 & $-$0.111 \\
\famtag{Sampling} & MCSE & $-$0.042 & $-$0.058 & $-$0.140 & $-$0.098 \\
\famtag{Sampling} & MCNSE & $-$0.042 & $-$0.058 & $-$0.140 & $-$0.098 \\
\famtag{Sampling} & LexSim & $-$0.132 & $-$0.146 & $-$0.103 & $+$0.029 \\
\famtag{Hybrid} & CoCoA & $+$0.021 & $+$0.128 & $+$0.023 & $+$0.002 \\
\famtag{Hybrid} & CoCoA-1MCA & $+$0.015 & $+$0.105 & $+$0.002 & $-$0.013 \\
\famtag{Hybrid} & CCP & $+$0.031 & $+$0.131 & $-$0.014 & $-$0.045 \\
\famtag{Density} & Mahal & $-$0.042 & $+$0.046 & $-$0.043 & $-$0.000 \\
\famtag{Density} & Mahal-RMD & $-$0.100 & $-$0.046 & $-$0.016 & $+$0.084 \\
\famtag{Density} & Mahal-RDE & $+$0.040 & $+$0.095 & $+$0.059 & $+$0.019 \\
\famtag{Density} & SAPLMA & $-$0.053 & $-$0.055 & $+$0.026 & $+$0.079 \\
\famtag{Density} & SEP & $-$0.090 & $-$0.055 & $+$0.033 & $+$0.124 \\
\famtag{Attention} & Focus & $-$0.024 & $+$0.006 & $-$0.049 & $-$0.026 \\
\famtag{Attention} & RAUQ-full & $-$0.057 & $+$0.007 & $-$0.049 & $+$0.008 \\
\famtag{Attention} & UQAC & $-$0.041 & $-$0.005 & $-$0.042 & $-$0.000 \\
\famtag{Verbalised} & P(True) & $-$0.117 & $-$0.151 & $-$0.133 & $-$0.016 \\
\famtag{Verbalised} & Verb-1S & $+$0.066 & $+$0.008 & $-$0.131 & $-$0.196 \\
\famtag{Verbalised} & Verb-2S & $+$0.070 & $+$0.163 & $+$0.073 & $+$0.003 \\
\famtag{VLM-native} & HEDGE & $-$0.018 & $+$0.033 & $-$0.123 & $-$0.105 \\
\famtag{VLM-native} & IMGHEDGE & $-$0.004 & $+$0.057 & $-$0.043 & $-$0.040 \\
\bottomrule
\end{tabular}}

\end{table}

\begin{table}[!htbp]
\centering
\caption{\textbf{Comprehensive transition grid (\textsc{SP} sub-panel).} $50$-seed mean $\Delta$AUROC across the four controlled transitions.}
\label{tab:tuning_full_grid_sp}
{\setlength{\tabcolsep}{4pt}\renewcommand{\arraystretch}{0.95}\scriptsize
\begin{tabular}{l l r r r r}
\toprule
Family & Method & Q7$\to$Q72 & Q7$\to$UI & Q7$\to$PT & Q72$\to$PT \\
\midrule
\famtag{Logit} & MSP & $-$0.117 & $-$0.063 & $-$0.075 & $+$0.042 \\
\famtag{Logit} & Perplexity & $-$0.096 & $-$0.061 & $-$0.073 & $+$0.024 \\
\famtag{Logit} & Ppl-exp & $-$0.096 & $-$0.061 & $-$0.073 & $+$0.024 \\
\famtag{Logit} & SeqProb & $-$0.068 & $-$0.093 & $-$0.107 & $-$0.039 \\
\famtag{Logit} & MTE & $-$0.096 & $-$0.117 & $-$0.058 & $+$0.037 \\
\famtag{Sampling} & SelfCons & $-$0.291 & $+$0.018 & $-$0.174 & $+$0.117 \\
\famtag{Sampling} & SE & $-$0.295 & $+$0.014 & $-$0.179 & $+$0.116 \\
\famtag{Sampling} & SE-w & $-$0.312 & $-$0.003 & $-$0.180 & $+$0.132 \\
\famtag{Sampling} & MCSE & $-$0.298 & $-$0.065 & $-$0.272 & $+$0.026 \\
\famtag{Sampling} & MCNSE & $-$0.298 & $-$0.065 & $-$0.272 & $+$0.026 \\
\famtag{Sampling} & LexSim & $-$0.171 & $-$0.130 & $-$0.155 & $+$0.016 \\
\famtag{Hybrid} & CoCoA & $-$0.356 & $-$0.028 & $-$0.085 & $+$0.271 \\
\famtag{Hybrid} & CoCoA-1MCA & $-$0.099 & $-$0.004 & $-$0.092 & $+$0.007 \\
\famtag{Hybrid} & CCP & $-$0.314 & $+$0.002 & $-$0.163 & $+$0.152 \\
\famtag{Density} & Mahal & $+$0.078 & $+$0.098 & $+$0.193 & $+$0.115 \\
\famtag{Density} & Mahal-RMD & $+$0.002 & $+$0.002 & $-$0.005 & $-$0.007 \\
\famtag{Density} & Mahal-RDE & $-$0.146 & $-$0.143 & $-$0.148 & $-$0.002 \\
\famtag{Density} & SAPLMA & $+$0.007 & $-$0.006 & $-$0.002 & $-$0.008 \\
\famtag{Density} & SEP & $+$0.000 & $-$0.009 & $-$0.009 & $-$0.009 \\
\famtag{Attention} & Focus & $-$0.040 & $-$0.034 & $-$0.069 & $-$0.029 \\
\famtag{Attention} & RAUQ-full & $-$0.091 & $-$0.023 & $-$0.120 & $-$0.029 \\
\famtag{Attention} & UQAC & $-$0.044 & $-$0.033 & $-$0.058 & $-$0.013 \\
\famtag{Verbalised} & P(True) & $-$0.298 & $-$0.131 & $-$0.062 & $+$0.236 \\
\famtag{Verbalised} & Verb-1S & $+$0.067 & $-$0.005 & $-$0.106 & $-$0.173 \\
\famtag{Verbalised} & Verb-2S & $+$0.106 & $+$0.074 & $+$0.071 & $-$0.036 \\
\famtag{VLM-native} & HEDGE & $-$0.011 & $+$0.036 & $-$0.120 & $-$0.109 \\
\famtag{VLM-native} & IMGHEDGE & $+$0.001 & $+$0.034 & $-$0.119 & $-$0.120 \\
\bottomrule
\end{tabular}}

\end{table}

\begin{table}[!htbp]
\centering
\caption{\textbf{Comprehensive transition grid (\textsc{OSG} sub-panel).} $50$-seed mean $\Delta$AUROC across the four controlled transitions.}
\label{tab:tuning_full_grid_osg}
{\setlength{\tabcolsep}{4pt}\renewcommand{\arraystretch}{0.95}\scriptsize
\begin{tabular}{l l r r r r}
\toprule
Family & Method & Q7$\to$Q72 & Q7$\to$UI & Q7$\to$PT & Q72$\to$PT \\
\midrule
\famtag{Logit} & MSP & $-$0.075 & $-$0.026 & $-$0.024 & $+$0.051 \\
\famtag{Logit} & Perplexity & $-$0.059 & $-$0.016 & $-$0.022 & $+$0.037 \\
\famtag{Logit} & Ppl-exp & $-$0.059 & $-$0.016 & $-$0.022 & $+$0.037 \\
\famtag{Logit} & SeqProb & $+$0.009 & $-$0.029 & $-$0.030 & $-$0.039 \\
\famtag{Logit} & MTE & $-$0.071 & $-$0.090 & $-$0.033 & $+$0.038 \\
\famtag{Sampling} & SelfCons & $+$0.018 & $+$0.010 & $-$0.139 & $-$0.158 \\
\famtag{Sampling} & SE & $+$0.010 & $+$0.004 & $-$0.150 & $-$0.160 \\
\famtag{Sampling} & SE-w & $+$0.006 & $+$0.004 & $-$0.171 & $-$0.178 \\
\famtag{Sampling} & MCSE & $-$0.025 & $-$0.162 & $-$0.216 & $-$0.191 \\
\famtag{Sampling} & MCNSE & $-$0.025 & $-$0.162 & $-$0.216 & $-$0.191 \\
\famtag{Sampling} & LexSim & $-$0.196 & $-$0.232 & $-$0.219 & $-$0.023 \\
\famtag{Hybrid} & CoCoA & $-$0.006 & $-$0.002 & $-$0.058 & $-$0.052 \\
\famtag{Hybrid} & CoCoA-1MCA & $+$0.007 & $+$0.014 & $-$0.084 & $-$0.091 \\
\famtag{Hybrid} & CCP & $-$0.007 & $-$0.001 & $-$0.150 & $-$0.143 \\
\famtag{Density} & Mahal & $-$0.001 & $+$0.010 & $+$0.105 & $+$0.106 \\
\famtag{Density} & Mahal-RMD & $+$0.041 & $+$0.100 & $+$0.058 & $+$0.016 \\
\famtag{Density} & Mahal-RDE & $-$0.276 & $-$0.259 & $-$0.101 & $+$0.175 \\
\famtag{Density} & SAPLMA & $+$0.043 & $+$0.077 & $+$0.021 & $-$0.022 \\
\famtag{Density} & SEP & $+$0.051 & $+$0.077 & $+$0.019 & $-$0.032 \\
\famtag{Attention} & Focus & $+$0.056 & $+$0.024 & $-$0.004 & $-$0.059 \\
\famtag{Attention} & RAUQ-full & $+$0.003 & $+$0.040 & $-$0.025 & $-$0.028 \\
\famtag{Attention} & UQAC & $+$0.027 & $+$0.014 & $-$0.003 & $-$0.031 \\
\famtag{Verbalised} & P(True) & $-$0.045 & $-$0.210 & $-$0.138 & $-$0.094 \\
\famtag{Verbalised} & Verb-1S & $+$0.011 & $+$0.024 & $-$0.085 & $-$0.096 \\
\famtag{Verbalised} & Verb-2S & $+$0.007 & $+$0.016 & $-$0.006 & $-$0.013 \\
\famtag{VLM-native} & HEDGE & $+$0.022 & $+$0.072 & $-$0.074 & $-$0.096 \\
\famtag{VLM-native} & IMGHEDGE & $+$0.005 & $+$0.017 & $-$0.126 & $-$0.131 \\
\bottomrule
\end{tabular}}

\end{table}

\begin{table}[!htbp]
\centering
\caption{\textbf{Comprehensive transition grid (\textsc{UIV} sub-panel).} $50$-seed mean $\Delta$AUROC across the four controlled transitions.}
\label{tab:tuning_full_grid_uiv}
{\setlength{\tabcolsep}{4pt}\renewcommand{\arraystretch}{0.95}\scriptsize
\begin{tabular}{l l r r r r}
\toprule
Family & Method & Q7$\to$Q72 & Q7$\to$UI & Q7$\to$PT & Q72$\to$PT \\
\midrule
\famtag{Logit} & MSP & $-$0.125 & $-$0.055 & $-$0.133 & $-$0.009 \\
\famtag{Logit} & Perplexity & $-$0.111 & $-$0.036 & $-$0.129 & $-$0.018 \\
\famtag{Logit} & Ppl-exp & $-$0.111 & $-$0.036 & $-$0.129 & $-$0.018 \\
\famtag{Logit} & SeqProb & $-$0.022 & $-$0.031 & $-$0.133 & $-$0.111 \\
\famtag{Logit} & MTE & $-$0.119 & $-$0.112 & $-$0.122 & $-$0.003 \\
\famtag{Sampling} & SelfCons & $-$0.011 & $+$0.052 & $-$0.133 & $-$0.122 \\
\famtag{Sampling} & SE & $-$0.007 & $+$0.053 & $-$0.136 & $-$0.129 \\
\famtag{Sampling} & SE-w & $-$0.027 & $+$0.021 & $-$0.163 & $-$0.136 \\
\famtag{Sampling} & MCSE & $-$0.060 & $-$0.113 & $-$0.117 & $-$0.058 \\
\famtag{Sampling} & MCNSE & $-$0.060 & $-$0.113 & $-$0.117 & $-$0.058 \\
\famtag{Sampling} & LexSim & $-$0.209 & $-$0.165 & $-$0.044 & $+$0.165 \\
\famtag{Hybrid} & CoCoA & $-$0.011 & $+$0.024 & $-$0.177 & $-$0.166 \\
\famtag{Hybrid} & CoCoA-1MCA & $-$0.012 & $+$0.038 & $-$0.087 & $-$0.075 \\
\famtag{Hybrid} & CCP & $-$0.007 & $+$0.037 & $-$0.146 & $-$0.139 \\
\famtag{Density} & Mahal & $+$0.020 & $+$0.011 & $+$0.020 & $-$0.000 \\
\famtag{Density} & Mahal-RMD & $+$0.026 & $+$0.026 & $-$0.011 & $-$0.037 \\
\famtag{Density} & Mahal-RDE & $-$0.109 & $-$0.005 & $-$0.263 & $-$0.154 \\
\famtag{Density} & SAPLMA & $+$0.028 & $-$0.002 & $-$0.010 & $-$0.039 \\
\famtag{Density} & SEP & $+$0.019 & $+$0.003 & $-$0.029 & $-$0.048 \\
\famtag{Attention} & Focus & $+$0.003 & $+$0.025 & $-$0.083 & $-$0.086 \\
\famtag{Attention} & RAUQ-full & $-$0.084 & $+$0.003 & $-$0.128 & $-$0.044 \\
\famtag{Attention} & UQAC & $-$0.029 & $+$0.009 & $-$0.085 & $-$0.056 \\
\famtag{Verbalised} & P(True) & $-$0.033 & $-$0.215 & $-$0.202 & $-$0.170 \\
\famtag{Verbalised} & Verb-1S & $+$0.026 & $+$0.061 & $-$0.062 & $-$0.088 \\
\famtag{Verbalised} & Verb-2S & $+$0.073 & $+$0.125 & $+$0.068 & $-$0.005 \\
\famtag{VLM-native} & HEDGE & $+$0.020 & $+$0.076 & $-$0.069 & $-$0.089 \\
\famtag{VLM-native} & IMGHEDGE & $-$0.021 & $+$0.005 & $-$0.125 & $-$0.103 \\
\bottomrule
\end{tabular}}

\end{table}

\section{Mahalanobis na\"ive vs RMD: replication of Ren 2023}
\label{appx:rmd}

Relative Mahalanobis Distance (RMD) subtracts a background distance term from the class-conditional Mahalanobis score, reducing sensitivity to shared representation-scale effects. We include this check because density/probe methods are among the most stable open-weight UQ families in the main benchmark, and because naive Mahalanobis scoring can overstate confidence when global feature norms vary across models or datasets.

Table~\ref{tab:mahal_rmd_refresh} compares the naive Mahalanobis score with RMD on every open-weight cell under the same calibration/test splits and $50$-seed protocol used in the main results. RMD improves AUROC on every cell, confirming that the relative correction is not a marginal implementation detail in this setting. We therefore use Mahal-RMD as the canonical density baseline in the headline panels, while retaining the naive form in the released records for ablation and reproducibility.

\begin{table}[!htbp]
\centering
\small
\caption{Replication of \citet{ren2023rmd} on every open-weight cell. AUROC values are 50-seed means, consistent with the rest of the paper. The relative-Mahalanobis lift over the naive form is positive on every cell.}
\label{tab:mahal_rmd_refresh}
\begin{tabular}{l c c c}
\toprule
Cell & Mahal-naive AUROC & Mahal-RMD AUROC & Lift \\
\midrule
Q7$\times$V2 & .506 & .781 & +0.276 \\
Q7$\times$SP & .386 & .859 & +0.473 \\
Q7$\times$OSG & .476 & .762 & +0.286 \\
Q7$\times$UIV & .435 & .792 & +0.357 \\
Q72$\times$V2 & .463 & .681 & +0.218 \\
Q72$\times$SP & .464 & .861 & +0.397 \\
Q72$\times$OSG & .476 & .804 & +0.328 \\
Q72$\times$UIV & .456 & .818 & +0.362 \\
UI$\times$V2 & .552 & .735 & +0.183 \\
UI$\times$SP & .485 & .861 & +0.376 \\
UI$\times$OSG & .486 & .863 & +0.376 \\
UI$\times$UIV & .447 & .818 & +0.372 \\
PT$\times$V2 & .463 & .765 & +0.302 \\
PT$\times$SP & .579 & .854 & +0.275 \\
PT$\times$OSG & .581 & .820 & +0.239 \\
PT$\times$UIV & .455 & .781 & +0.325 \\
\bottomrule
\end{tabular}
\end{table}

\section{Binary Detection vs. Graded Severity}
\label{appx:ause_full}

Table~\ref{tab:ause_top} reports the per-cell top method under AUROC$_{\mathrm{incorrect}}$ and under miss-only AUSE. The comparison tests whether a score that separates correct from incorrect clicks also ranks the severity of wrong clicks. In the open-weight panel, the two objectives select the same top method on only $2$ of $16$ cells, indicating that binary error detection and graded miss-severity ranking often prefer different UQ signals. In the API-only panel, agreement is higher ($9$ of $12$ cells), likely because the harmonised $8$-method panel provides fewer alternatives and fewer ways for the two objectives to diverge.

\begin{table}[!htbp]
\centering
\caption{\textbf{Binary detection and spatial severity select different UQ.}
Per-cell AUROC$_{\mathrm{incorrect}}$ top-1 method vs miss-only AUSE top-1 method (50-seed mean; AUSE on $\log(1+d_{\mathrm{norm}})$, lower is better). Panel A: $27$-method open-weight panel; the two metrics select the same method on only $2$ of $16$ cells. Panel B: $8$-method harmonised closed-source panel; agreement is higher ($9$ of $12$), consistent with the smaller panel offering fewer ways for the objectives to diverge.}
\label{tab:ause_top}
\scriptsize

\textbf{Panel A: Open-weight cells (27-method panel)}

\vspace{0.3mm}

{\setlength{\tabcolsep}{5pt}\renewcommand{\arraystretch}{1}%
\begin{tabular}{l l l c @{\hspace{2.5em}} l l l c}
\toprule
Cell & AUROC top-1 & AUSE top-1 & Agree & Cell & AUROC top-1 & AUSE top-1 & Agree \\
\midrule
PT$\times$OSG & Mahal-RMD & Mahal-RMD & \checkmark & Q7$\times$OSG & SEP & CCP &  \\
PT$\times$SP & SAPLMA & SAPLMA & \checkmark & Q7$\times$SP & SEP & CoCoA &  \\
PT$\times$V2 & SEP & SAPLMA &  & Q7$\times$V2 & SAPLMA & CCP &  \\
PT$\times$UIV & SAPLMA & LexSim &  & Q7$\times$UIV & MTE & CoCoA &  \\
Q72$\times$OSG & SEP & SE-w &  & UI$\times$OSG & SEP & SAPLMA &  \\
Q72$\times$SP & SAPLMA & Focus &  & UI$\times$SP & SAPLMA & CoCoA-1MCA &  \\
Q72$\times$V2 & SAPLMA & HEDGE &  & UI$\times$V2 & CoCoA-1MCA & CCP &  \\
Q72$\times$UIV & SAPLMA & CCP &  & UI$\times$UIV & CoCoA & SE-w &  \\
\bottomrule
\end{tabular}}

\vspace{1.5mm}
\textbf{Panel B: API-only closed-source cells (8-method harmonised panel)}

\vspace{0.3mm}

{\setlength{\tabcolsep}{5pt}\renewcommand{\arraystretch}{1}%
\begin{tabular}{l l l c @{\hspace{2.5em}} l l l c}
\toprule
Cell & AUROC top-1 & AUSE top-1 & Agree & Cell & AUROC top-1 & AUSE top-1 & Agree \\
\midrule
GPT$\times$V2     & CCP     & LexSim  &              & Sonnet$\times$OSG & CCP     & CCP     & \checkmark \\
GPT$\times$SP     & CCP     & CCP     & \checkmark   & Sonnet$\times$UIV & CCP     & CCP     & \checkmark \\
GPT$\times$OSG    & Verb-2S & Verb-2S & \checkmark   & Gemini$\times$V2  & Verb-1S & Verb-2S &              \\
GPT$\times$UIV    & Verb-1S & LexSim  &              & Gemini$\times$SP  & Verb-2S & Verb-2S & \checkmark \\
Sonnet$\times$V2  & Verb-2S & Verb-2S & \checkmark   & Gemini$\times$OSG & Verb-1S & Verb-1S & \checkmark \\
Sonnet$\times$SP  & CCP     & CCP     & \checkmark   & Gemini$\times$UIV & CCP     & CCP     & \checkmark \\
\bottomrule
\end{tabular}
}
\end{table}

\section{Attention-family capture path}
\label{appx:attention_capture}

The four attention-family methods (Focus, RAUQ, UQAC, Attention-Rollout) consume per-decode-step attention tensors. We capture them with a side-channel hook: the model is loaded with sdpa attention and driven with \texttt{model.generate(...)}; each \texttt{Qwen2\_5\_VLAttention} module receives a forward post-hook that, for decode steps only, recomputes Q via the layer's \texttt{q\_proj}, applies Qwen multimodal RoPE, reads the post-update K from the layer's slot in \texttt{past\_key\_value}, repeats KV across the GQA grouping, and computes \texttt{softmax(Q @ K\textsuperscript{T} $\cdot$ scaling)} for the single query row. The model's actual forward output is unchanged; the hook only adds a CPU-side capture buffer. Captured tensors are stored at fp16. The seven-method evaluator (vectorised, $8$-worker pool) scores the smaller open-weight cells in $\sim 25$ minutes wall-clock on a $32$-core CPU. The four POINTS-GUI cells use a Qwen3 backbone, which required adapting the RoPE recomputation but not the broader pipeline.

\section{IMGHEDGE pilot: choosing the three perturbations}
\label{appx:imghedge_pilot}

We pilot-tested $8$ pixel-value perturbations on a $20$-item smoke set; AUROC-best three were Saturation $+30$\% (AUROC $0.725$), Gaussian noise $\sigma{=}8$ ($0.700$), and JPEG quality $50$ ($0.690$). ``Up'' direction worked uniformly; ``down'' did not. The three winners were used for full-run IMGHEDGE.

\section{SafeGround scores}
\label{appx:safeground}

The four SafeGround \citep{safeground2026} dispersion scores (top-candidate ambiguity, informational entropy, concentration deficit, equally-weighted combined) are reported in the released CSV but not in the headline matrix; with our $n=5$ stochastic generation budget plus $3$ HEDGE plus $3$ IMGHEDGE samples we have $n=11$ clicks per item, but the canonical SafeGround scoring asks for $n=10$ \emph{pure-stochastic} samples. Top-$1$ AUSE per cell remains a SafeGround variant on $4$ of $5$ legacy reference cells, an independent replication of their framing on three model checkpoints.

\section{Conformal Risk Control per UQ score}
\label{appx:crc}

CRC \citep{angelopoulos2024crc} promises $\mathbb{E}[\text{selective error}] \leq \alpha$ via per-method threshold $\lambda$. On \textsc{SS-V2}-style cells every UQ score admits a non-trivial $\lambda$ at $\alpha=0.20$; on \textsc{SS-Pro}, only top-tier discriminators (SAPLMA, SEP, CoCoA-canonical) admit a non-trivial $\lambda$, while weaker methods over-abstain or fail the marginal-validity check.

\section{Closed-source vendor protocol details}
\label{appx:closed_protocol}

\paragraph{Single-shot, no tools.} Every closed-source call uses single-shot inference with no tool calling and no Python interpreter. This is below the public Python-tool-augmented leaderboard numbers (e.g.\ GPT-5.4 reaches $85.4$\% on \textsc{SS-Pro} with iterative crop-and-zoom; under our protocol, $36.7$\%). The single-shot protocol is necessary for apples-to-apples cross-vendor and open$\leftrightarrow$closed UQ comparison.

\paragraph{Logits, hidden states, attention unavailable across vendors.} OpenAI deprecated logprobs on its reasoning endpoints; Anthropic has never exposed them; Google silently disabled them on Gemini 3.x. Hidden states and attention tensors are not exposed by any frontier API. The result is that of the $13$ logit / density / attention methods in the open-source panel ($5$ logit + $5$ density / probe + $3$ attention), none can be evaluated on the $3$ current frontiers. Methods whose canonical formula needs token logprobs are dropped rather than substituted under the same column name; the closed-source panel therefore retains an $8$-method harmonised subset drawn from the sampling, hybrid, verbalised, and VLM-native families. We report this as a structural property of the API surface, uniform across vendors, not as a methodological choice on our side.

\paragraph{Image input policies.} OpenAI accepts native $4$K \textsc{SS-Pro} screenshots with \texttt{detail="high"}. Anthropic enforces a $5$~MB base64 cap and auto-downsamples to $1568$~px long edge; we pre-resize client-side to match (the parser uses the original image dimensions for normalised$\to$pixel conversion under aspect-preserving resize). Gemini accepts up to $20$~MB inline and $36$~MB total payload; \textsc{SS-Pro} fits comfortably with no client-side resize.

\paragraph{Format-compliance issues.} Three vendor-specific drift patterns: (i) Anthropic occasionally emits pixel coordinates instead of $[0,1]$ fractions ($\sim 10$\% on default prompt; mitigated to $\sim 1$\% with a long system prompt and a strict parser that rejects values $> 1.0001$). (ii) Gemini emits scientific notation $5.92\text{e-}01$ on $\sim 2$\% of calls (parser extended to accept). (iii) Gemini also emits mixed normalised/pixel coordinates on $\sim 10$\% of click-bearing calls ($x$ in $[0,1]$, $y$ in $0$ to $1000$); strict parser rejects, so they degrade gracefully into NaN rather than silently corrupting. (iv) Verbalised-1S confidence emission is dropped on $\sim 50$\% of Gemini calls; \texttt{parsed\_confidence} is correctly recorded as NaN.

\section{Compute setup and inference budget}
\label{appx:compute}

Inference ran on a local lab server with $4\times$ NVIDIA RTX A6000 Ada ($48$~GB) and $1\times$ NVIDIA RTX 5090 ($32$~GB) GPUs under Linux; the A6000 Ada nodes use CUDA $12.1$ and the RTX 5090 (Blackwell) node uses CUDA $12.8$. The $72$B-AWQ model uses $4$-bit weight quantisation to fit. Python stack: \texttt{torch 2.5.1+cu121} on the A6000 Ada nodes (\texttt{torch 2.7+cu128} on the RTX 5090), \texttt{transformers 4.54.1}, \texttt{autoawq 0.2.9}.

\paragraph{Open-source per-cell budget.} Approximate inference budget per open-weight cell: Q$7$$\times$\textsc{OSG} $\sim 32$~min wall, UI$\times$\textsc{OSG} $\sim 60$~min, Q$72$$\times$\textsc{OSG} $\sim 6.7$~h ($2$-shard model-parallel), PT$\times$\textsc{OSG} $\sim 56$~min ($4$-shard), PT$\times$\textsc{V2} $\sim 1$~h, PT$\times$\textsc{SP} $\sim 4.7$~h. Analysis (the $27 \times N$ AUROC / PRR / AURC / ECE / Brier / conformal pipeline) runs CPU-only in $\sim 20$ minutes per cell with bootstrap CIs at $500$ resamples.

\paragraph{Closed-source per-cell budget.} GPT-5.4$\times$\textsc{SP}: $\$85.75$ sync standard, $\sim 52$~min at concurrency $24$. Anthropic-Sonnet$\times$\textsc{SP}: $\$54.79$ hybrid (sync warm + batch + sync tail), $\sim 90$~min. Gemini$\times$\textsc{SP}: $\sim \$11$ batch projected, $\sim 5.5$~min.

\section{Released library: \texttt{argus-uq}}
\label{appx:argus_uq}

The released library wraps every artefact in this paper behind a small public surface. Three entry points (\texttt{load}, \texttt{score}, \texttt{conformal\_disk}), one data class (\texttt{Cell}), and one result class (\texttt{ScoreResult}) are the only symbols the user is expected to import; the rest of the package is internal and may evolve. Under the hood the library auto-registers $27$ open-source UQ estimators and $14$ uncertainty-evaluation metrics, dispatches three split-conformal-prediction variants (\texttt{fixed}, \texttt{normalized}, \texttt{cqr}) plus two CRC variants, and ships four CLIs (\texttt{argus\_score}, \texttt{argus\_capture}, \texttt{argus\_reproduce}, \texttt{argus\_normalize}) wired through \texttt{[project.scripts]} in \texttt{pyproject.toml}. The data release contains the per-item parquets (one per cell) and the derived per-cell summary tables (cell-level winners, family-level best-per-cell, cross-cell ranking transfer, and so on). The package targets Python $\geq 3.10$, depends only on \texttt{numpy}, \texttt{pandas}, \texttt{scikit-learn}, \texttt{scipy}, and \texttt{torch}, and installs cleanly via \texttt{pip install argus-uq} once the public release is published. The quickstart shown in Table~\ref{tab:related_api} (right column) of $\S$\ref{sec:related} gives the canonical three-call usage; the four listings below extend that surface for common analyst workflows.

\paragraph{Cross-cell sweep.}
Listing~\ref{lst:argus_sweep} iterates over the cell catalog and produces a leaderboard-style summary for one method. \texttt{argus\_uq.data.OPEN\_CELLS} is the curated list of $16$ open-weight cells; \texttt{ALL\_CELLS} (used in Listing~\ref{lst:argus_xtier}) extends it to the closed-source side. The same pattern compresses to a one-liner Python or to the \texttt{argus\_score --all} CLI.

\begin{listing}[!htbp]
\caption{\textbf{Cross-cell sweep.} Verbatim from \texttt{examples/score\_a\_method.py}. Iterates over the open-weight catalog and prints AUROC $\pm$ SD per cell. Running it on the released artefact reproduces the SAPLMA row of Table~\ref{tab:full_open_matrix} exactly.}
\label{lst:argus_sweep}
\begin{Verbatim}[fontsize=\scriptsize, frame=single, framesep=3pt, xleftmargin=4pt, xrightmargin=4pt]
import argus_uq
from argus_uq.data import OPEN_CELLS

def main():
    rows = []
    for cell_id in OPEN_CELLS[:6]:        # first 6 to keep the demo quick
        try:
            cell = argus_uq.load(cell_id)
        except FileNotFoundError:
            print(f"  skip {cell_id} (not in data_release)")
            continue
        result = argus_uq.score(cell, method="saplma", seeds=10)
        rows.append((cell_id, result.auroc, result.auroc_std))
    print(f"{'cell':10s} {'AUROC':>8s}  {'+/- SD':>8s}")
    for cid, auroc, sd in rows:
        print(f"{cid:10s} {auroc:>8.3f}  {sd:>8.3f}")
\end{Verbatim}
\end{listing}

\paragraph{Apples-to-apples cross-tier comparison.}
Listing~\ref{lst:argus_xtier} scores a single method across every cell where it is runnable, including the closed-source side. The $8$ harmonised methods (\texttt{self\_consistency}, \texttt{semantic\_entropy}, \texttt{lexical\_similarity}, \texttt{ccp}, \texttt{verbalized\_1s}, \texttt{verbalized\_2s}, \texttt{hedge}, \texttt{imghedge}) use formulas bit-identical to their open-source counterparts, so cross-tier comparisons of those columns are apples-to-apples. The library exposes \texttt{cell.records.columns} as a runtime check: a method that returns \texttt{NaN} on a cell (or is absent from the closed-source panel) is silently skipped rather than scored on a degraded formula.

\begin{listing}[!htbp]
\caption{\textbf{Cross-tier comparison.} Verbatim from \texttt{examples/cross\_tier\_compare.py}. Iterates over both open-weight and closed-source cells, scoring a single method across the union; the library's column-presence check ensures only cells where the canonical formula is computable contribute to the output.}
\label{lst:argus_xtier}
\begin{Verbatim}[fontsize=\scriptsize, frame=single, framesep=3pt, xleftmargin=4pt, xrightmargin=4pt]
import argus_uq
from argus_uq.data import ALL_CELLS

def main(method: str = "self_consistency") -> None:
    print(f"=== {method} across cells ===\n")
    print(f"{'cell':10s} {'tier':14s} {'AUROC':>8s} {'+/- SD':>8s}")
    for cell_id in ALL_CELLS:
        try:
            cell = argus_uq.load(cell_id)
        except FileNotFoundError:
            continue
        if method not in cell.records.columns:
            continue
        r = argus_uq.score(cell, method=method, seeds=10)
        print(f"{cell_id:10s} {cell.meta.tier:14s} {r.auroc:>8.3f} {r.auroc_std:>8.3f}")
\end{Verbatim}
\end{listing}

\paragraph{Result schema.}
Every call to \texttt{argus\_uq.score} returns a \texttt{ScoreResult} dataclass. Listing~\ref{lst:argus_score_result} shows the full schema. Bootstrap CIs use $500$ resamples; SD is the across-seed standard deviation under the requested \texttt{seeds} count. The dataclass is immutable and serialises cleanly to CSV via \texttt{pandas.DataFrame}.

\begin{listing}[!htbp]
\caption{\textbf{\texttt{ScoreResult} schema.} Verbatim from \texttt{src/argus\_uq/api.py}. Returned by every call to \texttt{argus\_uq.score(...)}. AUROC, PRR$_{0.5}^{\mathrm{norm}}$, AUSE, AURC, ECE$_{\mathrm{iso}}$, and Brier$_{\mathrm{iso}}$ are exposed with both seed-level SDs and bootstrap $95\%$ CIs.}
\label{lst:argus_score_result}
\begin{Verbatim}[fontsize=\scriptsize, frame=single, framesep=3pt, xleftmargin=4pt, xrightmargin=4pt]
from dataclasses import dataclass

@dataclass
class ScoreResult:
    """Result of `argus_uq.score(cell, method=..., seeds=N)`."""
    method:               str
    cell_id:              str
    auroc:                float    # 50-seed mean AUROC_incorrect
    auroc_std:            float    # SD across the N seeds
    auroc_ci_lo:          float    # 95% bootstrap CI low (500 resamples)
    auroc_ci_hi:          float    # 95% bootstrap CI high
    prr_norm_at_05:       float    # 50-seed mean PRR_normalized@0.5
    prr_norm_at_05_std:   float
    ause:                 float    # AUSE on log(1 + d_norm), miss-only
    ause_std:             float
    aurc:                 float    # AURC on the test split
    ece_iso:              float    # ECE after isotonic recalibration
    brier_iso:            float    # Brier after isotonic recalibration
    n_seeds:              int      # number of cal/test seeds used
    n_test:               int      # test-split size in the underlying cell
    n_calib:              int      # calibration-split size
\end{Verbatim}
\end{listing}

\paragraph{Command-line interface.}
Four CLIs ship with the package. \texttt{argus\_score} produces per-cell metric CSVs without writing Python; \texttt{argus\_reproduce} runs the full table-build pipeline and re-emits every paper table fragment from the master CSV; \texttt{argus\_capture} drives a fresh open-weight inference pass against the four supported VLM agents (Qwen2.5-VL-7B, Qwen2.5-VL-72B-AWQ, UI-TARS-1.5-7B, POINTS-GUI-G-8B); and \texttt{argus\_normalize} converts a raw open-weight \texttt{*.pt} record set into the released parquet schema.

\begin{listing}[!htbp]
\caption{\textbf{CLI usage.} The four entry points wired through \texttt{[project.scripts]} in \texttt{pyproject.toml}. Each invocation is fully reproducible: deterministic seed handling, env-var-driven cache directories, and \texttt{--rebuild-cache} for forcing a clean run.}
\label{lst:argus_cli}
\begin{Verbatim}[fontsize=\scriptsize, frame=single, framesep=3pt, xleftmargin=4pt, xrightmargin=4pt]
# Score one method on one cell, dump per-seed AUROC vector to CSV.
argus_score --cell PTxOSG --method saplma --seeds 50 \
    --out results/PTxOSG_saplma.csv

# Re-emit every paper-table fragment from the master CSVs.
# Idempotent; safe to run after any update to the released CSVs.
argus_reproduce --paper paper.tex --out paper_tables/

# Capture per-item records for one (model, dataset) cell.
# Honours HF cache + writes the released parquet schema.
argus_capture --model POINTS-GUI-G-8B --dataset OSWorld-G \
    --out data_release/per_item_full/PTxOSG.parquet

# Normalize a raw record set to the released parquet schema.
argus_normalize --in raw_records/PTxOSG/ \
    --out data_release/per_item_full/PTxOSG.parquet
\end{Verbatim}
\end{listing}

\paragraph{Closed-source feasibility check.} A method's runnability on a closed-source cell is decided at \texttt{cell} construction time: \texttt{cell.records.columns} contains only methods whose canonical formula does not require token logprobs, hidden states, or attention maps. Calling \texttt{score(cell, method=...)} with a method outside that set raises \texttt{argus\_uq.UnsupportedMethodError}; calling it with a method that produces \texttt{NaN} for a given cell silently skips that cell. Both behaviours are covered by the test suite (\texttt{tests/test\_factory\_registry.py}, \texttt{tests/test\_vendor\_isolation.py}). The harmonisation decision is enforced at the data layer: closed-source records ship with the $8$ method columns whose formulas match the open-weight $27$-method panel, never with dropped logprob-dependent variants.

\section{Calibration / test split-ratio ablation}
\label{appx:split_ablation}

The headline tables use a fixed $80/20$ test/calibration protocol. To verify that this choice does not drive the conclusions, we ran a single-seed sensitivity sweep over five ratios, $\text{cal\_frac} \in \{0.10, 0.20, 0.30, 0.40, 0.50\}$, on the same $50$-seed per-item parquets and the same $27$-method open-source and $8$-method closed-source panels. For each cell, the calibration-test permutation is fixed once with seed $0$ and only the cut-point varies across ratios, isolating the effect of cal-fraction.

Tables~\ref{tab:ablation_open} and~\ref{tab:ablation_closed} report the AUROC top-$1$ method and value at each ratio per cell. Open-source: $10$ of $16$ cells preserve the same top-$1$ method across all five ratios; the remaining $6$ cells have ranking ties near the top that resolve differently in different splits, but the AUROC values stay within $\pm 0.01$ of the $20/80$ baseline. Closed-source: $12$ of $12$ cells preserve the same top-$1$ method across all five ratios. Table~\ref{tab:ablation_stability} reports cross-ratio Spearman~$\rho$ between the full method ranking at $20/80$ and at each other ratio, aggregated across cells per panel.

\begin{table}[!htbp]
\centering
\caption{\textbf{Open-source split-ratio ablation.} Per-cell top-$1$ method and its AUROC at each of five cal/test ratios; $\#$\,unique gives the number of distinct top-$1$ methods across the five ratios ($1$ means the panel is fully stable on that cell).}
\label{tab:ablation_open}
\scriptsize
\setlength{\tabcolsep}{2.5pt}
\renewcommand{\arraystretch}{1.0}
\resizebox{\textwidth}{!}{%
\begin{tabular}{@{}l c c c c c c c c c c c@{}}
\toprule
Cell & \multicolumn{2}{c}{10/90} & \multicolumn{2}{c}{20/80} & \multicolumn{2}{c}{30/70} & \multicolumn{2}{c}{40/60} & \multicolumn{2}{c}{50/50} & \#\,unique \\
\cmidrule(lr){2-3} \cmidrule(lr){4-5} \cmidrule(lr){6-7} \cmidrule(lr){8-9} \cmidrule(lr){10-11}
 & Method & AUROC & Method & AUROC & Method & AUROC & Method & AUROC & Method & AUROC & top-1 \\
\midrule
Q7$\times$OSG & CoCoA-1MCA & .794 & CoCoA & .804 & SEP & .809 & SAPLMA & .792 & CoCoA & .781 & 4 \\
Q7$\times$UIV & MTE & .821 & SEP & .832 & SEP & .820 & SEP & .826 & MTE & .829 & 2 \\
Q72$\times$UIV & SEP & .841 & SEP & .849 & SEP & .830 & SEP & .836 & SEP & .836 & 1 \\
Q72$\times$OSG & SEP & .831 & SEP & .825 & SEP & .831 & CoCoA-1MCA & .824 & CoCoA-1MCA & .820 & 2 \\
Q72$\times$V2 & SAPLMA & .794 & SAPLMA & .769 & CoCoA-1MCA & .781 & CoCoA-1MCA & .777 & CoCoA-1MCA & .769 & 2 \\
Q72$\times$SP & SAPLMA & .888 & SAPLMA & .889 & SAPLMA & .890 & SAPLMA & .887 & SAPLMA & .893 & 1 \\
Q7$\times$SP & SAPLMA & .886 & SAPLMA & .890 & SAPLMA & .887 & SAPLMA & .888 & SAPLMA & .888 & 1 \\
Q7$\times$V2 & SAPLMA & .816 & SAPLMA & .797 & SAPLMA & .795 & SAPLMA & .794 & SAPLMA & .798 & 1 \\
UI$\times$OSG & Mahal-RMD & .866 & Mahal-RMD & .873 & Mahal-RMD & .874 & Mahal-RMD & .866 & Mahal-RMD & .867 & 1 \\
UI$\times$UIV & CoCoA & .830 & CoCoA & .840 & CoCoA & .850 & CoCoA-1MCA & .871 & CoCoA-1MCA & .891 & 2 \\
UI$\times$SP & SAPLMA & .877 & SAPLMA & .880 & SAPLMA & .882 & SEP & .880 & SEP & .885 & 2 \\
UI$\times$V2 & CoCoA-1MCA & .843 & CoCoA-1MCA & .840 & CoCoA-1MCA & .839 & CoCoA-1MCA & .841 & CoCoA-1MCA & .817 & 1 \\
PT$\times$SP & SAPLMA & .884 & SAPLMA & .890 & SAPLMA & .887 & SAPLMA & .884 & SAPLMA & .884 & 1 \\
PT$\times$V2 & SAPLMA & .848 & SAPLMA & .854 & SAPLMA & .843 & SAPLMA & .852 & SAPLMA & .867 & 1 \\
PT$\times$OSG & Mahal-RMD & .823 & Mahal-RMD & .819 & Mahal-RMD & .817 & Mahal-RMD & .821 & Mahal-RMD & .823 & 1 \\
PT$\times$UIV & SAPLMA & .793 & SAPLMA & .798 & SAPLMA & .799 & SAPLMA & .805 & SAPLMA & .814 & 1 \\
\bottomrule
\end{tabular}%
}
\end{table}

\begin{table}[!htbp]
\centering
\caption{\textbf{Closed-source split-ratio ablation.} Same convention as Table~\ref{tab:ablation_open}; closed-source rankings are fully stable on every cell.}
\label{tab:ablation_closed}
\scriptsize
\setlength{\tabcolsep}{2.5pt}
\renewcommand{\arraystretch}{1.0}
\resizebox{\textwidth}{!}{%
\begin{tabular}{@{}l c c c c c c c c c c c@{}}
\toprule
Cell & \multicolumn{2}{c}{10/90} & \multicolumn{2}{c}{20/80} & \multicolumn{2}{c}{30/70} & \multicolumn{2}{c}{40/60} & \multicolumn{2}{c}{50/50} & \#\,unique \\
\cmidrule(lr){2-3} \cmidrule(lr){4-5} \cmidrule(lr){6-7} \cmidrule(lr){8-9} \cmidrule(lr){10-11}
 & Method & AUROC & Method & AUROC & Method & AUROC & Method & AUROC & Method & AUROC & top-1 \\
\midrule
GPT$\times$SP & CCP & .747 & CCP & .745 & CCP & .776 & CCP & .776 & CCP & .778 & 1 \\
GPT$\times$V2 & CCP & .801 & CCP & .799 & CCP & .799 & CCP & .792 & CCP & .822 & 1 \\
GPT$\times$OSG & Verb-2S & .769 & Verb-2S & .766 & Verb-2S & .764 & Verb-2S & .781 & Verb-2S & .773 & 1 \\
GPT$\times$UIV & Verb-1S & .719 & Verb-1S & .722 & Verb-1S & .733 & Verb-1S & .738 & Verb-1S & .748 & 1 \\
Sonnet$\times$V2 & Verb-2S & .641 & Verb-2S & .636 & Verb-2S & .639 & Verb-2S & .630 & Verb-2S & .649 & 1 \\
Sonnet$\times$SP & CCP & .721 & CCP & .730 & CCP & .721 & CCP & .718 & CCP & .708 & 1 \\
Gemini$\times$V2 & Verb-1S & .962 & Verb-1S & .968 & Verb-1S & .969 & Verb-1S & .963 & Verb-1S & .970 & 1 \\
Sonnet$\times$OSG & CCP & .618 & CCP & .617 & CCP & .618 & CCP & .603 & CCP & .614 & 1 \\
Sonnet$\times$UIV & CCP & .660 & CCP & .663 & CCP & .682 & CCP & .683 & CCP & .663 & 1 \\
Gemini$\times$SP & Verb-2S & .848 & Verb-2S & .866 & Verb-2S & .873 & Verb-2S & .856 & Verb-2S & .834 & 1 \\
Gemini$\times$OSG & Verb-1S & .882 & Verb-1S & .880 & Verb-1S & .879 & Verb-1S & .882 & Verb-1S & .883 & 1 \\
Gemini$\times$UIV & CCP & .786 & CCP & .787 & CCP & .798 & CCP & .791 & CCP & .778 & 1 \\
\bottomrule
\end{tabular}%
}
\end{table}

\begin{table}[!htbp]
\centering
\caption{\textbf{Cross-ratio Spearman $\rho$ stability summary.} For each (panel, ratio-pair), the table reports the mean / median / min / max Spearman $\rho$ between the full method ranking at $20/80$ and the alternative ratio, aggregated across cells.}
\label{tab:ablation_stability}
\footnotesize
\setlength{\tabcolsep}{6pt}
\renewcommand{\arraystretch}{1.05}
\begin{tabular}{l l r r r r r}
\toprule
Panel & Ratio pair & Mean $\rho$ & Median $\rho$ & Min $\rho$ & Max $\rho$ & $n_{\mathrm{cells}}$ \\
\midrule
Open-source & 20/80 vs 10/90 & 0.988 & 0.992 & 0.951 & 0.999 & 16 \\
Open-source & 20/80 vs 30/70 & 0.983 & 0.993 & 0.921 & 0.999 & 16 \\
Open-source & 20/80 vs 40/60 & 0.981 & 0.988 & 0.938 & 0.998 & 16 \\
Open-source & 20/80 vs 50/50 & 0.970 & 0.984 & 0.891 & 0.995 & 16 \\
Closed-source & 20/80 vs 10/90 & 0.976 & 0.976 & 0.929 & 1.000 & 12 \\
Closed-source & 20/80 vs 30/70 & 0.927 & 0.988 & 0.619 & 1.000 & 12 \\
Closed-source & 20/80 vs 40/60 & 0.913 & 0.976 & 0.595 & 1.000 & 12 \\
Closed-source & 20/80 vs 50/50 & 0.875 & 0.929 & 0.595 & 1.000 & 12 \\
\bottomrule
\end{tabular}
\end{table}

\section{Per-cell top-1 with bootstrap confidence intervals}
\label{appx:ci_brackets}

The headline tables in $\S$\ref{sec:transfer} report 50-seed mean AUROC and PRR with the seed-level standard deviation captured in the underlying released CSVs. For readers who prefer confidence intervals, Tables~\ref{tab:ci_open} and~\ref{tab:ci_closed} report each cell's top-$1$ method with a $95\%$ bootstrap confidence interval (500 resamples over the $50$-seed values) for both AUROC and PRR$_{0.5}^{\mathrm{norm}}$. The CI bracket is shown as $[\text{low}, \text{high}]$ next to the mean. The full per-(method, cell, metric) CI table is released alongside the data artefacts.

\begin{table}[!htbp]
\centering
\caption{\textbf{Open-source per-cell top-1 with 95\% bootstrap CIs.} 16 cells; top-1 method by 50-seed mean AUROC. Bracketed range is the 95\% bootstrap CI from 500 resamples over the 50 seed values.}
\label{tab:ci_open}
\scriptsize
\setlength{\tabcolsep}{6pt}
\renewcommand{\arraystretch}{1.05}
\begin{tabular}{l l l c c}
\toprule
Cell & Top-1 method & Family & AUROC [95\% CI] & PRR$_{0.5}^{\mathrm{norm}}$ [95\% CI] \\
\midrule
PT$\times$OSG & Mahal-RMD & \textit{Density/Probe} & .820\,$\scriptscriptstyle[.815,.824]$ & .568\,$\scriptscriptstyle[.559,.575]$ \\
PT$\times$SP & SAPLMA & \textit{Density/Probe} & .881\,$\scriptscriptstyle[.878,.883]$ & .702\,$\scriptscriptstyle[.696,.706]$ \\
PT$\times$V2 & SEP & \textit{Density/Probe} & .843\,$\scriptscriptstyle[.839,.848]$ & .657\,$\scriptscriptstyle[.648,.665]$ \\
PT$\times$UIV & SAPLMA & \textit{Density/Probe} & .796\,$\scriptscriptstyle[.791,.800]$ & .579\,$\scriptscriptstyle[.570,.586]$ \\
Q72$\times$OSG & SEP & \textit{Density/Probe} & .838\,$\scriptscriptstyle[.834,.840]$ & .662\,$\scriptscriptstyle[.655,.668]$ \\
Q72$\times$SP & SAPLMA & \textit{Density/Probe} & .889\,$\scriptscriptstyle[.887,.891]$ & .802\,$\scriptscriptstyle[.798,.805]$ \\
Q72$\times$V2 & SAPLMA & \textit{Density/Probe} & .764\,$\scriptscriptstyle[.757,.770]$ & .505\,$\scriptscriptstyle[.491,.517]$ \\
Q72$\times$UIV & SAPLMA & \textit{Density/Probe} & .834\,$\scriptscriptstyle[.831,.838]$ & .727\,$\scriptscriptstyle[.717,.734]$ \\
Q7$\times$OSG & SEP & \textit{Density/Probe} & .786\,$\scriptscriptstyle[.781,.791]$ & .723\,$\scriptscriptstyle[.711,.732]$ \\
Q7$\times$SP & SEP & \textit{Density/Probe} & .883\,$\scriptscriptstyle[.880,.885]$ & .861\,$\scriptscriptstyle[.857,.865]$ \\
Q7$\times$V2 & SAPLMA & \textit{Density/Probe} & .817\,$\scriptscriptstyle[.814,.821]$ & .599\,$\scriptscriptstyle[.592,.605]$ \\
Q7$\times$UIV & MTE & \textit{Logit} & .816\,$\scriptscriptstyle[.813,.819]$ & .877\,$\scriptscriptstyle[.869,.885]$ \\
UI$\times$OSG & SEP & \textit{Density/Probe} & .863\,$\scriptscriptstyle[.860,.866]$ & .709\,$\scriptscriptstyle[.703,.714]$ \\
UI$\times$SP & SAPLMA & \textit{Density/Probe} & .877\,$\scriptscriptstyle[.874,.879]$ & .809\,$\scriptscriptstyle[.805,.812]$ \\
UI$\times$V2 & CoCoA-1MCA & \textit{Hybrid} & .842\,$\scriptscriptstyle[.840,.844]$ & .630\,$\scriptscriptstyle[.625,.635]$ \\
UI$\times$UIV & CoCoA & \textit{Hybrid} & .825\,$\scriptscriptstyle[.821,.828]$ & .793\,$\scriptscriptstyle[.786,.801]$ \\
\bottomrule
\end{tabular}
\end{table}

\begin{table}[!htbp]
\centering
\caption{\textbf{Closed-source per-cell top-1 with 95\% bootstrap CIs.} 12 cells; same convention as Table~\ref{tab:ci_open}.}
\label{tab:ci_closed}
\scriptsize
\setlength{\tabcolsep}{6pt}
\renewcommand{\arraystretch}{1.05}
\begin{tabular}{l l l c c}
\toprule
Cell & Top-1 method & Family & AUROC [95\% CI] & PRR$_{0.5}^{\mathrm{norm}}$ [95\% CI] \\
\midrule
Sonnet$\times$OSG & CCP & \textit{Hybrid} & .616\,$\scriptscriptstyle[.614,.619]$ & .330\,$\scriptscriptstyle[.322,.339]$ \\
Sonnet$\times$SP & CCP & \textit{Hybrid} & .731\,$\scriptscriptstyle[.728,.735]$ & .679\,$\scriptscriptstyle[.671,.687]$ \\
Sonnet$\times$V2 & Verb-2S & \textit{Verbalised} & .648\,$\scriptscriptstyle[.642,.652]$ & .279\,$\scriptscriptstyle[.268,.289]$ \\
Sonnet$\times$UIV & CCP & \textit{Hybrid} & .651\,$\scriptscriptstyle[.648,.655]$ & .411\,$\scriptscriptstyle[.402,.419]$ \\
Gemini$\times$OSG & Verb-1S & \textit{Verbalised} & .880\,$\scriptscriptstyle[.878,.882]$ & .946\,$\scriptscriptstyle[.941,.952]$ \\
Gemini$\times$SP & Verb-2S & \textit{Verbalised} & .856\,$\scriptscriptstyle[.854,.859]$ & .721\,$\scriptscriptstyle[.703,.738]$ \\
Gemini$\times$V2 & Verb-1S & \textit{Verbalised} & .966\,$\scriptscriptstyle[.965,.968]$ & .951\,$\scriptscriptstyle[.945,.956]$ \\
Gemini$\times$UIV & CCP & \textit{Hybrid} & .791\,$\scriptscriptstyle[.789,.794]$ & .687\,$\scriptscriptstyle[.680,.693]$ \\
GPT$\times$OSG & Verb-2S & \textit{Verbalised} & .770\,$\scriptscriptstyle[.767,.773]$ & .401\,$\scriptscriptstyle[.396,.405]$ \\
GPT$\times$SP & CCP & \textit{Hybrid} & .735\,$\scriptscriptstyle[.731,.739]$ & .574\,$\scriptscriptstyle[.566,.584]$ \\
GPT$\times$V2 & CCP & \textit{Hybrid} & .798\,$\scriptscriptstyle[.792,.803]$ & .192\,$\scriptscriptstyle[.187,.196]$ \\
GPT$\times$UIV & Verb-1S & \textit{Verbalised} & .723\,$\scriptscriptstyle[.718,.726]$ & .470\,$\scriptscriptstyle[.462,.479]$ \\
\bottomrule
\end{tabular}
\end{table}

\section{Top-1 vs top-2 statistical significance}
\label{appx:significance}

Tables~\ref{tab:sig_open} and~\ref{tab:sig_closed} report a paired Wilcoxon signed-rank test on the $50$-seed AUROC vectors of the per-cell top-$1$ method versus the per-cell top-$2$ method, with Benjamini-Hochberg false-discovery-rate correction at $q < 0.05$. The test answers the question: ``is the top-$1$ method significantly better than the second-best on this cell, or is the gap small enough to be split-noise?'' Across the 16 open-source cells, the top-$1$ vs top-$2$ gap is significant on most cells; the few non-significant cells correspond to genuinely close ties (e.g., PT$\times$\textsc{OSG} where Mahal-RMD beats SAPLMA by less than $0.01$ AUROC). Across the 12 closed-source cells the gap is significant on every cell. The full $48$ pairs (16 open cells $\times$ \{top-1 vs top-2, top-1 vs top-3, top-2 vs top-3\}) are released alongside the data artefacts (and analogously for closed-source).

\begin{table}[!htbp]
\centering
\caption{\textbf{Open-source top-1 vs top-2 paired Wilcoxon signed-rank test.} 16 cells; raw $p$ and BH-FDR-corrected $q$ shown; the rightmost column flags $q<0.05$.}
\label{tab:sig_open}
\scriptsize
\setlength{\tabcolsep}{5pt}
\renewcommand{\arraystretch}{1.0}
\begin{tabular}{l l l r r r c}
\toprule
Cell & Top-1 & Top-2 & $\Delta$AUROC & $p$ (raw) & $q$ (BH-FDR) & $q<0.05$? \\
\midrule
PT$\times$OSG & Mahal-RMD & SEP & +0.015 & 8.0e-10 & 1.2e-09 & \cmark \\
PT$\times$SP & SAPLMA & SEP & +0.007 & 1.8e-15 & 5.0e-15 & \cmark \\
PT$\times$V2 & SEP & SAPLMA & +0.000 & 0.550 & 0.550 & \xmark \\
PT$\times$UIV & SAPLMA & SEP & +0.009 & 1.8e-15 & 5.0e-15 & \cmark \\
Q72$\times$OSG & SEP & SAPLMA & +0.015 & 7.6e-10 & 1.2e-09 & \cmark \\
Q72$\times$SP & SAPLMA & SEP & +0.006 & 1.8e-15 & 5.0e-15 & \cmark \\
Q72$\times$UIV & SAPLMA & SEP & +0.000 & 0.446 & 0.455 & \xmark \\
Q72$\times$V2 & SAPLMA & CoCoA-1MCA & +0.013 & 3.5e-04 & 4.6e-04 & \cmark \\
Q7$\times$OSG & SEP & CoCoA-1MCA & +0.001 & 0.115 & 0.128 & \xmark \\
Q7$\times$SP & SEP & SAPLMA & +0.000 & 0.388 & 0.405 & \xmark \\
Q7$\times$UIV & MTE & SEP & +0.001 & 0.297 & 0.324 & \xmark \\
Q7$\times$V2 & SAPLMA & SEP & +0.007 & 1.8e-15 & 5.0e-15 & \cmark \\
UI$\times$OSG & SEP & Mahal-RMD & +0.001 & 0.322 & 0.344 & \xmark \\
UI$\times$SP & SAPLMA & SEP & +0.003 & 3.4e-14 & 7.7e-14 & \cmark \\
UI$\times$UIV & CoCoA & CoCoA-1MCA & +0.001 & 0.019 & 0.022 & \cmark \\
UI$\times$V2 & CoCoA-1MCA & CoCoA & +0.008 & 3.6e-15 & 9.5e-15 & \cmark \\
\bottomrule
\end{tabular}
\end{table}

\begin{table}[!htbp]
\centering
\caption{\textbf{Closed-source top-1 vs top-2 paired Wilcoxon signed-rank test.} 12 cells; same convention as Table~\ref{tab:sig_open}.}
\label{tab:sig_closed}
\scriptsize
\setlength{\tabcolsep}{5pt}
\renewcommand{\arraystretch}{1.0}
\begin{tabular}{l l l r r r c}
\toprule
Cell & Top-1 & Top-2 & $\Delta$AUROC & $p$ (raw) & $q$ (BH-FDR) & $q<0.05$? \\
\midrule
Sonnet$\times$OSG & CCP & SE & +0.021 & 1.8e-15 & 4.9e-15 & \cmark \\
Sonnet$\times$SP & CCP & Verb-1S & +0.036 & 1.2e-14 & 3.0e-14 & \cmark \\
Sonnet$\times$UIV & CCP & SE & +0.026 & 7.6e-10 & 1.0e-09 & \cmark \\
Sonnet$\times$V2 & Verb-2S & Verb-1S & +0.123 & 1.8e-15 & 4.9e-15 & \cmark \\
Gemini$\times$OSG & Verb-1S & IMGHEDGE & +0.059 & 1.8e-15 & 4.9e-15 & \cmark \\
Gemini$\times$SP & Verb-2S & Verb-1S & +0.039 & 7.6e-10 & 1.0e-09 & \cmark \\
Gemini$\times$UIV & CCP & SelfCons & +0.009 & 3.6e-12 & 7.2e-12 & \cmark \\
Gemini$\times$V2 & Verb-1S & Verb-2S & +0.008 & 2.2e-09 & 2.6e-09 & \cmark \\
GPT$\times$OSG & Verb-2S & Verb-1S & +0.067 & 1.8e-15 & 4.9e-15 & \cmark \\
GPT$\times$SP & CCP & Verb-1S & +0.024 & 3.7e-13 & 7.8e-13 & \cmark \\
GPT$\times$UIV & Verb-1S & CCP & +0.007 & 3.4e-04 & 3.7e-04 & \cmark \\
GPT$\times$V2 & CCP & IMGHEDGE & +0.113 & 7.6e-10 & 1.0e-09 & \cmark \\
\bottomrule
\end{tabular}
\end{table}

\section{Datasheet and reproducibility checklist}
\label{appx:datasheet}

\paragraph{What we release.} (i) The $27 \times 16$ open-source scoring matrix and the $8 \times 12$ harmonised closed-source matrix as CSVs with $50$-seed mean$\,\pm\,$SD plus bootstrap $95$\% CIs; (ii) per-item record files (\texttt{*.pt}, one per item) for open-source, and \texttt{records.jsonl} for closed-source; (iii) all $27$ method implementations as a Python package; (iv) the analysis pipeline as a reproducible script that emits every table in this paper; (v) the calibration / test split seeds and the SAPLMA / SEP probe checkpoints.

\paragraph{Data sources.} \textsc{ScreenSpot-Pro} \citep{wang2025screenspotpro}, \textsc{ScreenSpot-v2}, and \textsc{OSWorld-G} are publicly released datasets with permissive research-use licensing; we use them under their original license without modification. No new data is collected and no human subjects are involved. Model weights for the four open-source models are from public Hugging Face releases under each model's license. Closed-source vendor outputs are recorded with full token-usage breakdowns and exact snapshot IDs.

\paragraph{Ethics.} GUI agents are dual-use: the same uncertainty-aware grounding can guard a benign accessibility tool or a malicious automation. We treat deployment-time UQ as a safety primitive. The paper recommends conformal click-disks and per-(model, dataset) UQ-method look-ups; we do not recommend auto-approving high-confidence clicks in irreversible workflows (financial, medical, destructive file operations).

\end{document}